\documentclass[a4paper,fleqn]{cas-sc}

\usepackage[english]{babel}
\usepackage{apacite} 
\AtBeginDocument{
    
    \renewcommand{\BRetrieved}[1]{}
}
\makeatletter
\def\citep#1{\shortcite{#1}}
\def\citet#1{\shortciteA{#1}}
\makeatother

\usepackage{subcaption} 
\usepackage{amsmath, amssymb, amsfonts, amsthm}
\usepackage{bm}
\usepackage{xcolor}
\usepackage{algorithm}
\usepackage{graphicx}
\usepackage{float}
\usepackage[noend]{algpseudocode}

\newtheorem{theorem}{Theorem}

\newtheorem{definition}{Definition}
\newtheorem{assumption}{Assumption}
\newtheorem{proposition}{Proposition}
\theoremstyle{remark}
\newtheorem{remark}{Remark}

\def\tsc#1{\csdef{#1}{\textsc{\lowercase{#1}}\xspace}}
\tsc{WGM}
\tsc{QE}

\begin{document}
\let\WriteBookmarks\relax
\def\floatpagepagefraction{1}
\def\textpagefraction{.001}

\shorttitle{}    

\shortauthors{}  

\title [mode = title]{UbiQVision: Quantifying Uncertainty in XAI for Image Recognition}



%

\author[1,2]{Akshat Dubey}[orcid=0009-0008-4823-9375]
\cormark[1]
\cortext[1]{Corresponding author}
\ead{DubeyA@rki.de; dubea98@zedat.fu-berlin.de}


\author[1]{Aleksandar Anžel}[orcid=0000-0002-0678-2870]
\ead{AnzelA@rki.de}

\author[1]{Bahar İlgen}[orcid=0000-0001-5725-0850]
\ead{IlgenB@rki.de}

\author[1,2]{Georges Hattab}[orcid=0000-0003-4168-8254]
\ead{HattabG@rki.de; georges.hattab@fu-berlin.de}






\affiliation[1]{organization={Center for Artificial Intelligence in Public Health Research (ZKI-PH), Robert Koch Institute},
            addressline={Nordufer 20}, 
            city={Berlin},
         citysep={}, 
            postcode={13353}, 
            country={Germany}}






\affiliation[2]{organization={Department of Mathematics and Computer Science, Freie Universität Berlin},
            addressline={Arnimallee 14}, 
            city={Berlin},
         citysep={}, 
            postcode={14195}, 
            country={Germany}}




\begin{abstract}
Recent advances in deep learning have led to its widespread adoption across diverse domains, including medical imaging. This progress is driven by increasingly sophisticated model architectures, such as ResNets, Vision Transformers, and Hybrid Convolutional Neural Networks, that offer enhanced performance at the cost of greater complexity. This complexity often compromises model explainability and interpretability. SHAP has emerged as a prominent method for providing interpretable visualizations that aid domain experts in understanding model predictions. However, SHAP explanations can be unstable and unreliable in the presence of epistemic and aleatoric uncertainty. In this study, we address this challenge by using Dirichlet posterior sampling and Dempster-Shafer theory to quantify the uncertainty that arises from these unstable explanations in medical imaging applications. The framework uses a belief, plausible, and fusion map approach alongside statistical quantitative analysis to produce quantification of uncertainty in SHAP. Furthermore, we evaluated our framework on three medical imaging datasets with varying class distributions, image qualities, and modality types which introduces noise due to varying image resolutions and modality-specific aspect covering the examples from pathology, ophthalmology, and radiology, helping us to study the epistemic uncertainty.
\end{abstract}



\maketitle
\section{Introduction}
The adoption of deep learning (DL) in medical imaging~\citep{lee2017deep, eli2024deep} has accelerated due to the obviation of the need for manual feature engineering~\citep{li2023medical} and enhanced diagnostic performance~\citep{prasad2024revolutionizing}. The pervasive digitization of medical records through electronic health records (EHRs) and digital imaging systems has given rise to extensive repositories of diverse imaging data (X-rays, MRIs, CTs, ultrasounds) that provide the substantial large-scale datasets necessary for training deep learning models~\citep{morris2018reinventing, adeghe2024role, oyeniyi2024emerging}. The efficacy of these algorithms in critical clinical tasks, such as image classification, lesion detection, organ segmentation, and disease progression prediction, has been well-documented~\citep{mall2023comprehensive}. A notable benefit of these algorithms is the significant reduction in diagnostic time and cognitive burden on physicians that they provide~\citep{shin2023impact}. The integration of automated early disease detection, accelerated and streamlined diagnostic workflows, and the capacity to amalgamate imaging data with other patient-specific information has resulted in a paradigm shift within the domain of medical imaging, thereby evolving it into a personalized, data-driven diagnostic~\citep{ahmad2024early, taguelmimt2025towards, esteban2025integrating}. 
Convolutional neural networks (CNNs) have achieved remarkable success in medical diagnostics, often matching or surpassing the performance of board-certified physicians in specific fields~\citep{liu2019comparison}. In dermatology, for example, CNNs have been shown to match or exceed the accuracy of experienced dermatologists in classifying skin lesions as malignant or benign. Multiple studies involving up to 157 dermatologists indicate that CNNs consistently deliver comparable sensitivity and specificity~\citep{esteva2021deep}. Notably, the introduction of CNN assistance has led to a significant improvement in dermatologists' accuracy when interpreting cutaneous tumors. Dermatologists benefit the most from such technological support~\citep{ba2022convolutional}. In ophthalmology, CNNs demonstrate physician-level grading capabilities for diabetic retinopathy by analyzing fundus photographs~\citep{manikandan2023deep}. These systems have received FDA clearance as diagnostic tools~\citep{esteva2021deep}. Furthermore, CNNs can diagnose various vision-threatening conditions, such as diabetic macular edema, age-related macular degeneration, and glaucoma~\citep{manikandan2023deep}. The recent emergence of Vision Transformers (ViTs) represents a significant architectural advancement in medical imaging, as they are increasingly surpassing the performance of traditional CNNs~\citep{aburass2025vision}. ViTs use self-attention mechanisms to process image patches, which allows them to capture both local and global dependencies across images~\citep{barekatain2025evaluating}. A review of over 200 studies reveals that ViTs achieve promising results in various medical imaging applications, including breast cancer detection, brain tumor analysis, lung disease diagnosis, retinal evaluations, and assessing patients with suspected SARS-CoV-2 infection~\citep{aburass2025vision}. The transformer architecture's inherent attention mechanism allows models to prioritize different input sections based on their significance, making ViTs particularly effective for tasks that require long-range dependency modeling. 
The hybrid CNN-ViT model integrates convolutional layers, which facilitate local texture extraction, with Vision Transformer attention, which enables global spatial reasoning~\citep{liu2025enhancing}. This integration results in a model that exhibits superior performance on complex multi-organ segmentation and volumetric medical imaging tasks when compared to models that utilize only one architecture.\\
\\
Despite these impressive performance gains across CNNs, ViTs, and hybrid architectures in critical tasks such as lesion detection and disease prediction, their clinical deployment remains fundamentally limited by a lack of transparent and reliable reasoning mechanisms. This motivates the need to study explainability and uncertainty together, as their intersection is essential for enabling trustworthy and clinically robust medical imaging AI~\citep{dubey2024nested, dubey2024ai}. To address this challenge, the field of explainable artificial intelligence (XAI) has developed a suite of validation frameworks such as saliency maps, gradient methods, concept activation vectors, Shapley Additive exPlanations (SHAP), \textit{etc}.  Gradient based method which includes Grad-CAM~\citep{selvaraju2017grad}, GradCAM++~\citep{chattopadhay2018grad}, XGradCAM~\citep{fu2020axiom}, AblationCAM~\citep{ramaswamy2020ablation}, LayerCAM~\citep{jiang2021layercam}, and integrated gradients~\citep{sundararajan2017axiomatic}. They have been implemented in conjunction with concept-based approaches such as Testing with Concept Activation Vectors (TCAV)~\citep{kim2018interpretability} and attention-based~\citep{vaswani2017attention} mechanisms. Other critical methodologies include LIME (Local Interpretable Model-agnostic Explanations)~\citep{ribeiro2016should},  and counterfactual explanations leveraging generative models like GANterfactual~\citep{mertes2022ganterfactual} and diffusion autoencoders~\citep{atad2024counterfactual}. 
SHAP has been recognized as the preeminent framework for medical imaging, primarily due to its ability to generate pixel-level, human-readable visualizations that facilitate the comprehension of "black-box" models~\citep{rahman2025enhanced, muhammad2024unveiling, ren2025here}. To address the specific complexity of deep neural networks, DeepSHAP~\citep{lundberg2017unified} was developed to combine DeepLIFT~\citep{ponce2024practical} with the Shapley value framework. This adaptation facilitates the efficient back-propagation of feature importance through the numerous layers of a CNN, thereby enabling clinicians to validate precisely which anatomical features or image regions are driving a diagnostic prediction~\citep{de2023explainable}. Specialized SHAP variants have emerged to handle sophisticated visual tasks. Generalized DeepSHAP (G-DeepSHAP)~\citep{chen2022explaining} extends the capabilities of heterogeneous model series frequently encountered in intricate clinical pipelines. Contemporary innovations, such as SHAP-Attention~\citep{wu2025shapattention}, employ dynamic integration of attention mechanisms to weight visual feature importance in an adaptive manner, offering a more comprehensive understanding of modern vision transformers. Furthermore, multi-modal SHAP (MM-SHAP)~\citep{parcalabescu2023mm} addresses the fusion of visual and textual data, quantifying how image features interact with clinical reports to produce a diagnosis. These variants ensure that even the most complex, multi-input imaging models remain transparent and compliant~\citep{muhammad2024unveiling}. 
SHAP values have important limitations in medical image analysis that can undermine clinical reliability, particularly because they are sensitive to model choice and feature correlations~\citep{hossain2025explainable}.
One of the key reasons is that standard SHAP formulations effectively assume independent features, while medical images have correlations~\citep{ponce2024practical, salih2025perspective}.
In correlated settings, SHAP's use of marginal distributions i.e., treating each pixel or region as if it varied independently, can generate unrealistic feature combinations that may not occur in real clinical data, leading to misleading attributions.
This is particularly problematic for medical decision-making, where clinicians may over-trust visually appealing heatmaps that do not faithfully reflect the model's true reasoning~\citep{muhammad2024unveiling, ponce2024practical}. This could be attributed to the concept of automation bias in clinical decision-making.
Therefore, it is crucial to quantify the uncertainty of SHAP explanations so users can see when attributions are unstable or poorly supported by the data and model~\citep{hossain2025explainable}.
Without such uncertainty estimates, SHAP visualizations can hide both aleatoric uncertainty (noise in images or labels) and epistemic uncertainty (lack of similar training examples)~\citep{dubey2025ubiqtree}. For example, the aleatoric uncertainty may be attributed to the medical images from devices with different specifications~\citep{nguyen2025aleatoric}. The epistemic uncertainty~\citep{lohr2024towards} could be attributed to the lack of data, which led to the overlooking of the visualizations that were present in most of the images. For example, the structure of the brain for Alzheimer's analysis was overlooked instead of looking at the ventricles~\citep{shi2015studying}.\\
\\
Existing attempts to quantify uncertainty typically rely on approaches such as resampling, ensembles, or Bayesian / dropout-style approximations around SHAP, but they are computationally expensive for high‑dimensional images and complex deep models.
These methods also inherit SHAP's underlying assumptions about feature independence or its background distribution, so the reported uncertainty may still not correspond to clinically meaningful importance under realistic feature dependencies. Recent advancements in SHAP uncertainty quantification (UQ) frameworks represents a promising direction, combining robust statistical approaches and novel theoretical foundations to enhance trustworthy AI deployment in medical imaging.
Quantifying uncertainty in SHAP values is increasingly vital to ensure clinicians can distinguish between explanations arising from data noise or distributional shifts. This empowers safer decision-making even when models encounter challenging or novel imaging inputs~\citep{singh2025beyond, huang2024review}. Contemporary methods such as variance decomposition, ensemble-based algorithms like UbiQTree~\citep{dubey2025ubiqtree}, and Monte Carlo dropout deepen uncertainty~\citep{faghani2023quantifying} insights for both tree and deep learning models~\citep{lambert2024trustworthy}. Importantly, the integration of evidence theory and Bayesian approaches including Dirichlet process hypothesis sampling, now enables a principled separation of aleatoric and epistemic uncertainties, allowing for more calibrated and reliable explanatory confidence measures~\citep{lambert2024trustworthy}.  The Mean Uncertainty in Explanations (MUE)~\citep{shen2025probabilistic} and distributional SHAP frameworks~\citep{chiaburu2025uncertainty} provide initial steps toward modeling uncertainty in feature importance, but are either limited to point estimates or rely on distributional ranges that can be difficult to verify and may yield wide intervals in practice. Information Theoretic Shapley Values~\citep{watson2023explaining} have been introduced that adapts the Shapley framework so that each feature's contribution is measured not to the prediction itself, but to the conditional entropy (information-theoretic uncertainty) of the model output. It defines new Shapley games whose value functions are based on information measures, and provides inference procedures with finite-sample error guarantees, enabling model‑agnostic explanations of different types of predictive uncertainty useful for tasks like covariate shift detection and active learning. UbiQTree represents a rigorous advance by leveraging Dempster–Shafer evidence theory~\citep{bloch1996some, shafer1992dempster} and Dirichlet-based hypothesis sampling~\citep{follmer2006dirichlet} to decompose SHAP variance into aleatoric, epistemic, and entanglement components, but it is currently tailored to tree ensembles and tabular settings~\citep{dubey2025ubiqtree}. Utilization of Dempster–Shafer theory and related evidence-theoretic approaches is critical because they enable a principled partition of uncertainty sources in SHAP explanations, supporting more reliable feature-importance assessment in high-stakes domains such as healthcare. A research gap remains for deep learning–based medical imaging, where integrating Dempster–Shafer theory, broader evidence theory, and Dirichlet-style sampling with visually meaningful explanations could extend these uncertainty decompositions to CNNs and Vision Transformers, improving domain-expert trust in vision-model outputs.\\
\\
Our research introduces UbiQVision, a framework that transforms medical diagnostics from a deterministic classification task into a transparent evidential reasoning process. Our work bridges the gap between high-performance deep learning, XAI, and the reliability of explanations by unifying three mathematical disciplines. 
\begin{enumerate}
    \item We constructed a heterogeneous ensemble and used Bayesian meta-learning (Dirichlet sampling) to dynamically assign reliability weights to each model based on its validation performance. 
    \item We use SHAP to extract pixel-wise feature attributions, which our algorithms then transform into basic probability assignments. 
    \item Finally, our framework uses Dempster–Shafer theory (DST)~\citep{bloch1996some, shafer1992dempster} to fuse these inputs, producing granular outputs: belief (confirmed), plausibility (potential risk), and uncertainty (total ignorance). This enables clinicians to distinguish between confirmed disease, conflicting model opinions, and insufficient data, thereby mitigating the risks of model overconfidence \& XAI reliability. The project is available at \url{https://github.com/xxx}
\end{enumerate}

\section{Related Work}
Uncertainty-aware XAI is essential for safely deploying deep learning in healthcare, particularly in medical imaging. Clinicians need to understand why a model makes a decision and how stable the explanation is under realistic sources of variability. Neglecting uncertainty in XAI can create a false sense of reliability in tasks such as cancer detection, triage, and treatment planning~\citep{faghani2023quantifying, loftus2022uncertainty}. Further, in the field of medical imaging, uncertainty estimation plays an important role. Previous studies on uncertainty estimation have shown that aleatoric and epistemic uncertainty quantification can enhance uncertainty-awareness and provide safety and calibration in high-stakes diagnostic models. However, these methods focus on predictive uncertainty rather than explanatory uncertainty~\citep{zou2023review}.\\
In medical image analysis, uncertainty can arise from noisy acquisitions, heterogeneous scanners, variable annotations, and model misspecifications. These factors affect both predictions and their explanations~\citep{huang2024review, lambert2024trustworthy}. 
In the context of predictive uncertainty, recent research has proposed multi-domain weakly decoupled domain-generalization networks~\citep{sun2025multi} and open-set classification methods via latent representation prompts~\citep{sun2025open} for the study of unknown fault recognition in diagnosis. The domain-generalization approach targets unknown operating conditions and focuses on learning features that remain robust across domains. This motivates testing explanation uncertainty under domain shift rather than only within closed, in-distribution settings. The open-set recognition method addresses the challenge of unknown classes by combining latent representations with open-set detection, which aligns closely with the challenge of handling samples that lie outside the training distribution.
Therefore, uncertainty quantification (UQ) in explainable artificial intelligence (XAI) aims to characterize how explanations change when inputs, model parameters, or training data are perturbed. UQ provides a bridge between model internals and radiologists' or clinicians' confidence in decisions~\citep{huang2024review, lambert2024trustworthy}.
A useful conceptual view models an explanation as a function $e_\theta(x,f)$, where for a given deep model $f$, an image $x$ (for example, CT, MRI, or fundus photograph), and explanation parameters $\theta$, the output assigns a contribution score to each feature or region. According to this perspective, the UQ methodically examines the fluctuations in $e_\theta(x,f)$ across a range of plausible models and inputs. This approach transforms saliency maps or attribution scores into random variables, thereby deviating from the conventional fixed point estimates~\citep{nazir2023survey, salvi2025explainability}.\\
\\
Empirical approaches generally employ Monte Carlo (MC) procedures, including test-time augmentation, MC dropout~\citep{milanes2021monte, zeevi2024monte, gal2016dropout}, and deep ensembles~\citep{lakshminarayanan2017simple}, to generate multiple predictions and corresponding explanations for a given medical image. Through an analysis of the variability in attribution maps or feature scores across these iterations, researchers can identify regions where explanations are unstable. These regions frequently coincide with image artifacts, low signal-to-noise areas, or underrepresented patterns in the training set~\citep{huang2024review, faghani2023quantifying}.
Analytical techniques, on the other hand, examine how minor variations in image intensities or model parameters influence the explanation, resulting in covariance matrices that quantify the localization or diffusion of uncertainty across the image. Metrics such as a normalized trace of this covariance (analogous to a "mean uncertainty in explanations") provide a single number that can be compared across architectures, training regimes, or XAI methods in radiology and other imaging specialties~\citep{lambert2024trustworthy, huang2024review}.\\
\\
The use of SHAP and related Shapley-based methods in healthcare has experienced a notable increase, with these methods being employed to attribute importance to image regions, radiomic features, or hybrid image–clinical variables. However, conventional SHAP operates under the assumption of a well-specified input distribution, a condition that is frequently unavailable in multi-center imaging datasets. In the event that the underlying data distribution is misspecified or estimated from small, biased samples, SHAP values can become unstable, leading to misleading rankings of imaging biomarkers or spurious emphasis on artifacts~\citep{li2020efficient, tang2021data, nazir2023survey}.
Recent studies have initiated the treatment of SHAP values as distribution-dependent objects. These studies compute SHAP values over ranges of plausible input distributions and derive confidence intervals or credible regions for feature importance. These interval-valued explanations elucidate the sensitivity of feature rankings to dataset shifts, scanner effects, and label noise. They frequently demonstrate that apparently dominant image features are not robust when uncertainty about the data distribution is acknowledged. In the context of high-dimensional imaging, the efficacy of stochastic SHAP schemes, equipped with inherent uncertainty estimations, has been investigated. These schemes have been demonstrated to enhance the computational efficiency of such analyses, while concurrently preserving the variability of attributions across resampled images or submodels~\citep{seoni2023application, huang2024review}.\\
\\
Information-theoretic approaches are gaining prominence in imaging because they naturally quantify both predictive confidence and explanation reliability through quantities such as entropy, mutual information, and information gain. In the domain of medical image synthesis and segmentation, the decomposition of predictive uncertainty into aleatoric and epistemic components can serve as a high-level explanation, facilitating the discernment of whether ambiguous boundaries or absent structures in a scan are attributable to intrinsic imaging limitations or model ignorance~\citep{barbano2022uncertainty, sudre2022uncertainty}.
A number of imaging studies~\citep{gillmann2021uncertainty} have integrated information theory with attribution methods. For instance, they have done so by weighting saliency or SHAP-like scores with measures of information content. The purpose of this integration is to highlight regions that both drive the prediction and carry high uncertainty. These combinations support uncertainty-aware visualization, in which radiologists observe heatmaps accompanied by confidence cues or uncertainty contours. Consequently, radiologists can more accurately determine when to trust model-highlighted lesions or when additional imaging or expert review is necessary.\\
\\
The majority of contemporary uncertainty-aware XAI methods for medical imaging~\citep{chen2024evidence, huang2025deep} continue to be closely associated with sampling-based schemes and probability calibration. These methods primarily reflect variability among estimators rather than addressing underlying structural issues, such as dataset shift, scanner drift, or evolving case-mix. Consequently, pixel- and region-level explanation uncertainty is seldom propagated into the clinical workflows where it is most significant, including longitudinal follow-up, multi-modality fusion (e.g., PET–CT, MRI sequences), and multidisciplinary tumor boards, where explanations must maintain coherence across visits, devices, and imaging protocols. This discrepancy becomes particularly salient in deep learning–based imaging pipelines, CNNs and ViTs where visually compelling heatmaps can obscure the fact that the underlying attributions are highly unstable under realistic changes in data distribution or model configuration. Information theory and Dempster–Shafer Theory (DST)~\citep{bloch1996some, shafer1992dempster} offer methodologies to model known, uncertain, and unknown aspects in explanations. Information-theoretic approaches~\citep{watson2023explaining} view explanations as information bottlenecks aimed at maximizing mutual information while reducing irrelevant variability. DST enhances this by assigning belief masses to hypothesis sets, allowing for distinct representation of genuine ignorance (vacuity) and conflicting evidence (dissonance). This capability enables deep models to differentiate low-SNR data from high-quality lesions and supports the integration of these evidential states into explanation maps. The extension of frameworks such as UbiQTree~\citep{dubey2025ubiqtree} from tabular or tree settings to convolutional and transformer-based encoders, by defining evidential belief over superpixels, feature maps, or latent clusters and combining them across time and modalities, has the potential to yield explanation heatmaps that not only highlight the regions of the network that are focused on but also quantify the extent to which this focus is well-supported, conflicted, or simply ignorant. This, in turn, could make explanatory uncertainty actionable in real clinical decision-making.
\section{Problem Formulation} 
\subsection{Definition \& Assumptions}

Let $\mathcal{D} = \{(x_i, y_i)\}_{i=1}^N$ be a dataset where $x \in \mathbb{R}^{H \times W \times C}$ represents input images of dimensions $height (H) \times width (W) \times channels (C)$ (\textit{e.g.}, Retinal Fundus, Brain MRI, or Pathological Blood Sample) and $y \in \Theta$ represents the class labels.
\label{section:pixel}
\begin{assumption}[Frame of Discernment]
The set of possible classes $\Theta = \{\theta_1, \theta_2, \dots, \theta_K\}$ is mutually exclusive and exhaustive. In the binary case (Healthy vs. Sick), $\Theta = \{Healthy, Sick\}$. The power set $2^\Theta$ includes singleton sets (containing exactly one element), the empty set (containing no elements) $\emptyset$, and the universal set (containing all the elements), $\Omega = \Theta$ (representing total ignorance).
\end{assumption}
In context to our work, the above assumption can be interpreted as following. The power set classifies the patient's condition into a hierarchy of certainty ranging from precise knowledge to complete ignorance. At the most specific level, the singleton sets $\{Healthy\}$ and $\{Sick\}$ represent definite belief, where evidence points exclusively to one condition and rules out the other. In contrast, the universal set, $\{Healthy, Sick\}$, represents total ignorance. While we know the patient is one of these two options, we have no evidence to distinguish between them. Finally, the empty set, represented by the $\emptyset$, represents a logical impossibility or conflict. Because the options are exhaustive, assigning belief to this set implies that the answer does not lie within the list. This usually signals that the data sources contradict one another (e.g., one prediction says "healthy" while another prediction says "sick").
\begin{assumption}[Ensemble Heterogeneity]
We possess an ensemble of $M$ independent models $\mathcal{F}~=~\{f_1, f_2, \dots, f_M\}$. These ML models are heterogeneous (varying architectures or training epochs or any other varying architectures) such that their errors are assumed to be not perfectly correlated. 
\end{assumption}
\subsection{Bayesian Meta-Learning (Model Weighting)}

We note the reliability of each model $f_j$ as a random variable $w_j$. We seek the posterior distribution of these weights based on validation performance.

\subsubsection{The Dirichlet-Multinomial Model}
For the ensemble of models indexed by $\mathcal{M} = \{1, 2, \dots, M\}$, let $\mathbf{c} = [c_1, c_2, \dots, c_M]$ be the vector of correct prediction counts, derived from a validation set of size $V$.
In this framework, $\mathbf{c}$ serves as the empirical measure of reliability. The count $c_M$ represents the evidential support for model $M$. In the context of evidence fusion, this metric dictates the discounting process: Models with high $c_M$ are treated as experts, while models with low $c_M$ are discounted, shifting their belief mass toward the universal set (total ignorance) to prevent unreliable data from skewing the decision.
\begin{definition}[Prior Distribution]
We assume a non-informative uniform prior over the latent model weights $\mathbf{w}$, modeled by a Dirichlet distribution:
\begin{equation}
\mathbf{w} \sim \text{Dir}(\boldsymbol{\alpha}_0)
\end{equation}
where $\boldsymbol{\alpha}_0 = [1, 1, \dots, 1]$ represents the initial concentration parameters (hyperparameters).
\end{definition}
\begin{definition}[Likelihood]
The observed counts $\mathbf{c}$ follow a Multinomial distribution given the weights $\mathbf{w}$:
\begin{equation}
P(\mathbf{c} | \mathbf{w}) \propto \prod_{j=1}^M w_j^{c_j}
\end{equation}
\end{definition}
\begin{theorem}[Posterior Derivation]
Using the conjugacy of the Dirichlet-Multinomial distributions and applying a temperature scaling to control the entropy of the distribution (Tempered Bayesian Inference)~\citep{van2025tempering, zanella2019scalable, kapoor2022uncertainty}, the posterior is derived as:
\begin{equation}
\mathbf{w} | \mathbf{c} \sim \text{Dir}(\boldsymbol{\alpha}_{post})
\end{equation}
where the updated concentration parameters are:
\begin{equation}
\boldsymbol{\alpha}_{post} = \boldsymbol{\alpha}_0 + \frac{\mathbf{c}}{T}
\end{equation}
Here, $T$ is the temperature parameter.
\end{theorem}
\begin{proof}[Proof of Convergence]
This proof establishes the asymptotic behavior of the model weighting. As the validation set size $V \to \infty$, the variance of the Dirichlet distribution $\text{Var}(w_j) \to 0$. The expected weight converges to the model's \textit{relative} performance within the ensemble:
\begin{equation}
    \mathbb{E}[w_j] = \frac{\alpha_{post}^{(j)}}{\sum_{k=1}^M \alpha_{post}^{(k)}} \approx \frac{c_j}{\sum_{k=1}^M c_k}
\end{equation}
Thus, the sampling process inherently favors the robust models while maintaining probabilistic diversity controlled by $T$.
\end{proof}

\subsection{Feature Attribution (SHAP)}
For a specific input $x$, model $f_j$ produces a prediction. We decompose this prediction into pixel-wise contributions using Shapley Values~\citep{lundberg2017unified}.

\begin{definition}[Additive Feature Attribution]
For a feature $i$ (pixel), its SHAP value $\phi_i$ is defined as~\citep{lundberg2017unified}:
\begin{equation}
    \phi_i(k, x) = \sum_{S \subseteq K \setminus \{i\}} \frac{|S|! (|K|-|S|-1)!}{|K|!} \left[ k_x(S \cup \{i\}) - k_x(S) \right]
\end{equation}
Where $K$ is the set of all features and $k_x(S)$ is the model output given feature subset $S$.
\end{definition}

\begin{remark}
The SHAP values satisfy the efficiency property: $\sum_{i} \phi_i(k, x) = k(x) - \mathbb{E}[k(x)]$.
\begin{itemize}
    \item If $\phi_i > 0$: Pixel $i$ contributes evidence \textit{FOR} the target class.
    \item If $\phi_i < 0$: Pixel $i$ contributes evidence \textit{AGAINST} the target class.
\end{itemize}
\end{remark}

\subsection{Evidence Transformation}
This step converts real-valued SHAP scores $\phi \in \mathbb{R}$ into Evidential Mass $m \in [0, 1]$.

\begin{definition}[BPA Mapping Function]
We define a basic probability assignment (BPA) function $m_{j,i}: 2^\Theta \to [0, 1]$ for each pixel $i$ and model $j$. Let $\theta$ be the target hypothesis. Let $\lambda$ be a sensitivity scalar and $w_j$ be the derived Bayesian weight. The mass assignment is:
\begin{align}
    m_{j,i}(\{ \theta \}) &= w_j \cdot \max(0, \tanh(\lambda \cdot \phi_{j,i})) \\
    m_{j,i}(\{ \neg \theta \}) &= w_j \cdot \max(0, -\tanh(\lambda \cdot \phi_{j,i})) \\
    m_{j,i}(\Omega) &= 1 - \left( m_{j,i}(\{ \theta \}) + m_{j,i}(\{ \neg \theta \}) \right)
\end{align}
\end{definition}

\begin{proposition}[Validity of Mass Function]
The mapping constitutes a valid Dempster-Shafer mass function.
\end{proposition}

\begin{proof}
For $m$ to be valid, the sum of masses over the power set must equal 1.
\begin{align}
    \sum_{A \subseteq \Theta} m(A) &= m(\{ \theta \}) + m(\{ \neg \theta \}) + m(\Omega) \\
    &= m(\{ \theta \}) + m(\{ \neg \theta \}) + \left[ 1 - (m(\{ \theta \}) + m(\{ \neg \theta \})) \right] \\
    &= 1
\end{align}
The condition is satisfied by construction.
\end{proof}

\subsection{Applying Dempster-Shafer for Fusion}
We fuse the evidence from the ensemble $\mathcal{F} = \{f_1, \dots, f_M\}$. Let $m_1$ and $m_2$ be mass functions from two models.

\begin{definition}[Dempster's Rule of Combination]~\citep{sentz2002combination}
The orthogonal sum $m_{1,2} = m_1 \oplus m_2$ is defined for any non-empty set $A \subseteq \Theta$ as:
\begin{equation}
    m_{1,2}(A) = \frac{1}{1 - K} \sum_{B \cap C = A} m_1(B) \cdot m_2(C)
\end{equation}
where $K$ is the measure of conflict:
\begin{equation}
    K = \sum_{B \cap C = \emptyset} m_1(B) \cdot m_2(C)
\end{equation}
\end{definition}
The term $(1-K)^{-1}$ acts as a normalization factor that redistributes the conflicting mass to the remaining plausible sets.

\subsection{Epistemic Decision Metrics}
Finally, we extract three metrics for visualization and decision making.

\begin{definition}[Belief]
The lower probability bound, representing total evidence specifically supporting $A$.
\begin{equation}
    \text{Bel}(A) = \sum_{B \subseteq A} m(B) = m(\{A\}) \quad \text{(for singleton sets)}
\end{equation}
\end{definition}

\begin{definition}[Plausibility]
The upper probability bound, representing total evidence consistent with $A$.
\begin{equation}
    \text{Pl}(A) = \sum_{B \cap A \neq \emptyset} m(B)
\end{equation}
\end{definition}

\begin{definition}[Uncertainty]
The epistemic gap representing ignorance.
\begin{equation}
    U(A) = \text{Pl}(A) - \text{Bel}(A)
\end{equation}
\end{definition}

\begin{theorem}[Plausibility-Belief Duality]
For any proposition $A$, the plausibility of $A$ is the complement of the belief in its negation:
\begin{equation}
    \text{Pl}(A) = 1 - \text{Bel}(A^c)
\end{equation}
\end{theorem}

\begin{proof}
By definition, the sum of all masses is 1: $\sum_{B \subseteq \Theta} m(B) = 1$.
We partition the set of all subsets $B$ into two groups: those that intersect with $A$ and those that are disjoint from $A$ (i.e., subsets of $A^c$).
\begin{equation}
    1 = \sum_{B \cap A \neq \emptyset} m(B) + \sum_{B \subseteq A^c} m(B)
\end{equation}
Substituting the definitions of Plausibility and Belief:
\begin{equation}
    1 = \text{Pl}(A) + \text{Bel}(A^c)
\end{equation}
Rearranging the terms yields:
\begin{equation}
    \text{Pl}(A) = 1 - \text{Bel}(A^c)
\end{equation}
\end{proof}
\section{Methodology}
The proposed framework presents a pipeline for reliable medical diagnostics by unifying Bayesian meta-learning, cooperative game theory (SHAP), and Dempster-Shafer theory (DST). Contrary to conventional deterministic deep learning methodologies, which offer point estimates of probability (softmax), the framework generates a comprehensive epistemic uncertainty profile and interval estimates for each prediction.

The workflow is comprised of four distinct phases: The following four topics will be discussed: (1) heterogeneous ensemble construction, (2) Bayesian model reliability estimation, (3) game-theoretic evidence extraction, and (4) evidential fusion and uncertainty quantification.

\subsection{Heterogeneous Ensemble Construction}
To capture a variety of feature representations, a heterogeneous ensemble is constructed, denoted by $\mathcal{F} = \{f_1, f_2, \dots, f_M\}$. This ensemble comprises neural networks with disparate architectures (e.g., ResNets, Vision Transformers) or varying training capacities. This structural diversity guarantees that error manifolds are not perfectly correlated, a prerequisite for effective evidential fusion.

\subsection{Bayesian Meta-Learning for Model Weighting}
Traditional ensembles often average predictions uniformly, ignoring that some models may be less robust than others. To address this, we implement a stochastic weighting mechanism based on validation performance (Refer: Algorithm~\ref{alg:bayesian_weights}).

We model the reliability of the $M$ models as a random variable $\mathbf{w}$ which is modeled with a Dirichlet process. We assume a non-informative uniform prior $\text{Dir}(\boldsymbol{\alpha}_0)$, where $\boldsymbol{\alpha}_0 = [1, \dots, 1]$. By treating the validation process as a multinomial experiment, we update the concentration parameters based on the vector of correct prediction counts $\mathbf{c}$. The posterior distribution is derived as:
\begin{equation}
    \mathbf{w} \sim \text{Dir}\left(\boldsymbol{\alpha}_0 + \frac{\mathbf{c}}{T}\right)
    \label{eq:bayesian_model_weighing}
\end{equation}
where $T$ is a temperature parameter controlling the entropy of the weight distribution. By sampling the final weights $\mathbf{w}$ from this posterior, the framework automatically penalizes underperforming models (lowering their voting power) and rewards robust models, providing a mathematically grounded metric.

\subsection{Game-Theoretic Feature Attribution}
For a given input image $x$, we employ SHAP (SHapley Additive exPlanations) to decompose the model's output into pixel-wise contributions. Using the \texttt{DeepExplainer} approximation, we compute the Shapley value $\phi_{ij}$ for each pixel $i$ and model $j$. Crucially, we preserve the sign of the SHAP values:
\begin{itemize}
    \item Positive $\phi$: Indicates the pixel contributes evidence \textit{supporting} the target hypothesis $\theta$.
    \item Negative $\phi$: Indicates the pixel contributes evidence \textit{refuting} the target hypothesis (supporting $\neg \theta$).
\end{itemize}

\subsection{Construction of Evidential Mass Functions}
A core contribution of this framework is the mapping of real-valued SHAP scores into the Dempster-Shafer mass domain (See:~Algorithm~\ref{alg:mass_construction}). We define a Basic Probability Assignment (BPA) $m_j$ for each model. To bound the unbounded SHAP values $\phi$ into the unit interval $[0,1]$, we utilize a hyperbolic tangent transformation scaled by a sensitivity factor $\lambda$. The mass assignment follows a bipolar logic:
\begin{align}
    m_j(\{ \theta \}) &= w_j \cdot \max(0, \tanh(\lambda \phi)) \\
    m_j(\{ \neg \theta \}) &= w_j \cdot \max(0, -\tanh(\lambda \phi))
    \label{eq:lambda}
\end{align}
The remaining unassigned mass is attributed to the universal set $\Omega$, representing \textit{Ignorance}:
\begin{equation}
    m_j(\Omega) = 1 - (m_j(\{ \theta \}) + m_j(\{ \neg \theta \}))
\end{equation}
This transformation ensures that pixels carrying no signal result in high ignorance rather than false confidence.

\subsection{Dempster-Shafer Fusion and Conflict Resolution}
To synthesize the opinions of the ensemble, we employ Dempster's Rule of Combination, implemented in Algorithm:~\ref{alg:ds_fusion}. This rule computes the orthogonal sum of the mass functions:
\begin{equation}
    m_{1 \oplus 2}(A) = \frac{1}{1-K} \sum_{B \cap C = A} m_1(B) \cdot m_2(C)
\end{equation}
The normalization factor $(1-K)^{-1}$ plays a critical role in robustness. The conflict constant $K$ measures the degree of disagreement between models. By normalizing against $K$, the framework mathematically discounts evidence where models strongly disagree, effectively filtering out noise and hallucinations from weaker models.

\subsection{Clinical Implications and Interpretability}
The resulting framework provides a set of three epistemic metrics. The utilization of these metrics enables a direct mapping from mathematical states to clinical interpretations, thereby establishing a robust framework for ensuring the reliability.

\subsubsection{Belief Map ($\text{Bel}$)}
Defined as the lower bound of probability, $\text{Bel}(A) = m(\{A\})$. This acts as a confirmation signal
\begin{itemize}
    \item Visual Consequence: Appears as dark green regions.
    \item Clinical Implication: These regions represent high-probability regions contributing towards the detected class. The ensemble has reached a consensus that specific anatomical features (\textit{e.g.}, lesions, enlarged ventricles) are present. This provides the \textit{localization} of the disease.
\end{itemize}

\subsubsection{Plausibility Map ($\text{Pl}$)}
Defined as the upper bound of probability, $\text{Pl}(A) = 1 - \text{Bel}(\neg A)$. This is identifies the conflicts.
\begin{itemize}
    \item Visual Consequence: Appears as blue regions. Lighter spots (low plausibility) indicate areas where at least one model found evidence \textit{against} the diagnosis.
    \item Clinical Implication: When the Belief is elevated yet Plausibility is diminished, this signifies an internal model conflict (\textit{e.g.}, one model detects a healthy texture while another detects an abnormal region). This alerts the user of possible adversarial phenomena.
\end{itemize}

\subsubsection{Uncertainty Map ($U$)}
Defined as the epistemic gap, $U = \text{Pl} - \text{Bel}$.
\begin{itemize}
    \item Visual Consequence: Appears as bright yellow (High $U$) or dark purple (Low $U$).
    \item Clinical Implication: This quantifies \textit{Total Ignorance}.
    \begin{itemize}
        \item \textit{Yellow Regions (Background):} The model correctly identifies that it has no knowledge about empty space.
        \item \textit{Purple Regions (Anatomy):} The model confirms it has sufficient information to make a decision.
    \end{itemize}
    \item If the ROI (Region of Interest) is yellow, the system is signaling an Out-of-Distribution (OOD) sample or acquisition failure. The AI is effectively supporting prompting the human intervention.
\end{itemize}
\subsection{The Temperature Parameter (\textit{T}) and the Sensitivity Parameter ($\lambda$)}
\begin{figure}
    \centering
    \includegraphics[width=1.0\linewidth]{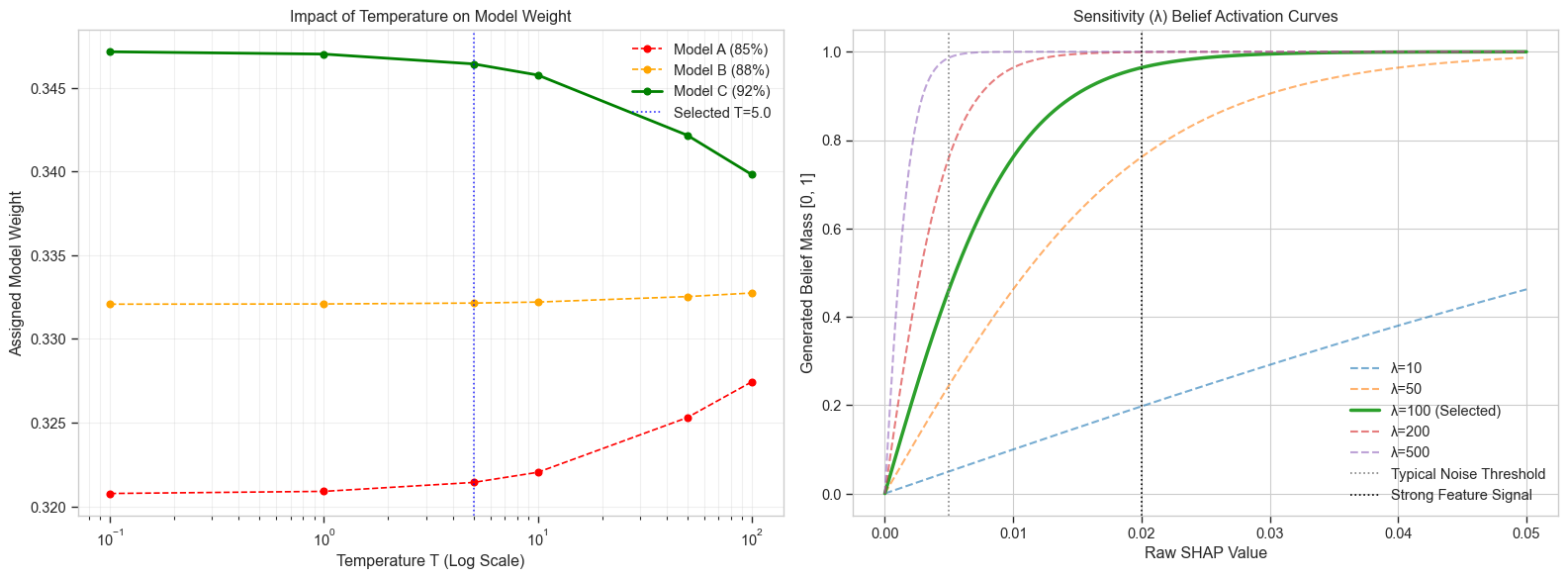}
    \caption{On the left, the plot demonstrates the impact of the temperature parameter (\textit{T}) on the model weights for Models A, B, and C, as evaluated on the test dataset using a specific metric on a logarithmic scale. A low \textit{T} value means that a model with even a slightly higher F1 score receives the highest weight, while a high \textit{T} value means that the models will be allotted equal weight, irrespective of their performance. On the right, the plot shows the impact of the sensitivity parameter $\lambda$ on the generated belief mass function. This parameter controls the signal-to-noise ratio in the uncertainty maps. When $\lambda$ is low, the resultant maps will assume that everything is noise. With a high $\lambda$, the very low background noise would be counted as maximum belief, resulting in a binary mask with no useful uncertainty
    maps.}
    \label{fig:lambda_T_analysis}
\end{figure}

The equation:~\ref{eq:bayesian_model_weighing}, which weighs the model using Bayesian or in-specific Dirichlet processes, is controlled by a temperature parameter \textit{T}, denoted by \textit{T}. (Refer figure:~\ref{fig:lambda_T_analysis}). This parameter controls the model's contribution as per the F1 score. For this study, the F1 score was used to weigh the models, but in other studies or extensions of this research, researchers should use specific metrics to weigh models, such as precision, recall, weighted recall, \textit{etc}. The framework supports selection of varied number of evaluation metrics. A low \textit{T} value means that a model with even a slightly higher F1 score receives the highest weight. This would eliminate the benefit of using ensemble models to produce uncertainty maps because it would ignore the unique features learned by heterogeneous models. With a high \textit{T} value, the performance differences among the models would be understated, as it would randomly weight the models involved in the ensemble. The goal is to calibrate and obtain a \textit{T} that balances the models according to their performance, while still taking the weaker models into consideration, as they have learned strong local evidence. At \textit{T}=0.1, Model C (92\%) demonstrates a near 100\% weight capacity. It has been determined that models A and B are no longer applicable. This is a significant concern because if Model C makes an error, the system will fail. When, \textit{T}=100.0, all models align to a consensus of 33\% weight. The high performance of Model C is overlooked. At (\textit{T}$\approx$5.0), Model C's weight is clearly higher than that of Model A, with a ratio of approximately 0.36 to 0.30. However, Model A retains sufficient influence to correct Model C if Model C is uncertain (e.g., if Model C outputs 0.51 vs. 0.49, Model A's strong vote can sway the fusion). We have selected \textit{T}=5.0 because it enables a collaborative environment where models are weighted based on their performance depending upon the chosen metric, while minority opinions are preserved to ensure the integrity of the results, which is a key objective of DST theory.\\
\\
The equation:~\ref{eq:lambda}, which is used in the SHAP-to-Belief conversion, is controlled by the parameter $\lambda$. This parameter controls the signal-to-noise ratio in the uncertainty maps (see Figure~\ref{fig:lambda_T_analysis}).
This parameter is necessary because the raw SHAP values are very small compared to the absolute value of the pixels in the image. Raw SHAP values may range from $10^{-1}$ to $10^{-4}$. If the value of $\lambda$ is low, then the resultant maps will assume that everything is noise. This will result in valid features having belief mass near zero, leading to 100\% uncertainty (ignorance). With a high $\lambda$, the very low background noise would be counted as maximum belief, resulting in a binary mask with no useful uncertainty maps. The goal is to calibrate the lambda value such that pixels or regions in the image that contribute to the prediction are mapped to the range of the tanh function, while keeping background noise close to insignificant. At $\lambda$=10, a strong SHAP signal (0.02) only yields a belief mass of $~$0.2. The system indicates a high degree of uncertainty, even for features that are clearly identifiable. This approach is considered too restrictive. At $\lambda$=500, it is estimated that a belief mass greater than 0.75 is produced by tiny noise (0.002). The heatmap becomes crowded \& bloated, covering the entire background. For $\lambda\approx$100, the, for SHAP values less than 0.005 (noise), the curve remains relatively flat, which maintains a low level of belief. For SHAP values greater than 0.02 (strong features), the curve approaches saturation (0.9+). It maintains a linear gradient between 0.005 and 0.02, allowing the visualization to show degrees of importance rather than just a binary blob. We have selected $\lambda$=100 (assuming normalized image inputs) because it aligns the tanh activation with the typical magnitude of GradientShap values for deep learning models, thereby maximizing the contrast between the object of interest and the background.

\begin{algorithm}[h]
\caption{Bayesian Meta-Learning for Model Reliability}\label{alg:bayesian_weights}
\begin{algorithmic}[1]
\Require Ensemble of models $\mathcal{F} = \{f_1, \dots, f_M\}$, Validation Dataset $\mathcal{D}_{val}$ of size $V$, Temperature $T$, Prior $\boldsymbol{\alpha}_0 = [1, \dots, 1]$
\Ensure Bayesian weight vector $\mathbf{w} \sim \text{Dir}(\boldsymbol{\alpha}_{post})$

\State \textbf{Initialization:}
\State Set count vector $\mathbf{c} \gets [0, \dots, 0]$

\State \textbf{Evidence Accumulation:}
\For{$i = 1$ to $V$} \Comment{Iterate over validation set}
    \State $(x, y) \gets \mathcal{D}_{val}[i]$
    \For{$j = 1$ to $M$}
        \State $\hat{y} \gets \text{argmax}(f_j(x))$
        \If{$\hat{y} = y$}
            \State $c_j \gets c_j + 1$ \Comment{Update correct prediction counts}
        \EndIf
    \EndFor
\EndFor

\State \textbf{Tempered Posterior Update:}
\State $\boldsymbol{\alpha}_{post} \gets \boldsymbol{\alpha}_0 + \frac{\mathbf{c}}{T}$ \Comment{Scale counts by temperature}

\State \textbf{Weight Sampling:}
\State $\mathbf{w} \sim \text{Dir}(\boldsymbol{\alpha}_{post})$ \Comment{Sample latent model weights}
\State \Return $\mathbf{w}$
\end{algorithmic}
\end{algorithm}

\begin{algorithm}[h]
\caption{Evidential Mass Construction (BPA Mapping)}\label{alg:mass_construction}
\begin{algorithmic}[1]
\Require Input image $x$, Model $f_j$, Target Class $\theta$, Bayesian Weight $w_j$ (from Alg. 1), Sensitivity $\lambda$
\Ensure Mass function $m_{j,i}: 2^\Theta \to [0, 1]$ for pixel $i$

\State \textbf{Feature Attribution:}
\State $\Phi \gets \text{DeepExplainer}(f_j, x)$ \Comment{Compute SHAP values}

\For{each pixel $i$ in $x$}
    \State $\phi_{j,i} \gets \Phi[i]$ \Comment{Pixel contribution}
    
    \State \textbf{Hyperbolic Transformation:}
    \State $\psi \gets \tanh(\lambda \cdot \phi_{j,i})$ 
    
    \State \textbf{Basic Probability Assignment:}
    \State $m_{j,i}(\{ \theta \}) \gets w_j \cdot \max(0, \psi)$ \Comment{Evidence FOR $\theta$}
    \State $m_{j,i}(\{ \neg \theta \}) \gets w_j \cdot \max(0, -\psi)$ \Comment{Evidence AGAINST $\theta$}
    
    \State \textbf{Ignorance Assignment:}
    \State $m_{j,i}(\Omega) \gets 1 - (m_{j,i}(\{ \theta \}) + m_{j,i}(\{ \neg \theta \})) $ \Comment{Total Ignorance}
    
    \State \textit{Note: Normalization is not required as $\sum m = 1$ by construction.}
\EndFor

\State \Return $\{m_{j,i}\}_{\forall i}$
\end{algorithmic}
\end{algorithm}

\begin{algorithm}[h]
\caption{Dempster-Shafer Fusion \& Epistemic Metrics}\label{alg:ds_fusion}
\begin{algorithmic}[1]
\Require Set of Mass functions $\{m_1, \dots, m_M\}$ for pixel $i$, Target Class $\theta$
\Ensure Belief $\text{Bel}$, Plausibility $\text{Pl}$, Uncertainty $U$

\State Initialize fused mass $m_{fused} \gets m_1$

\State \textbf{Sequential Fusion:}
\For{$j = 2$ to $M$}
    \State Let $m_{curr} = m_{fused}$ and $m_{next} = m_j$
    
    \State \textbf{Calculate Conflict}
    \State $K \gets \sum_{A \cap B = \emptyset} m_{curr}(A) \cdot m_{next}(B)$
    \State \Comment{Specifically: $m_{curr}(\{\theta\})m_{next}(\{\neg\theta\}) + m_{curr}(\{\neg\theta\})m_{next}(\{\theta\})$}

    \State \textbf{Apply Dempster's Rule }
    \For{each $C \in \{\{\theta\}, \{\neg \theta\}, \Omega\}$}
        \State $m_{new}(C) \gets \frac{1}{1-K} \sum_{A \cap B = C} m_{curr}(A) \cdot m_{next}(B)$
    \EndFor
    \State $m_{fused} \gets m_{new}$
\EndFor

\State \textbf{Extract Epistemic Metrics:}
\State $\text{Bel}(\theta) \gets m_{fused}(\{ \theta \})$ \Comment{Lower Bound}
\State $\text{Pl}(\theta) \gets 1 - m_{fused}(\{ \neg \theta \})$ \Comment{Upper Bound}
\State $U(\theta) \gets \text{Pl}(\theta) - \text{Bel}(\theta)$ \Comment{Epistemic Gap}

\State \Return $\text{Bel}(\theta), \text{Pl}(\theta), U(\theta)$
\end{algorithmic}
\end{algorithm}
\section{Results}
\subsection{Experimental Setup}
We designed experimentations spanning three distinct medical imaging domains, histology, ophthalmology, and neuro-imaging to rigorously evaluate the framework. These experiments were designed to test the framework's ability to generalize across different noise levels, anatomical structures, and pathological complexities. We used three publicly available medical datasets ranging from balanced binary classification to highly imbalanced multi-class classification. The Malaria Cell Images dataset (Microscopy)~\citep{yang2019deep, kassim2021diagnosing}, sourced from the official NIH repository, is a binary classification benchmark in histopathology. This dataset contains 27,558 cell images with a perfectly balanced distribution, split equally into 13,779 images for the parasitized class and 13,779 images for the uninfected class. The dataset consists of Giemsa-stained thin blood smear slides. Alpha blending was applied during preprocessing to enhance the separation of cellular structures from the background. This tested the model's ability to detect focal parasitic inclusions. For neuroimaging, we used the Alzheimer's MRI dataset~\citep{yakkundi2023alzheimer, marcus2007open, lamontagne2019oasis}, which comprises T1-weighted MRI scans that are categorized into four stages of dementia progression. Unlike the malaria dataset, this collection exhibits significant class imbalance, presenting a realistic challenge for the evidential weighting mechanism. The distribution is heavily skewed toward earlier stages: Non-Demented (3,200 files), Very Mild Demented (2,240 files), Mild Demented (896 files), and Moderate Demented (64 files). The extreme scarcity of the moderate dementia class ($n$ = 64) specifically tests the reliability of the explanations and generating confidence belief maps on data highlighting epistemic uncertainty. Finally, for ophthalmology, we used a subset of the diabetic retinopathy (DR) dataset from Kaggle Competition sponsored California Healthcare Foundation and retinal images provided by the EyePACS~\citep{dugas7diabetic}, to evaluate the framework on ordinal disease grading. This dataset includes retinal fundus photographs labeled into five increasing severity levels: Healthy (525 images), Mild DR (370 images), Moderate DR (599 images), Severe DR (202 images), and Proliferative DR (290 images). To obtain a class-balanced, and computationally manageable cohort, a subset of images was randomly selected from the original Kaggle EyePACS competition dataset, which contains 35,126 labeled fundus images. This dataset challenges the model to distinguish subtle severity transitions rather than simple binary distinctions. This requires the fusion process to handle higher inter-model conflict.

In terms of preprocessing, we resized all images to a standard resolution of 128 × 128 pixels to maintain computational efficiency while preserving essential pathological features. To ensure compatibility with pretrained feature extractors and stable SHAP convergence, all images were normalized using the standard ImageNet mean ($\mu = [0.485, 0.456, 0.406]$) and standard deviation ($\sigma = [0.229, 0.224, 0.225]$). We constructed a heterogeneous ensemble, $\mathcal{F} = \{f_1, f_2, f_3\}$, comprising three distinct neural network architectures. This diversity is critical to ensuring that error manifolds are uncorrelated, which is a prerequisite for effective evidential fusion. Model 1 (Custom CNN~\citep{lecun2002gradient}) is a lightweight convolutional neural network with three convolutional blocks ($16 \to 32 \to 64$ filters) and max pooling. It serves as the expert with low-bias for local texture features. Model 2 (ResNet-18)~\citep{he2016deep} is a deep residual network with skip connections that provides robust feature extraction for global structural patterns, though it is prone to high-frequency noise. Model 3 (Vision Transformer — ViT)~\citep{dosovitskiy2020image} with configuration \textit{ViT\_B\_16\_Weights.IMAGENET1K\_V1}~\citep{dosovitskiy2020image},  is a patch-based attention model. While powerful, its non-local attention mechanism introduces different failure modes, such as blocky artifacts, compared to CNNs. This adds necessary epistemic diversity to the ensemble.

All models were implemented using the PyTorch Lightning (version 2.5.5) framework and default parameters to ensure reproducibility. We used the Adam optimizer with a learning rate of $\eta = 1e^{-3}$ and standard cross-entropy loss for individual model training. To induce further heterogeneity, the models were trained for varying lengths of epochs (e.g., eight, twelve, and sixteen epochs), which prevented the ensemble from converging to the same local minimum and enriched the diversity of the evidential signal. The datasets were partitioned into 80\% training and 20\% testing sets. A randomly sampled subset of the test set ($N=100$) was reserved for the Bayesian validation phase. The reliability estimation module was configured with a non-informative uniform prior, $\boldsymbol{\alpha}_0 = [1, 1, 1]$. We evaluated the ensemble on the reserved validation subset by applying a temperature factor of $T = 5.0$ to the Dirichlet posterior update ($\boldsymbol{\alpha}_{post} = \boldsymbol{\alpha}_0 + c/T$). This hyperparameter controls the entropy of the distribution, ensuring that even weaker models contribute nonzero masses to the fusion process. To improve interpretability and fusion, we used the SHAP Deep Explainer with default parameters with a background dataset of 50 randomly sampled training images to approximate the conditional expectations. The sensitivity parameter $\lambda$ for the hyperbolic tangent conversion ($\phi \to m$) was set to $\lambda=100.0$. This high sensitivity was chosen empirically to amplify weak signals in medical images (e.g., subtle microaneurysms)~\citep{asha2021detection} and convert them into meaningful evidential mass. We experimented this parameter on smaller datasets such as MNIST digits and MNIST fashion as well. We found that the framework performed well even with low $T$ \& $\lambda$ as for these datasets the signals were strong but in case of medical imaging datasets, the $T$ \& $\lambda$ should be higher. The fusion engine used Dempster's Rule of Combination with a dynamic conflict check and extracted epistemic metrics (Belief, Plausibility, and Uncertainty) at the pixel level to generate spatial maps. 
\subsection{Malaria Infection Detection}
\begin{figure}[]
    \centering
    \begin{subfigure}[H]{ \textwidth}
        \centering
        \includegraphics[width= 0.5 \textwidth]{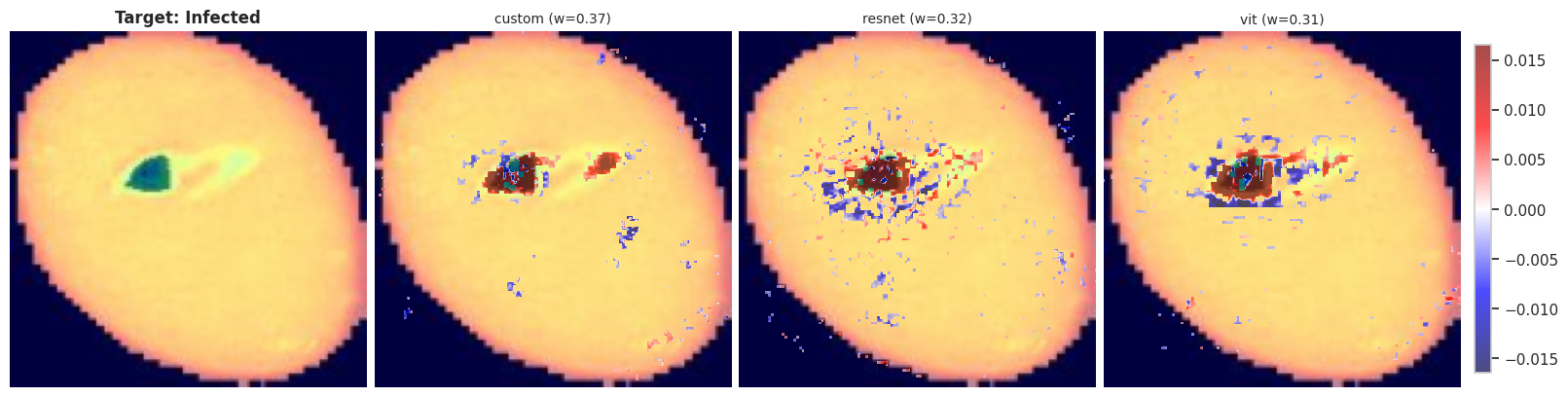}
        \caption{}
    \end{subfigure}
    \hfill 
    \begin{subfigure}[H]{ \textwidth}
        \centering
        \includegraphics[width= 0.5 \textwidth]{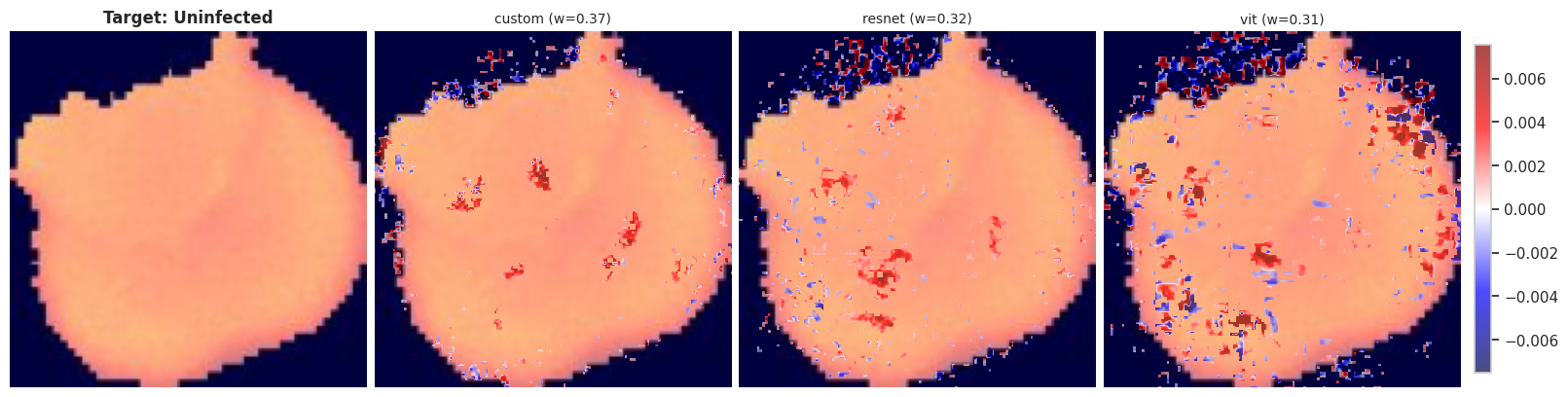}
        \caption{}
    \end{subfigure}
    \caption{The visualization shows the SHAP values for the two classes of the malaria dataset from the predictions of three models: a custom CNN model, a ResNet model, and a ViT model, with weights of 0.37, 0.32, and 0.31, respectively. The "uninfected" class receives positive attribution ($\phi > 0$) from the models for the erythrocyte’s interior regions with low-frequency spatial variation. Specifically, the smooth, homogeneous cytoplasm receives positive attribution. Conversely, the models assign high negative importance ($\phi < 0$) to high-contrast, high-frequency anomalies, such as ring-like chromatin structures.}
    \label{fig:malaria_shap}
\end{figure}

\begin{figure}[]
    \centering
    \begin{subfigure}[b]{ \textwidth}
        \centering
        \includegraphics[width= 0.5 \textwidth]{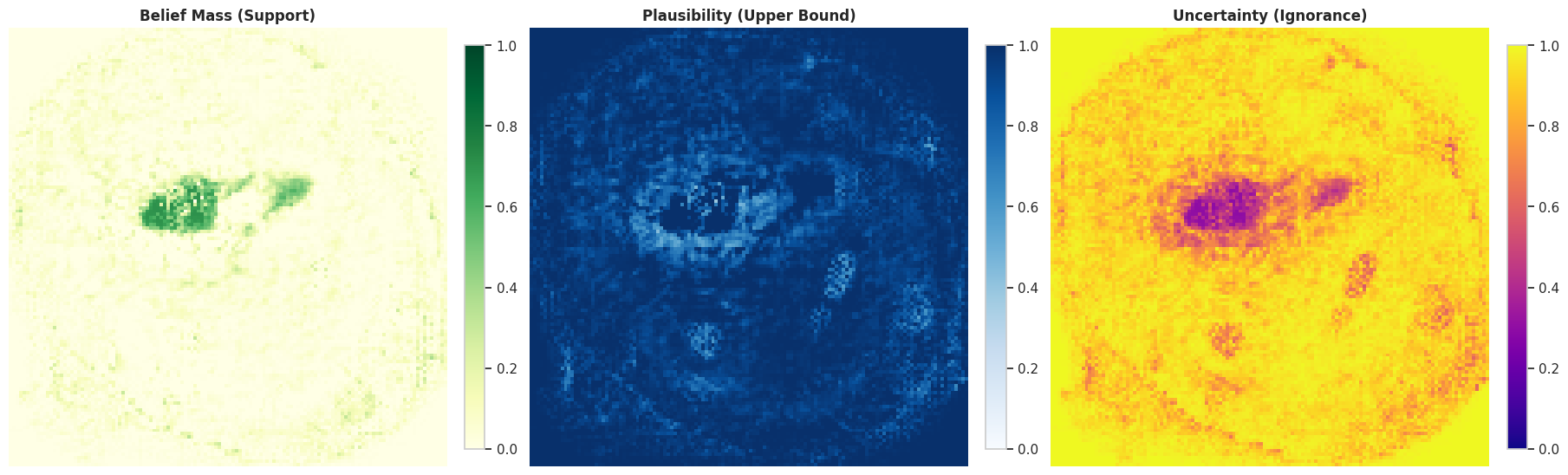}
        \caption{}
    \end{subfigure}
    \hfill 
    \begin{subfigure}[b]{\textwidth}
        \centering
        \includegraphics[width= 0.5 \textwidth]{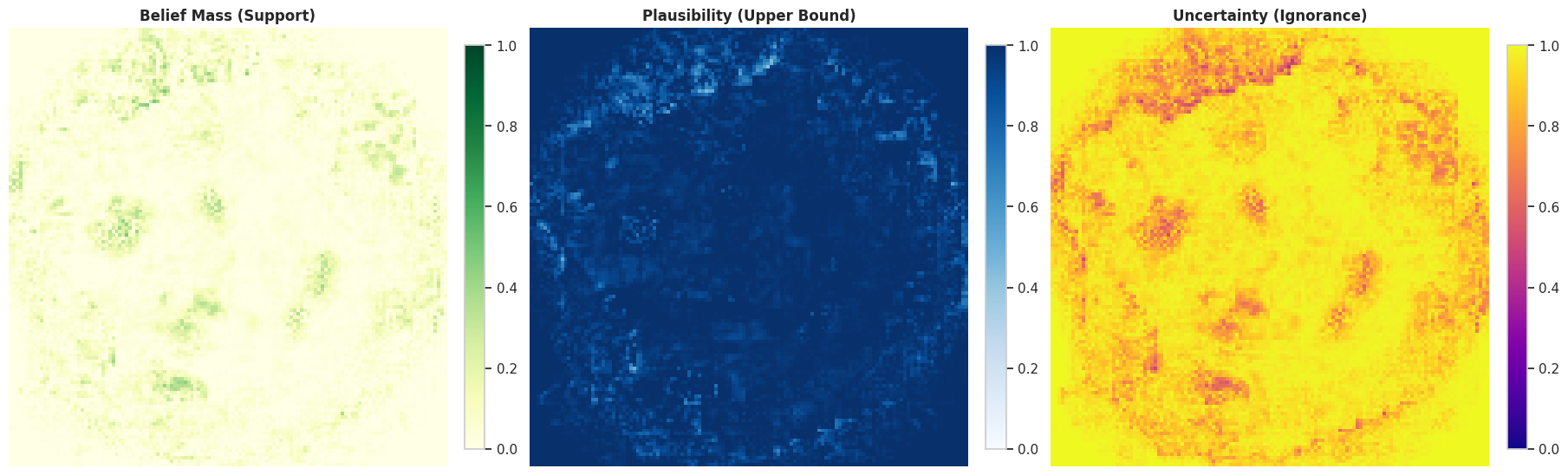}
        \caption{}
    \end{subfigure}
    \caption{This figure shows the fusion of SHAP explanations from the weighted model ensemble (Custom CNN, ResNet, and ViT) for a malaria-infected and uninfected sample. The figure (a) is of the parasitized sample. The belief mass (support) map shows a concentrated area of high belief (dark green), which precisely localizes the parasite. This indicates that the models found strong, consistent evidence at this location. This confidence is reflected in the uncertainty (ignorance) map, where the central purple region represents a significant reduction in uncertainty (approaching 0.0). According to Dempster-Shafer theory, this reduction in uncertainty indicates that the models are highly confident and in agreement about the feature of interest. In contrast, the surrounding cytoplasm (yellow) has high epistemic uncertainty due to a lack of distinguishing features. The figure (b) is of an uninfected sample. This figure illustrates the fusion results for a healthy, uninfected cell, which contrasts sharply with the parasitized sample. The belief mass map appears diffuse and scattered, with no strong, centralized support features. This correctly reflects the absence of a foreign body (parasite). Consequently, the uncertainty (ignorance) map demonstrates high uncertainty (bright yellow) across the entire cell structure. This uniform distribution of ignorance indicates that, although the ensemble likely predicts the class correctly, the decision relies on the general absence of features rather than the detection of a specific object. This results in higher overall quantified uncertainty compared to the infected sample.}
    \label{fig:malaria_ubiqcon}
\end{figure}

\begin{figure}[]
    \centering
    \begin{subfigure}[b]{ \textwidth}
        \centering
        \includegraphics[width= 0.45 \textwidth]{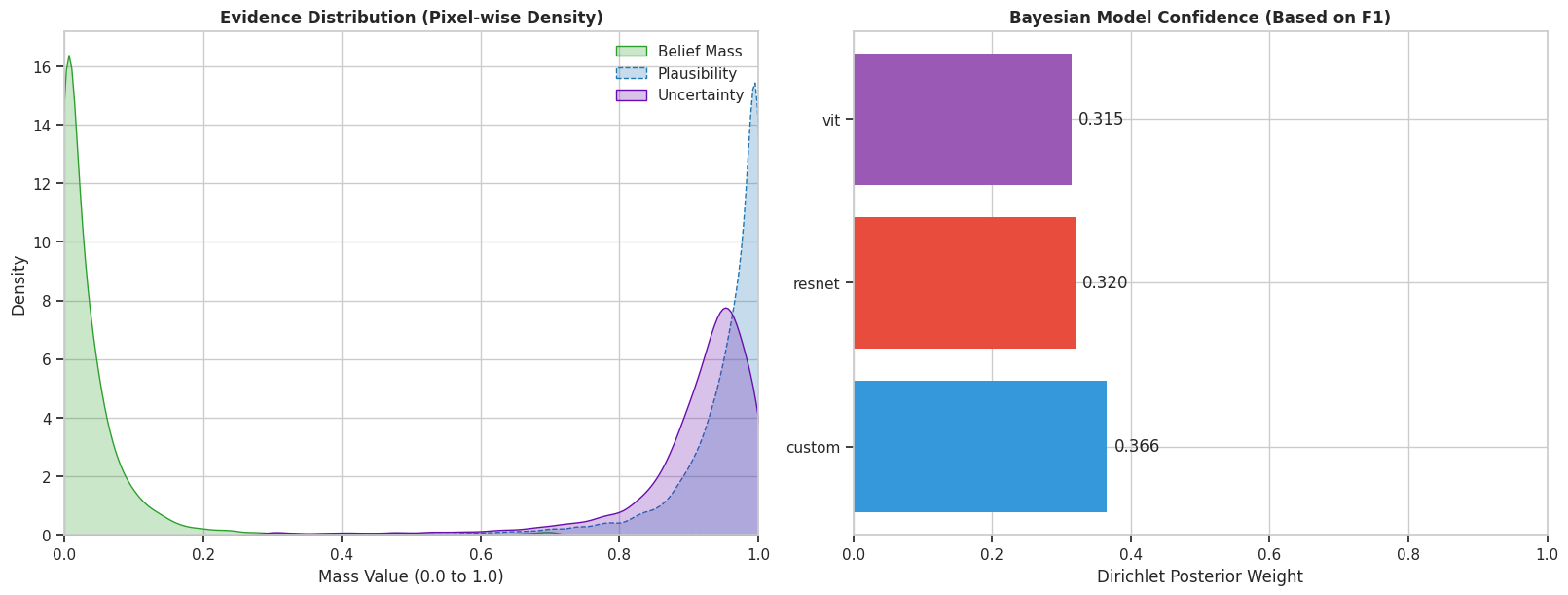}
        \caption{}
    \end{subfigure}
    \hfill 
    \begin{subfigure}[b]{\textwidth}
        \centering
        \includegraphics[width= 0.45 \textwidth]{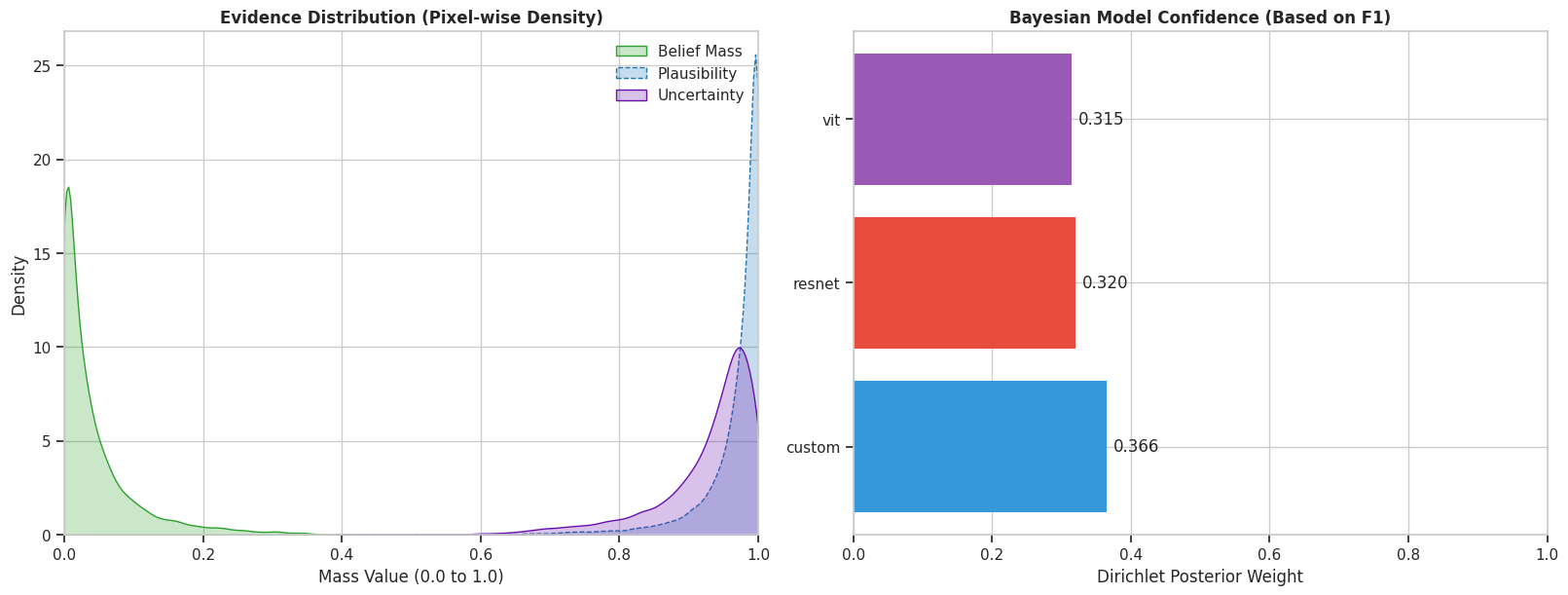}
        \caption{}
    \end{subfigure}
    \caption{Figure (a) shows the distribution of evidence and model confidence for the parasitized sample. It presents the statistical analysis of the fusion process for the infected erythrocyte. The left panel shows the kernel density estimation (KDE) of pixel-wise mass values. The belief mass (green) is heavily skewed toward zero, but it has a noticeable tail that extends into higher values. This statistically confirms the presence of a specific, localized region of strong support (the parasite) amidst a background of low belief. The uncertainty (purple) shows wider variance, reflecting confident detection in the center and high ignorance in the cytoplasm. The right panel shows the Bayesian weights assigned to the ensemble via a Dirichlet distribution based on validation F1 scores. In this case, the custom CNN received the highest posterior weight (w = 0.366), followed by ResNet (w = 0.320) and ViT (w = 0.315). This indicates that the simpler architecture offered slightly higher robustness for this specific split. Figure (b) shows the statistical profile of uninfected cell fusion and clearly contrasts with the parasitized analysis. In the KDE plot (left), the belief mass (green) is compressed near 0.0 with no significant tail. This quantitatively verifies that the models found almost no pixel-wise evidence to support the parasitized class. Consequently, the uncertainty (purple) and plausibility (blue) distributions are tightly clustered near 1.0, signifying high global ignorance. This confirms that the system's decision was driven by a lack of positive evidence rather than the detection of contradictory features. The Bayesian model confidence (right) remains consistent with the previous figure and illustrates the fixed weighting scheme applied across the dataset during the inference phase.}
    \label{fig:malaria_analytics}
\end{figure}

\begin{figure}
    \centering
    \includegraphics[width=0.5\linewidth]{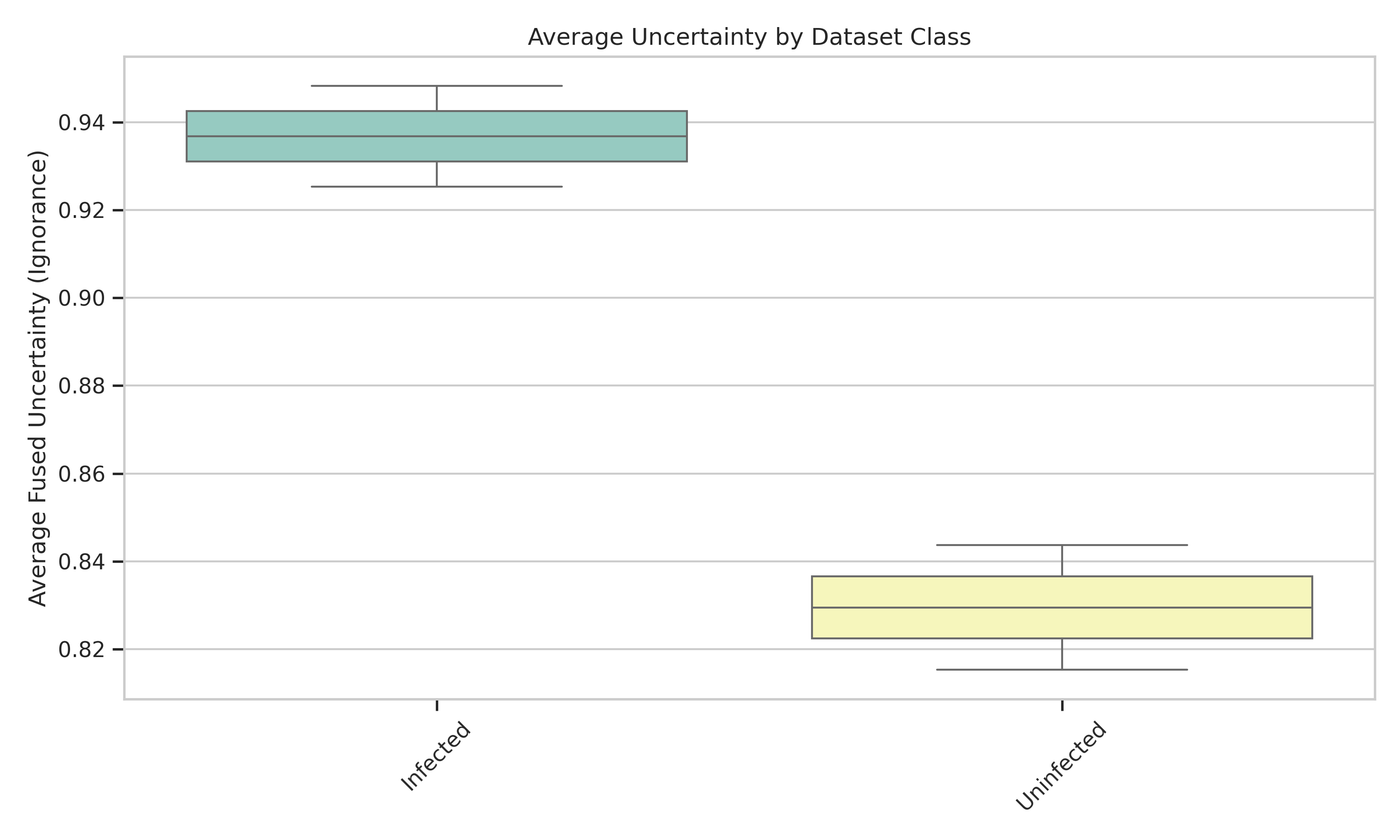}
    \caption{This figure denotes a box plot allowing us to analyze the mean average Uncertainty (Ignorance) for the Malaria dataset. The Infected class has the highest mean fused uncertainty while the Uninfected class has the lowest.}
    \label{fig:malaria_error_per_class}
\end{figure}
\begin{figure}
    \centering
    \includegraphics[width=0.5\linewidth]{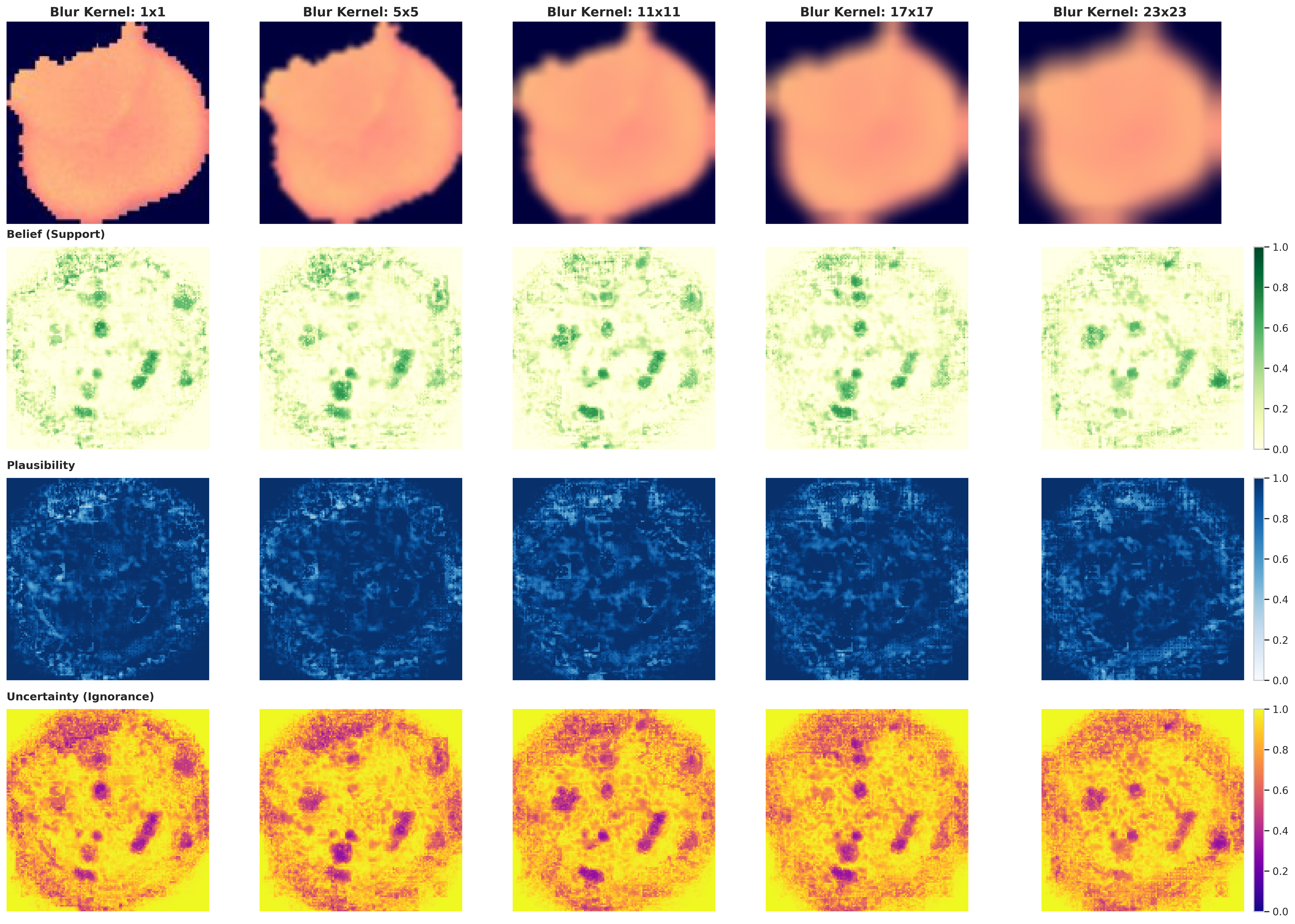}
    \caption{The visual illustration of the framework's  for progressive structural degradation through the introduction of the Gaussian blur. The top row of the figure displays the input image at increasing blur kernel sizes, ranging from 1x1 to 23x23. Subsequent rows illustrate the corresponding Belief (Support), Plausibility, and Uncertainty (Ignorance) maps generated by the DS-SHAP fusion framework for the Malaria dataset. The increase in the Gaussian kernel increases the uncertainty and the boundaries start to fade away.}
    \label{fig:malaria_ablation_study}
\end{figure}
\begin{figure}
    \centering
    \includegraphics[width=0.5\linewidth]{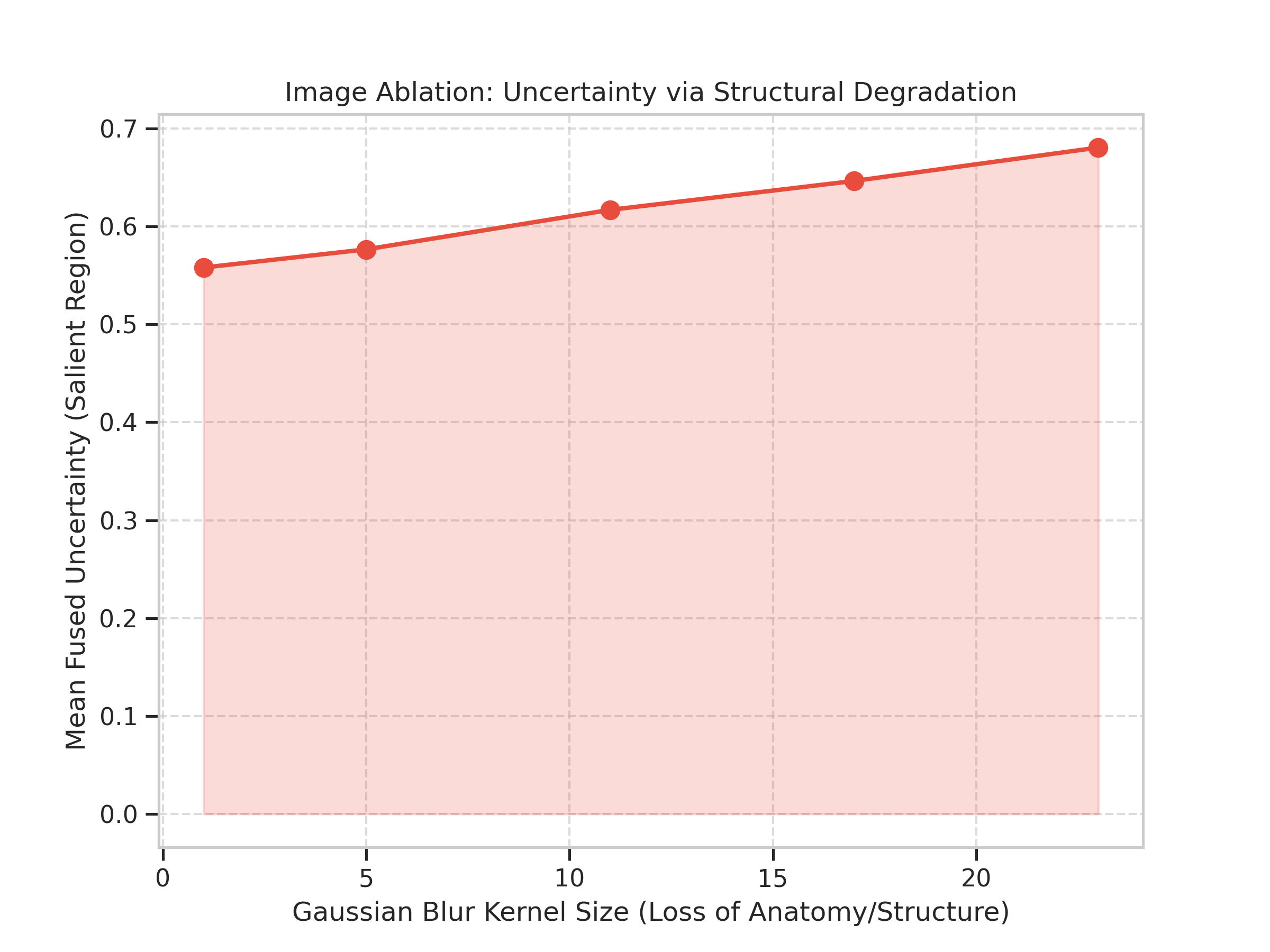}
    \caption{This figure illustrates the impact of structural degradation, manifested through the incorporation of a Gaussian blur kernel, on the mean fused uncertainty. The augmentation of the Gaussian blur kernel has been demonstrated to enhance the mean fused uncertainty for the malaria dataset. The following line chart is intended to quantify the relationship between structural integrity and the framework's calculated epistemic uncertainty. The x-axis of this figure represents the increasing Gaussian blur kernel size, or the loss of anatomical detail, while the y-axis shows the mean fused uncertainty, or the degree of ignorance, measured exclusively within the salient anatomical regions.}
    \label{fig:malaria_ablation_chart}
\end{figure}
\begin{figure}
    \centering
    \includegraphics[width=0.5\linewidth]{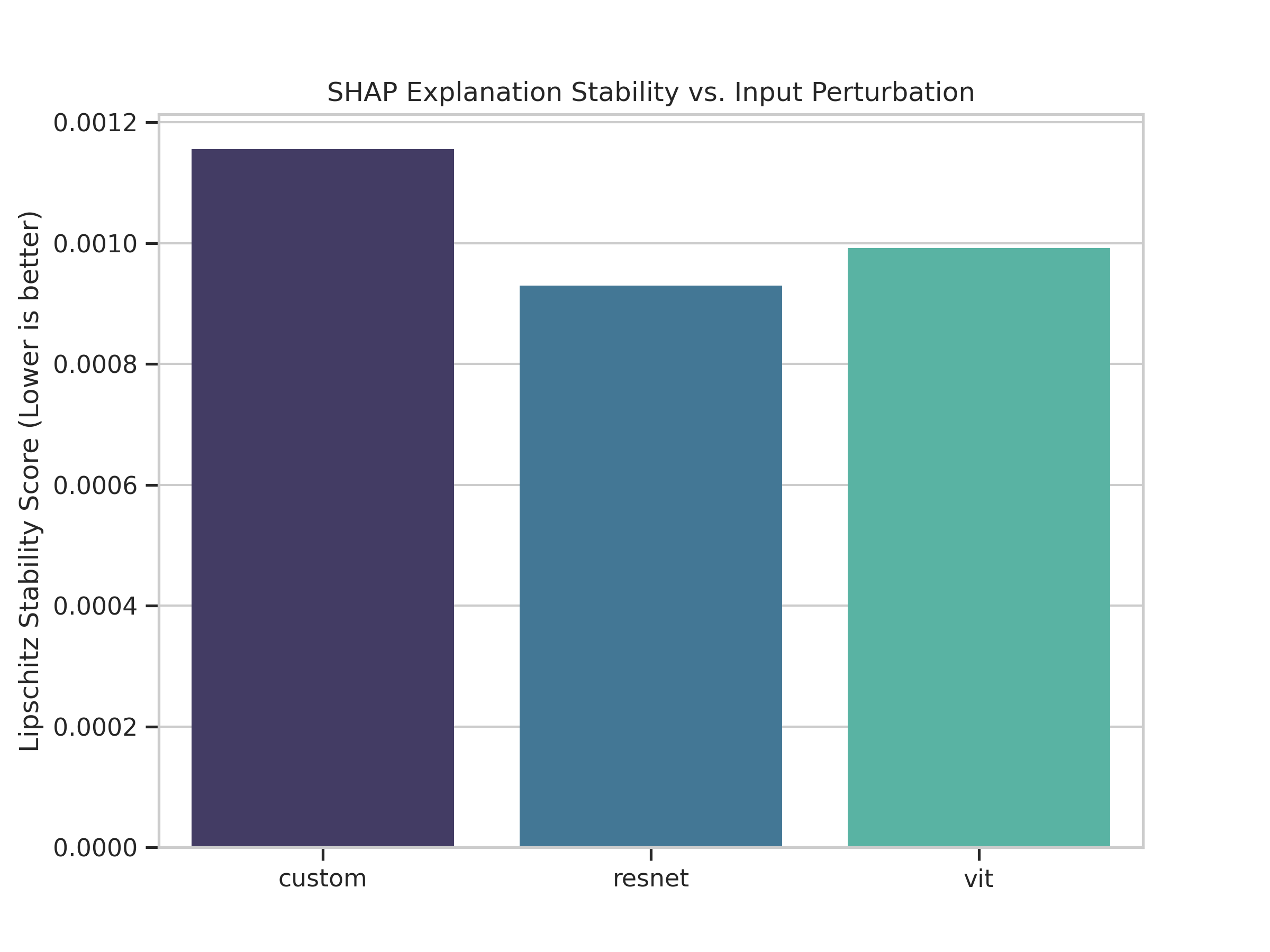}
    \caption{This figure indicates the stability of the SHAP attributions for the architectures employed in the analysis of images from the Malaria dataset. The custom CNN demonstrates the highest Lipschitz score (lower the better), indicating the least stable attributions in comparison to the ViT and ResNet architectures. However, the scores are low, which indicates that the attribution which is used in our framework to quantify uncertainty is overall stable.}
    \label{fig:shap_lipschitz_malaria}
\end{figure}
\begin{figure}
    \centering
    \includegraphics[width=0.5\linewidth]{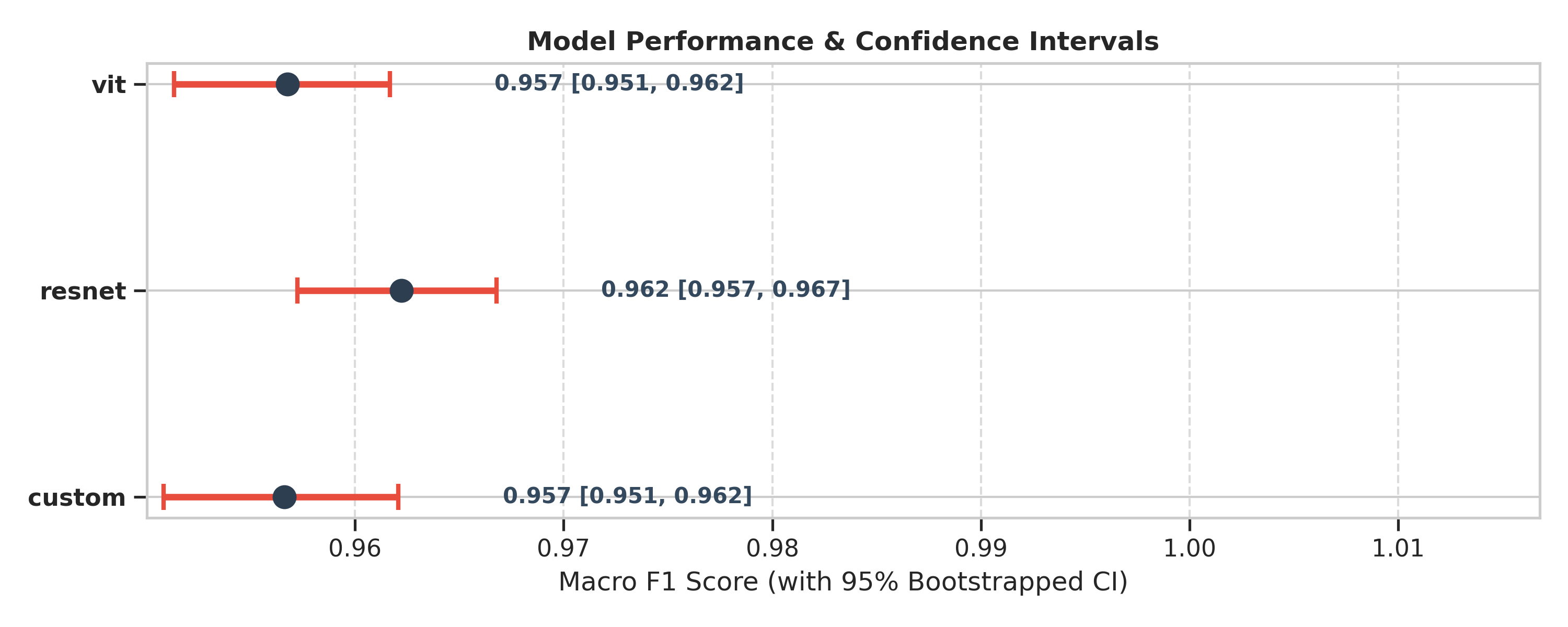}
    \caption{Forest plot with the mean macro F1 performance of each architectures (ResNet, CustomCNN, and ViT) for the Malaria dataset along with the 95\% confidence interval (CI). The dot represents the mean performance while the red bars denote the 95\% CI interval.}
    \label{fig:confidence_interval_malaria}
\end{figure}
The findings of the SHAP analysis (see Figure:~\ref{fig:malaria_shap}) are consistent with extant research; however, our framework extends these insights through rigorous quantification. In the context of malaria blood-smear models, the framework utilizes DeepLIFT-approximated Shapley values ($\phi$) to decompose the model's log-odds output. For the "uninfected" class, the models assign positive attribution ($\phi > 0$) to the erythrocyte's interior regions exhibiting low-frequency spatial variation, specifically, the smooth, homogeneous cytoplasm. Conversely, high negative importance ($\phi < 0$) is assigned to high-contrast, high-frequency anomalies, such as ring-like chromatin structures~\citep{cui2010chromatin}. This finding validates the model's emphasis on the absence of compact parasitic inclusions and the presence of globally uniform cell morphology as paramount evidence for a non-parasitic pathological image, as reported in the prior works~\citep{ahamed2025improving, lamoureux2024biophysical}.
For the Infected class, the framework demonstrates a superior capacity for precise feature localization and noise suppression (see Figure: ~\ref{fig:malaria_shap}). A visual analysis of the data indicates that the Bayesian Meta-Learning mechanism correctly identified the custom model CNN as the primary expert ($w = 0.37$). The technical advantage of this approach is evident in the transformation of raw SHAP scores into Basic Probability Assignments (BPA). The framework employs a methodology that entails the scaling of the evidential mass of each model by its Bayesian weight, as opposed to the mere averaging of feature maps. The subsequent Dempster-Shafer fusion (see Figure: ~\ref{fig:malaria_ubiqcon}) computes the orthogonal sum of these masses, which mathematically penalizes disagreement. Due to the fact that the ResNet and ViT models yielded conflicting evidence, their contribution to the Belief Mass ($Bel$) was discounted via the conflict constant ($K$). Consequently, the fused Belief map (dark green) does not merely highlight the cell but specifically isolates the parasitic biomarker, thereby effectively performing semantic segmentation of the pathology. This approach ensures that high confidence is derived exclusively from the parasite, effectively filtering out the cell wall and cytoplasm as irrelevant background.  In the case of the uninfected subject, the framework validates distributed textural evidence through the separation of Plausibility and Uncertainty (see Figure: ~\ref{fig:malaria_shap}). In contrast to the focal detection that is necessary for the identification of infections, the confirmation of "health" necessitates a holistic validation of cytoplasmic integrity. The technical superiority of the aforementioned framework is evident in the Plausibility Map (PM), which is mathematically defined as the upper probability bound $PM=1-Bel(Infected)$.  As illustrated in Figure:~\ref{fig:malaria_ubiqcon}, the absence of high-contrast features detected by the expert models to support the "Infected" hypothesis results in the "Infected" mass remaining near zero. Consequently, the Plausibility of "Healthy" remains maximal across the cell body. This process effectively functions as a "consistency check," thereby confirming that no hidden features contradict the diagnosis. Finally, the Uncertainty Map ($U$), calculated as the epistemic gap $U = \text{Pl} - \text{Bel}$, enhances system safety by explicitly delineating the anatomical Region of Interest (ROI). The bright yellow regions, which approximate a value of 1.0, signify a complete absence of knowledge regarding the slide background. In contrast, the cellular region demonstrates a reduction in uncertainty. This distinction serves to affirm that the model does not rely on background noise or slide artifacts but rather engages in the active processing of the morphological integrity of the cell itself.\\
\\
Furthermore, a distribution analysis of the framework was conducted (Figure:~\ref{fig:malaria_analytics}). This analysis functions as a quantitative validation layer. The purpose of this layer is to confirm the statistical foundation of the pixel-wise fusion maps and the functionality of the ensemble weighting mechanism. The Evidence Distribution Kernel Density Estimate (KDE) plot is a visual representation of the probability density function of the epistemic metrics, showing the Belief Mass (Green) peaking sharply near 0.0 with a thin tail extending toward 1.0. This distribution is mathematically valid for medical imaging, as pathological features like parasites typically occupy less than 10\% of the image area. The tail confirms the model successfully found high-confidence evidence in these specific, localized regions. The Uncertainty Mass (purple) exhibits a pronounced peak near 1.0, thereby validating the architecture in which the model accurately identifies a state of total ignorance regarding the vast, empty background. The Plausibility (Blue) curve reflects the Uncertainty, thereby confirming the mathematical relationship defined by the framework. In addition, the Bayesian Model Confidence bar chart demonstrates the Dirichlet Posterior Weights, thereby identifying the custom CNN as the most reliable expert ($w=0.366$). Conversely, the ResNet and ViT models are penalized for noise, with weights of $w=0.320$ and $w=0.315$, respectively. The balanced variance indicates that the ensemble prevents mode collapse by leveraging diversity rather than depending on a single architecture.
These charts function as statistical aggregations of the spatial maps generated earlier in the pipeline, thereby providing a global summary of the local explanations. The Bayesian Confidence chart directly dictates the influence of the SHAP feature attributions; by assigning a higher weight to the "custom" model, its clean and focused feature maps mathematically overpower the noisy contributions of the weaker models during the weighted sum operation of the fusion step. In addition, the Evidence Distribution chart functions effectively as a histogram of the fusion maps. The visual \textit{Green Spots} of the parasite in the Belief Map correspond directly to the tail of the green density curve ($x > 0.5$), while the vast \textit{White Background} aligns with the peak at $x=0$. In a similar manner, the \textit{Bright Yellow} background of the Uncertainty Map is represented by the substantial purple peak at approximately 1.0, and the \textit{Dark Purple} regions of certainty correspond to the dip in the purple curve. This indicates a direct correlation between the spatial visualizations and the statistical data. 
A thorough examination of the charts reveals three critical behaviors of the framework. Firstly, the precipitous decline of the Belief curve signifies high specificity and low false positives. The model circumvents the hallucination of evidence across the image, reserving high belief exclusively for relevant pixels. This is imperative for mitigating alarm fatigue in diagnostics. Secondly, the prevalence of the Uncertainty curve underscores the utilization of explicit ignorance modeling, a distinguishing feature that differentiates the framework from conventional Softmax models, which arbitrarily segment background space into classes. The framework has been developed to ensure that the models do not identify significant pixels in majority of the image. The framework guarantees that the final diagnosis is derived exclusively from the relevant pixels which are present in the minority with respect to the whole image. The Bayesian weights ultimately demonstrate ensemble robustness, as no single model dominates ($w < 0.40$ for all). This finding lends further credence to the notion that the system is dependent on collective intelligence, thereby ensuring a robust, consensus-based decision process that is less vulnerable to the potential inaccuracies of a single, potentially fallible architecture.

Further, for the aforementioned dataset, we have performed a study to analyze the consistency of our framework. The predictive performance of the models were evaluated using 10-fold stratified cross-validation. From the corresponding forest plot (See Figure:~\ref{fig:confidence_interval_malaria}), we can observe that the three acrchitectures have demonstrated a highly consistent Macro F1 Score. The ResNet architecture has an average score of 96.2\% with the 95\% confidence interval (CI) being in the range of 95.2\% \& 96.2\%. The Custom CNN and the ViT achieved scores of 95.7\% (95\% CI: 95.1\%, 96.2\%) rounded up to the nearest decimal place.  The overlapping confidence suggests that the ensemble's baseline predictive power is highly stable across the 10-folds and that the uncertainty calculation is performed with reliable evidences. 
To justify the integration of the SHAP values with DST framework, we calculated the local Lipschitz~\citep{demertzis2021lipschitz,simpson2024probabilistic} stability of the attributions. The SHAP explanation with respect to the input perturbation (See Figure:~\ref{fig:shap_lipschitz_malaria}) allows us to study the effect \& stability of SHAP for the ensemble architectures for the given input image. The ResNet model has the lowest score followed by ViT among the architectures. However, the scores are very low, for the highest score being exhibited by the custom CNN model. All three models exhibit stable explanations with the scores being $<0.0012$. This allows us to observe that the model's spatial reasoning is mathematically stable and robust to minor input perturbations. This validates the SHAP maps fused by our framework represent actual features than unstable gradient noise. 
The ablation study allows us to study the empirical results that the framework captures the epistemic uncertainty which is the ignorance due to a lack of the data, in our case, the structure of the image. From the figure (See Figure:~\ref{fig:malaria_ablation_study} \&~\ref{fig:malaria_ablation_chart}), we can observe that the baseline i.e. 1x1 kernel, the model's Belief (Support) is highly localized in the dark, intra-erythrocytic parasitic inclusions which is a characteristic of Plasmodium infection~\citep{motta2023parasite}. Correspondingly, the Uncertainty (Ignorance) shows dark purple (low uncertainty) over these specific regions. As we increase the intensity of the blur to 5x5 until 23x23, which causes the progressive structural degradation, we can observe that the distinct structural boundaries of the parasite dissolves leading to lighter purple color in the Uncertainty (Ignorance) maps. The Mean Fused Uncertainty starts to increase to increase as well. 
The average uncertainty by dataset class box-plot (See Figure:~\ref{fig:malaria_error_per_class}) points to an observation that how the ensemble models interpret the different states of the pathology. Over the 10-folds, the framework expressed significantly higher uncertainty for Infected cells compared to uninfected cells.

\subsection{Alzheimer's Detection Using Brain MRI}
\begin{figure}[]
    \centering
    \begin{subfigure}[b]{\textwidth}
        \centering
        \includegraphics[width= 0.5 \textwidth]{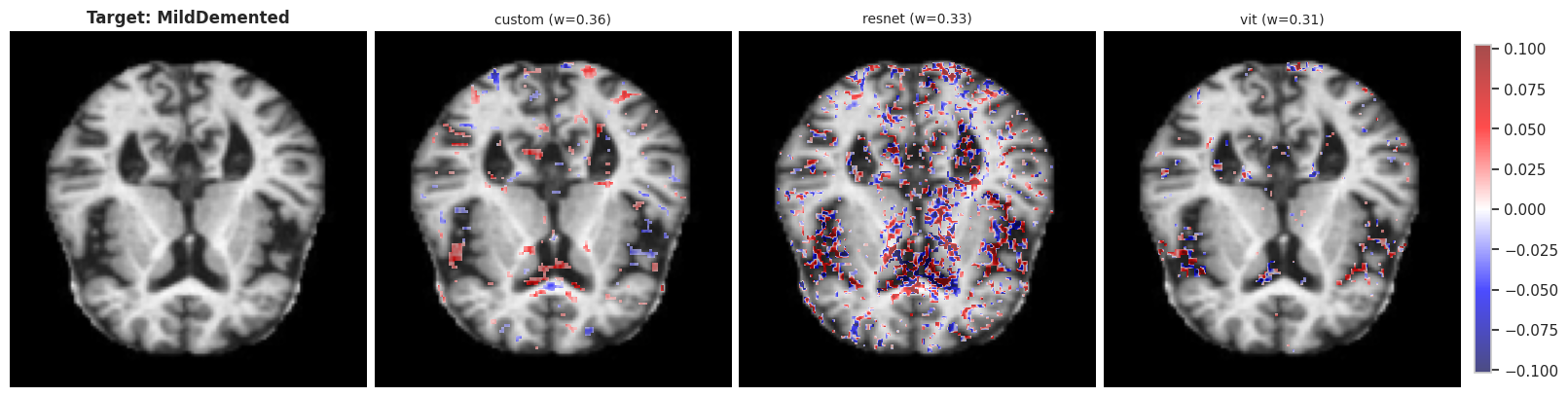}
        \caption{}
    \end{subfigure}
    \hfill 
    \begin{subfigure}[b]{\textwidth}
        \centering
        \includegraphics[width= 0.5 \textwidth]{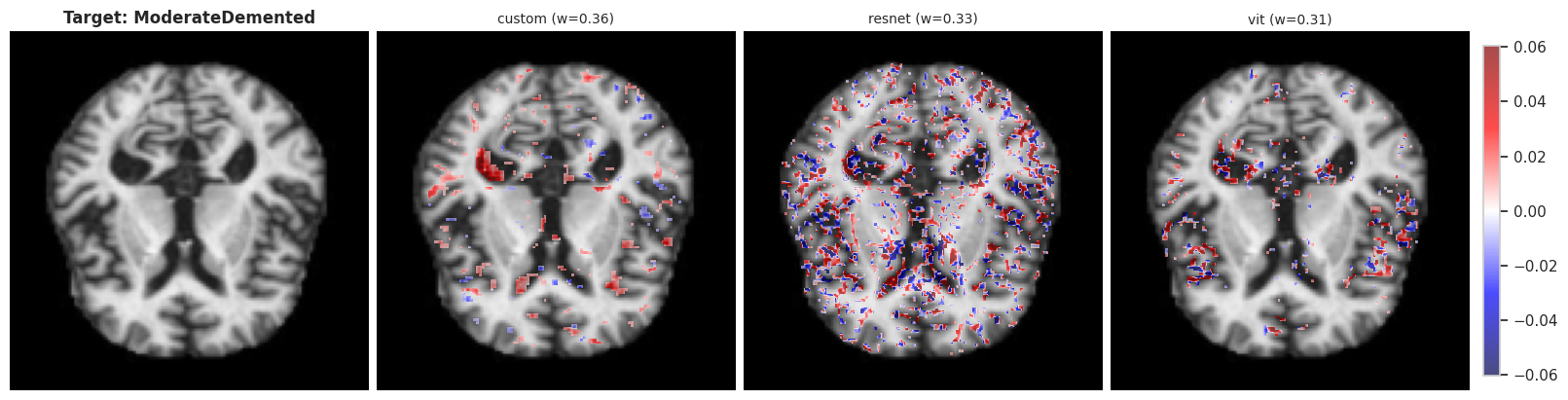}
        \caption{}
    \end{subfigure}
    \begin{subfigure}[b]{\textwidth}
        \centering
        \includegraphics[width= 0.5 \textwidth]{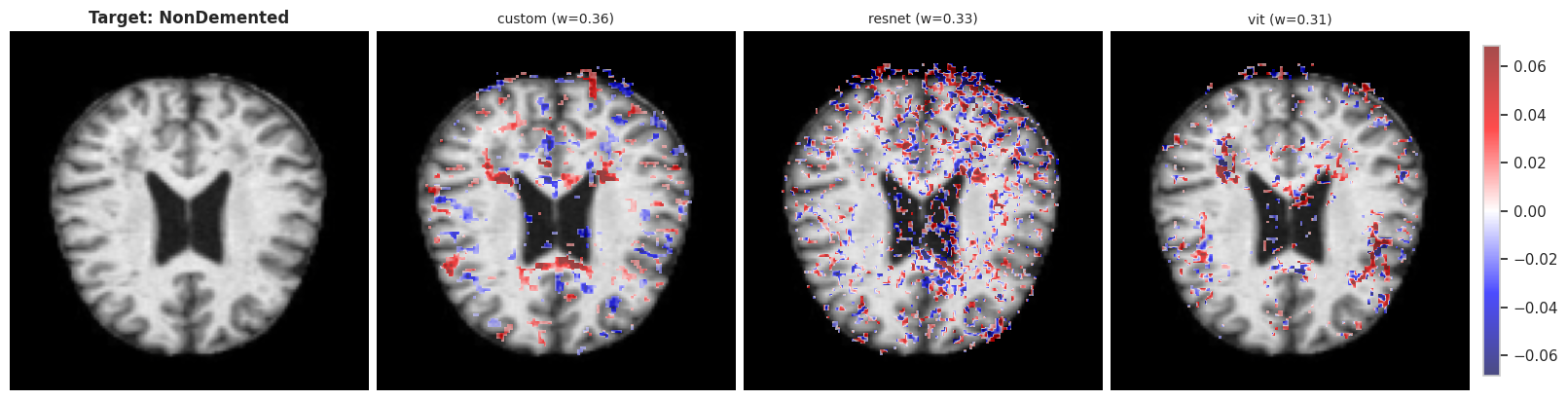}
        \caption{}
    \end{subfigure}
    \begin{subfigure}[b]{\textwidth}
        \centering
        \includegraphics[width= 0.5 \textwidth]{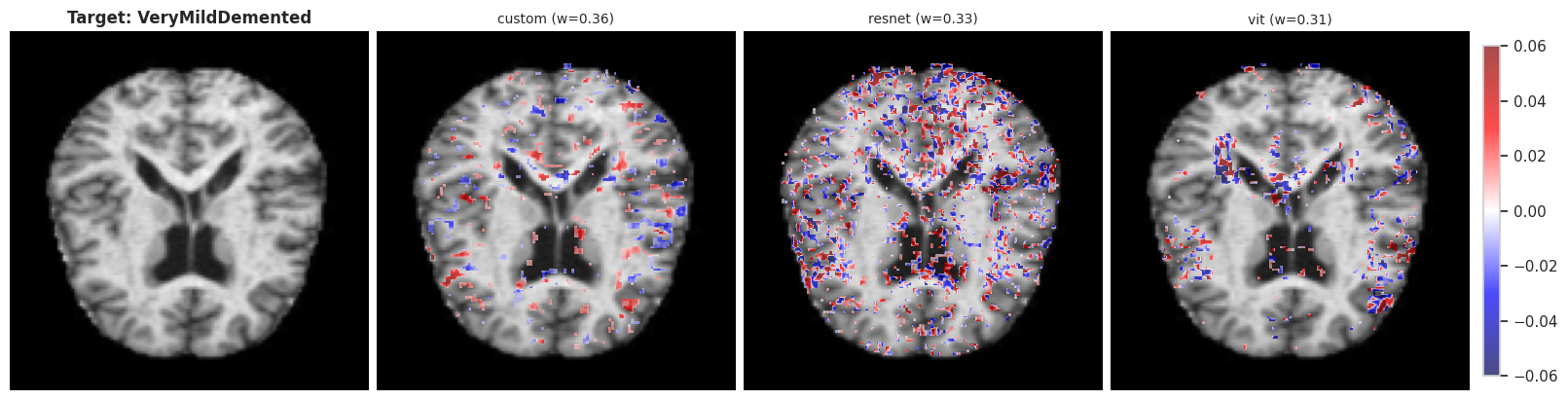}
        \caption{}
    \end{subfigure}
    \caption{This visualization shows the SHAP attribution maps ($\phi$) for each dementia stage. It details the additive feature attribution scores for the ensemble members, where the color intensity corresponds to the impact on the model's log-odds output. Red pixels denote positive SHAP values (phi > 0), indicating morphological regions, such as enlarged ventricles or cortical atrophy, that drive classification toward a specific dementia stage. Conversely, blue pixels indicate negative SHAP values ($\phi < 0$), representing features that lower the probability of the target class. The spatial distribution of the SHAP values varies significantly by architecture. For example, the ResNet model frequently exhibits scattered, high-frequency pixel importance, while the Custom CNN and ViT models display contiguous regions of semantic relevance. This demonstrates how distinct internal representations lead to different pixel-wise evidence-gathering strategies.}
    \label{fig:brain_shap}
\end{figure}
\begin{figure}[]
    \centering
    \begin{subfigure}[b]{\textwidth}
        \centering
        \includegraphics[width= 0.5 \textwidth]{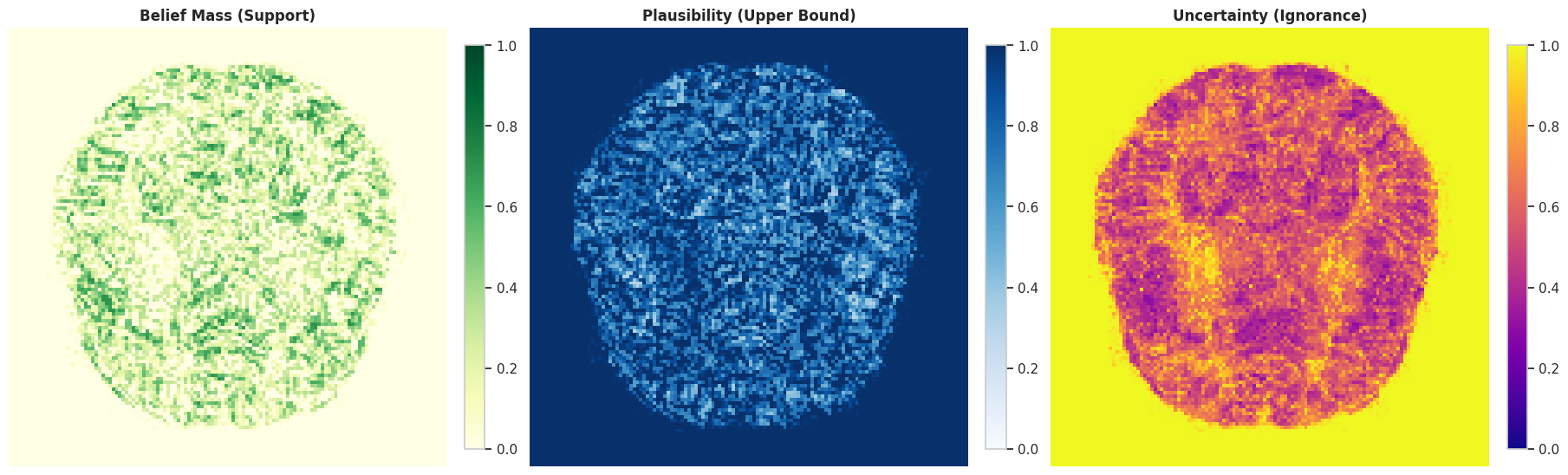}
        \caption{}
    \end{subfigure}
    \hfill 
    \begin{subfigure}[b]{\textwidth}
        \centering
        \includegraphics[width= 0.5 \textwidth]{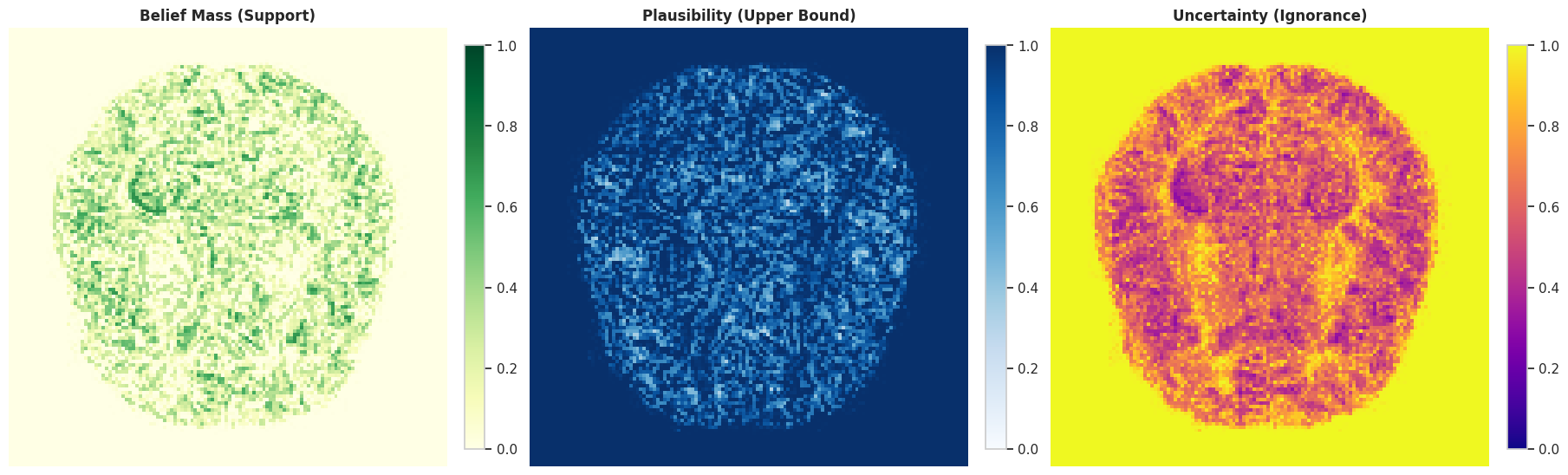}
        \caption{}
    \end{subfigure}
    \begin{subfigure}[b]{\textwidth}
        \centering
        \includegraphics[width= 0.5 \textwidth]{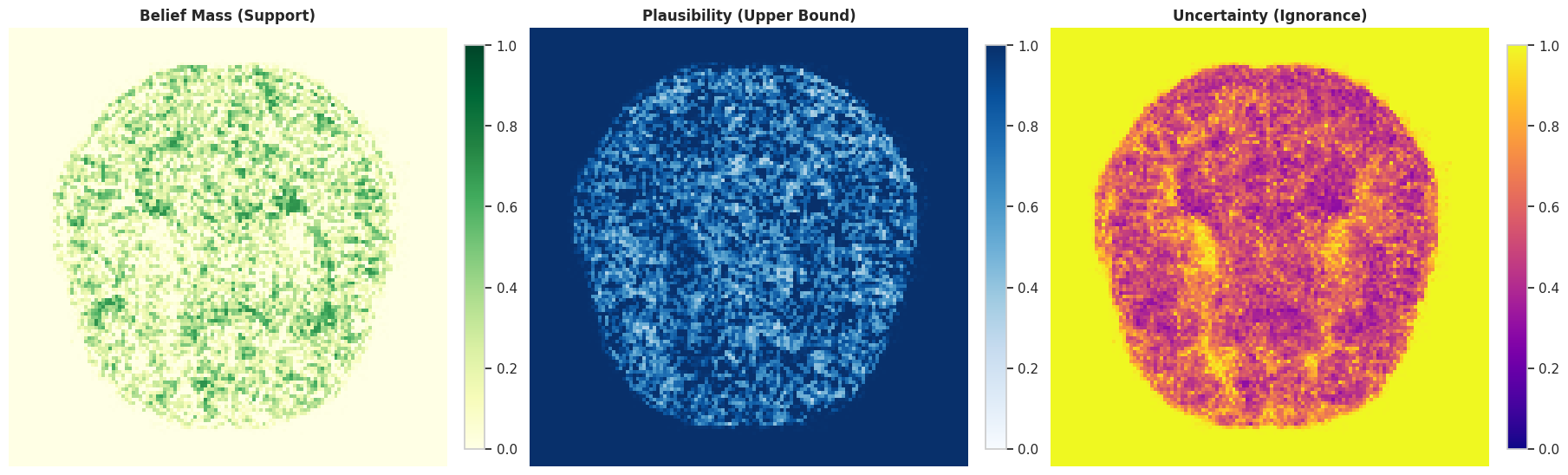}
        \caption{}
    \end{subfigure}
    \begin{subfigure}[b]{\textwidth}
        \centering
        \includegraphics[width= 0.5 \textwidth]{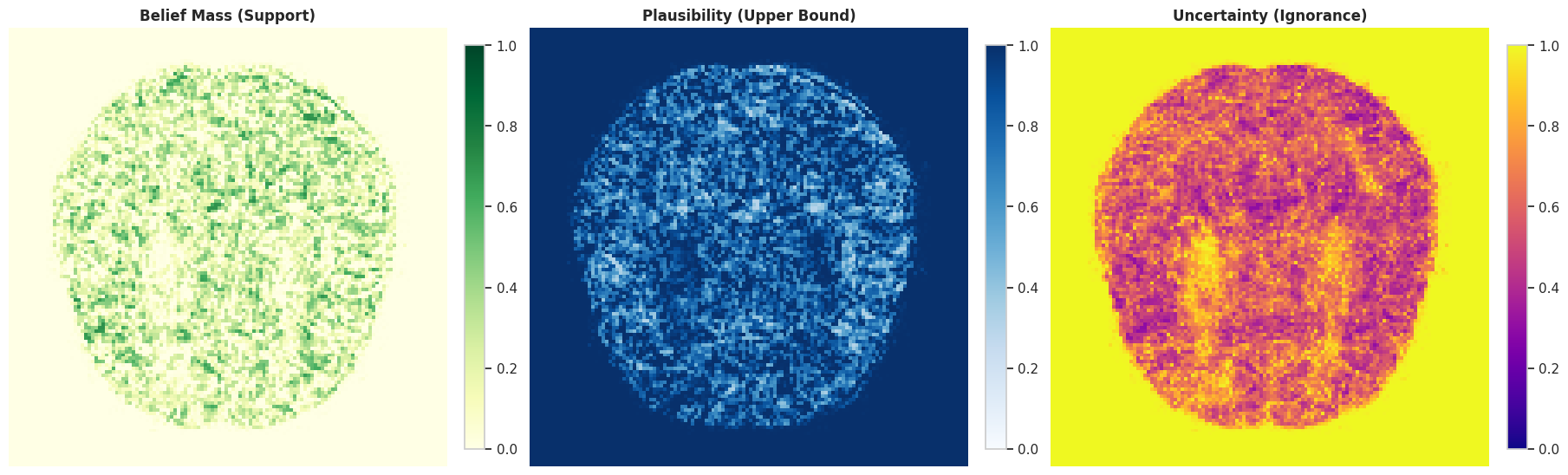}
        \caption{}
    \end{subfigure}
    \caption{This figure illustrates the correlation between feature distinctness and model confidence by presenting the pixel-wise fusion of SHAP explanations from the weighted model ensemble (Custom CNN, ResNet, and ViT) across four stages of dementia. (a) Mild Dementia: The belief mass (support) map shows localized clusters of evidence (green) that correspond to emerging pathological features. The uncertainty map shows a mix of orange and purple. This indicates that the ensemble detects specific biomarkers but has moderate epistemic uncertainty compared to advanced stages. (b) Moderately Demented: Representing the model’s optimal performance zone, this sample shows the strongest reduction in uncertainty (dark purple regions approaching 0.0 ignorance). The concentrated belief mass aligns with distinct structural atrophy, confirming that the ensemble reaches a high-confidence consensus when processing unambiguous neurodegenerative lesions. (c) Non-demented: In sharp contrast, the healthy brain exhibits diffuse, scattered belief and uniform, high uncertainty (bright yellow). This indicates a cautious decision-making process, as the prediction relies on the absence of disease markers rather than the detection of positive features. This correctly reflects the ensemble's high level of ignorance regarding the disparate background tissue. (d) Very mild dementia: Reflecting a performance limitation, this map closely mirrors the non-demented profile, with widespread high uncertainty and weak feature support. This suggests that the ensemble struggles to distinguish the subtle microstructural changes of early-stage pathology from healthy noise.}
    \label{fig:brain_ubiqcon}
\end{figure}

\begin{figure}[]
    \centering
    \begin{subfigure}[b]{\textwidth}
        \centering
        \includegraphics[width= 0.45 \textwidth]{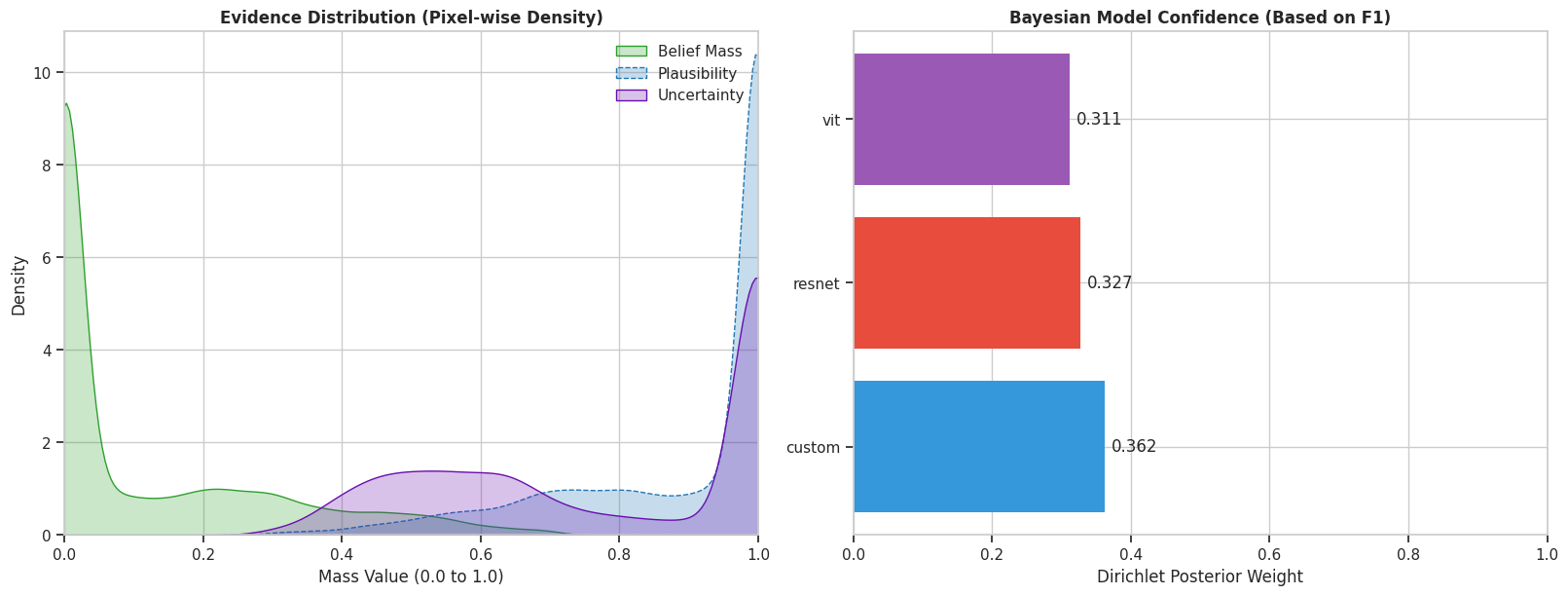}
        \caption{}
    \end{subfigure}
    \hfill 
    \begin{subfigure}[b]{\textwidth}
        \centering
        \includegraphics[width= 0.45 \textwidth]{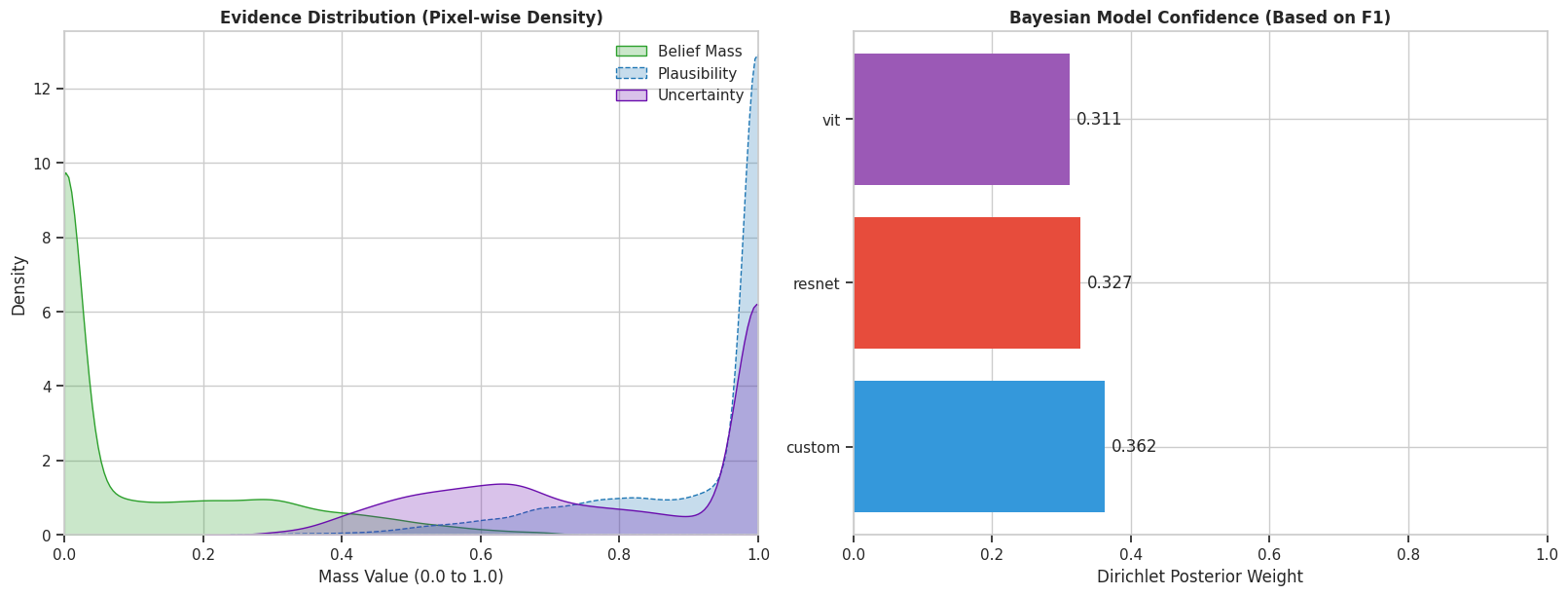}
        \caption{}
    \end{subfigure}
    \begin{subfigure}[b]{\textwidth}
        \centering
        \includegraphics[width= 0.45 \textwidth]{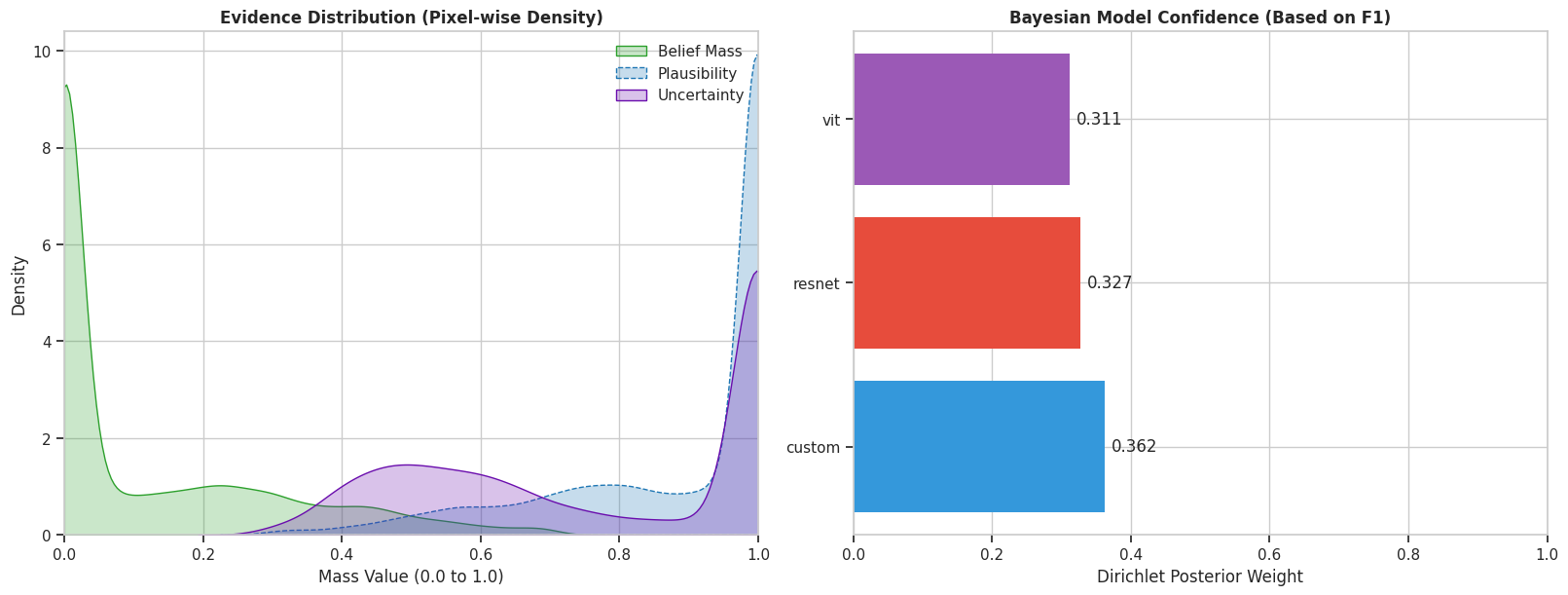}
        \caption{}
    \end{subfigure}
    \begin{subfigure}[b]{\textwidth}
        \centering
        \includegraphics[width= 0.45 \textwidth]{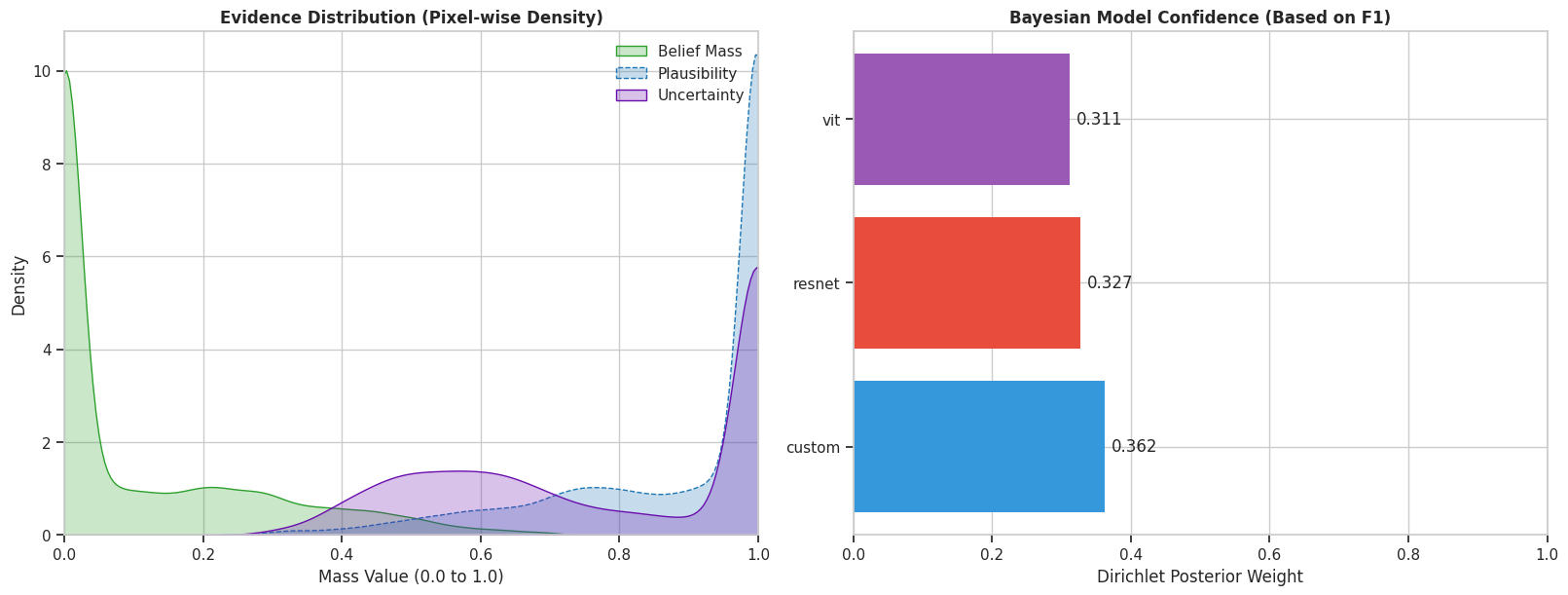}
        \caption{}
    \end{subfigure}
    \caption{This figure shows the quantitative analytics of the fusion engine by contrasting the statistical evidence distribution on the left with the assigned Bayesian model confidence on the right. The bar charts confirm that Custom CNN ($w=0.362$) has the greatest influence on the ensemble, followed closely by ResNet and ViT. This indicates a preference for local texture features over global dependencies. The kernel density estimation (KDE) plots reveal the pixel-wise behavior of the evidential masses across the four classes. In the "Moderate Demented" sample, the belief mass (green) has a tail that extends to higher values. This validates the detection of distinct pathological features. The uncertainty distribution (purple) captures the reduction in ignorance within the lesions. In contrast, for the non-demented and very mild cases, the belief mass is strictly compressed near zero, and the uncertainty curve is heavily skewed toward 1.0. This confirms that, for healthy or ambiguous samples, the model's decision is driven primarily by epistemic uncertainty (lack of evidence) rather than positive feature detection.}
    \label{fig:brain_analytics}
\end{figure}
\begin{figure}
    \centering
    \includegraphics[width=0.5\linewidth]{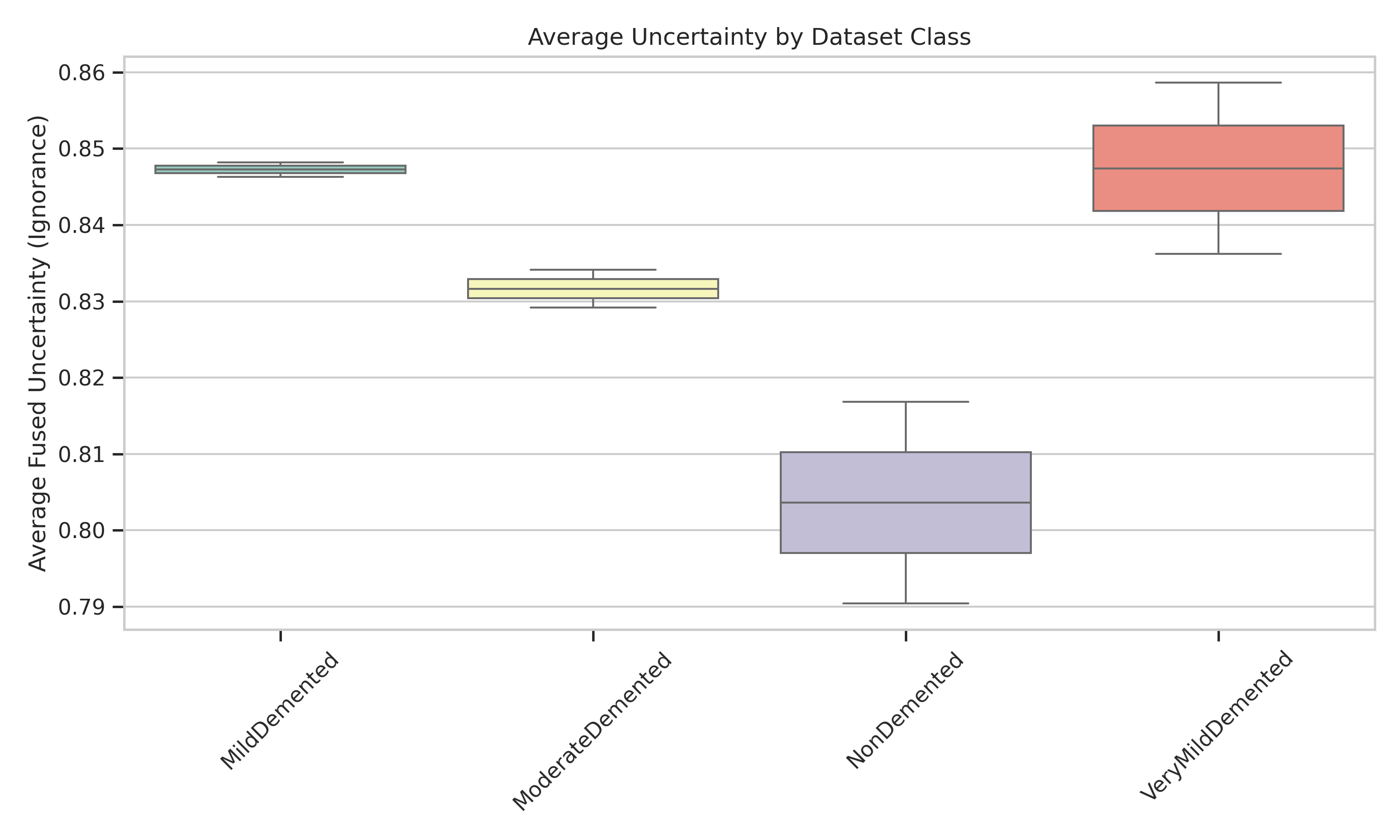}
    \caption{This figure denotes a box plot allowing us to analyze the mean average Uncertainty (Ignorance) for the Brain MRI dataset to study alzheimers. The NonDemented class has the lowest mean fused Uncertainty (Ignorance) while the VeryMildDemented has the highest. The MilDemented has the lowest spread followed by the ModerateDemented class.}
    \label{fig:brain_error_per_class}
\end{figure}
\begin{figure}
    \centering
    \includegraphics[width=0.5\linewidth]{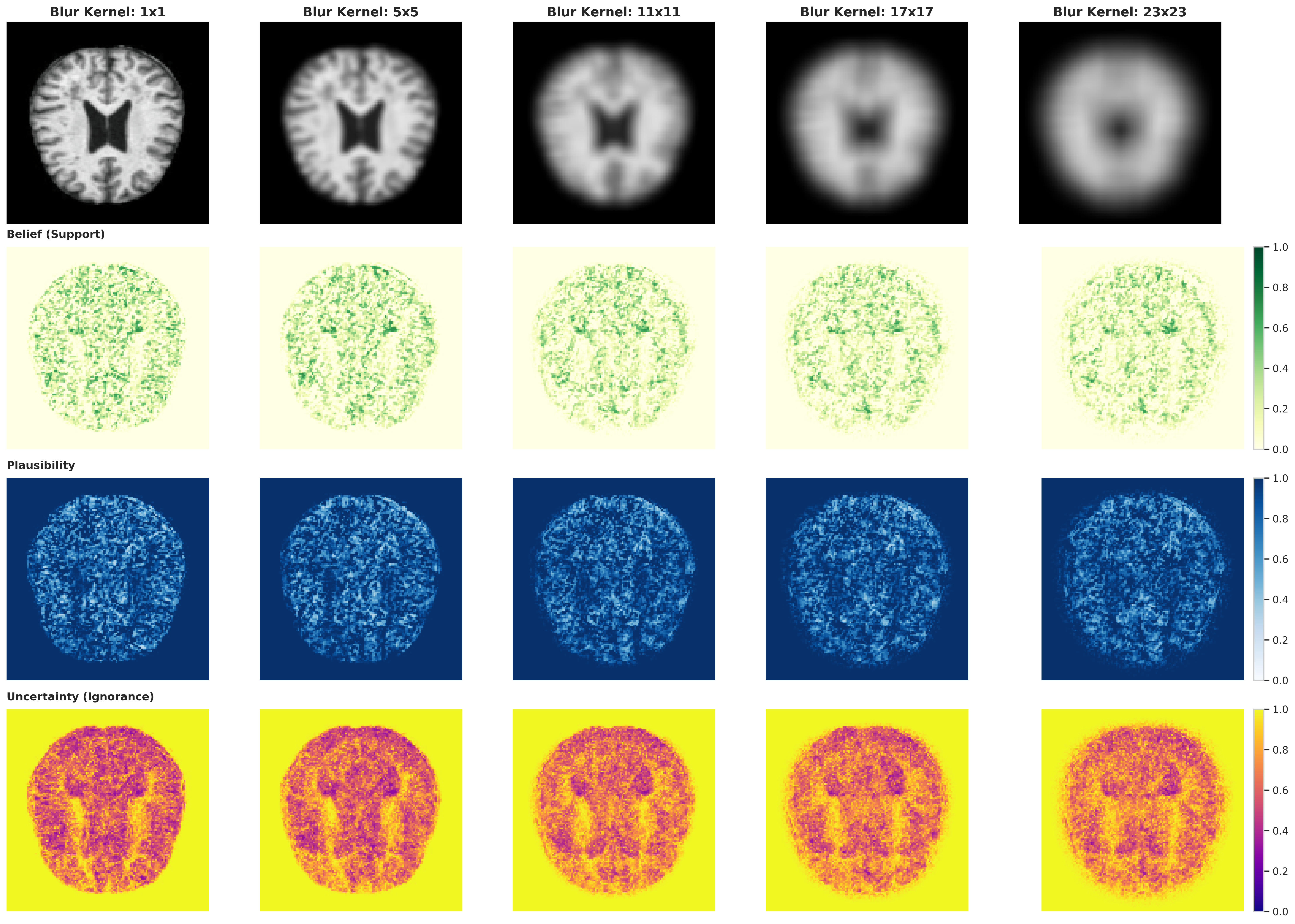}
    \caption{The visual illustration of the framework's  for progressive structural degradation through the introduction of the Gaussian blur. The top row of the figure displays the input image at increasing blur kernel sizes, ranging from 1x1 to 23x23. Subsequent rows illustrate the corresponding Belief (Support), Plausibility, and Uncertainty (Ignorance) maps generated by the DS-SHAP fusion framework for the Brain MRI dataset. The increase in the Gaussian kernel increases the uncertainty and the boundaries start to fade away leading to higher Uncertainty (Ignorance).}
    \label{fig:brain_ablation_study}
\end{figure}
\begin{figure}
    \centering
    \includegraphics[width=0.5\linewidth]{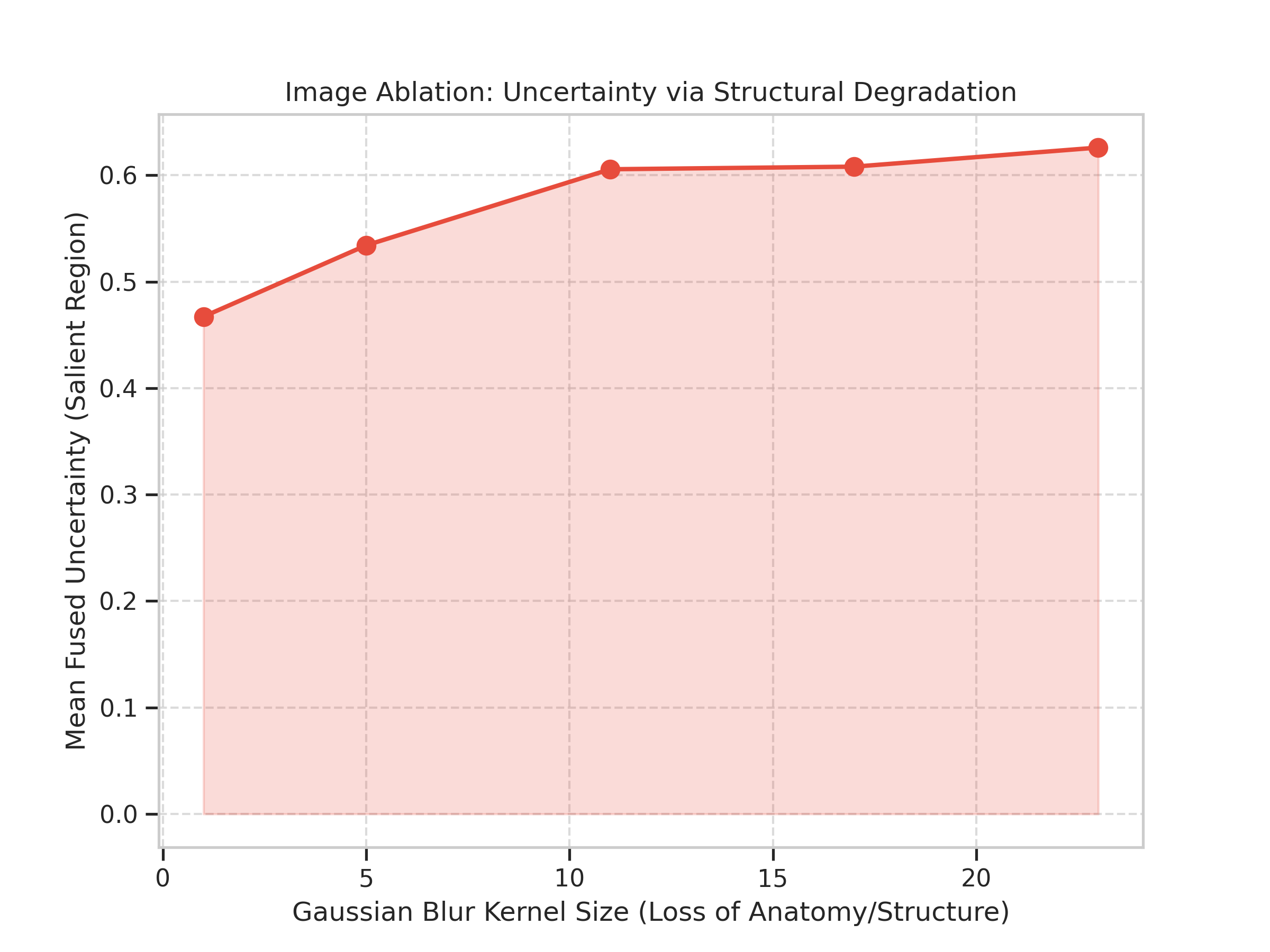}
    \caption{This figure illustrates the impact of structural degradation, manifested through the incorporation of a Gaussian blur kernel, on the mean fused uncertainty. The augmentation of the Gaussian blur kernel has been demonstrated to enhance the mean fused uncertainty for the Brain MRI dataset. The following line chart is intended to quantify the relationship between structural integrity and the framework's calculated epistemic uncertainty. The x-axis of this figure represents the increasing Gaussian blur kernel size, or the loss of anatomical detail, while the y-axis shows the mean fused uncertainty, or the degree of ignorance, measured exclusively within the salient anatomical regions.}
    \label{fig:brain_ablation_chart}
\end{figure}
\begin{figure}
    \centering
    \includegraphics[width=0.5\linewidth]{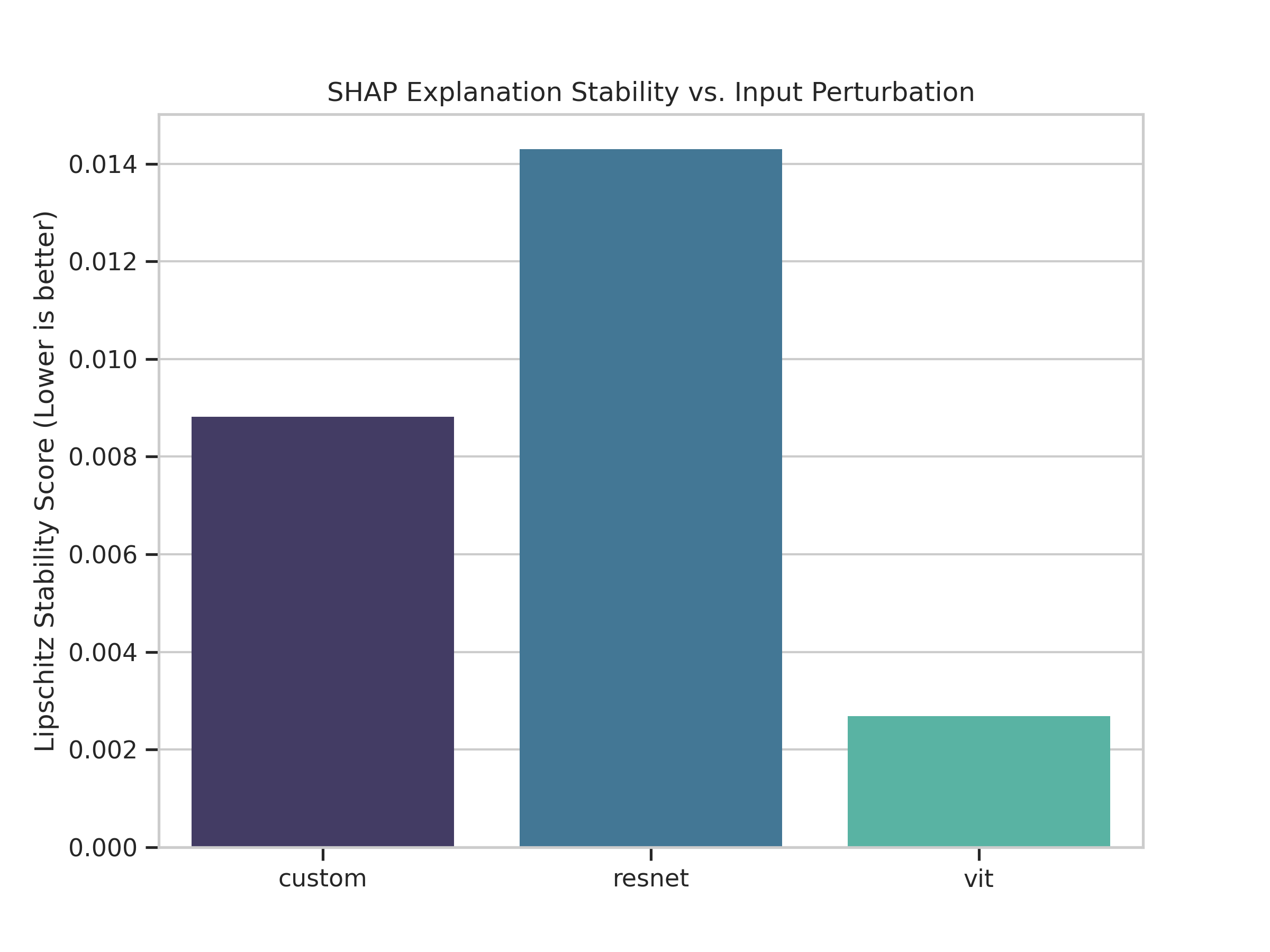}
    \caption{This figure indicates the stability of the SHAP attributions for the architectures employed in the analysis of images from the Brain MRI dataset. The Resnet demonstrates the highest Lipschitz score (lower the better), indicating the least stable attributions in comparison to the CustomCNN and ViT architectures. However, the scores are low, which indicates that the attribution which is used in our framework to quantify uncertainty is overall stable.}
    \label{fig:shap_lipschitz_brain}
\end{figure}
\begin{figure}
    \centering
    \includegraphics[width=0.5\linewidth]{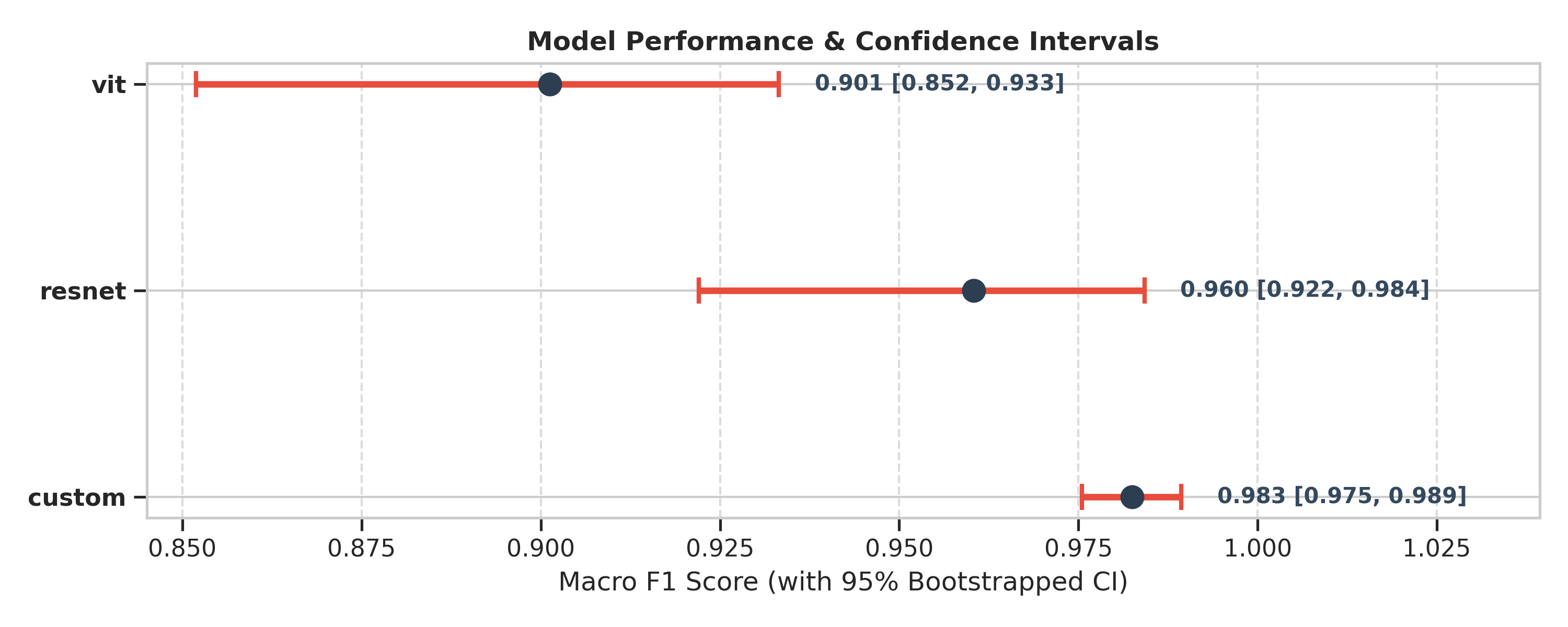}
    \caption{Forest plot with the mean macro F1 performance of each architectures (ResNet, CustomCNN, and ViT) for the Brain MRI dataset for alzheimers analysis along with the 95\% confidence interval (CI). The dot represents the mean performance while the red bars denote the 95\% CI interval.}
    \label{fig:confidence_interval_brain}
\end{figure}
The proposed framework was evaluated using a multi-class Alzheimer's MRI dataset, which was employed to differentiate between individuals exhibiting no dementia, those with very mild dementia, those with mild dementia, and those with moderate dementia, based on structural T1-weighted scans. The Bayesian Meta-Learning module exhibited noteworthy consistency in evaluating the architectural suitability for neuroimaging tasks, effectively replicating its performance on the retinal dataset. The custom CNN architecture was identified as the primary expert with a posterior reliability weight of approximately 0.36, while the ResNet and ViT models were assigned lower weights of approximately 0.33 and 0.31, respectively. This probabilistic ranking is visually substantiated by the feature attribution maps (Figure:~\ref{fig:brain_shap}), where the activations of the custom model are notably sparse and localized, adhering strictly to ventricular boundaries and regions of cortical atrophy. In stark contrast, the Resnet model exhibited substantial aleatoric uncertainty, manifesting as high-frequency noise and activation in random white matter regions that were not pertinent to the pathology. Conversely, the ViT model generated coarser, block-like activations due to its patch-based tokenization approach. The framework's inherent capacity to dynamically attenuate the noisy ResNet model is of paramount clinical significance, as it mathematically mitigates stochastic hallucinations that could otherwise result in false-positive diagnoses in a standard averaging ensemble.
The framework demonstrated a sophisticated capacity to capture the non-linear progression of brain atrophy, the primary biomarker for Alzheimer's disease (Figure:~\ref{fig:brain_shap}), by modulating its evidential confidence according to disease severity. In instances of moderate dementia, characterized by substantial enlargement of the lateral ventricles due to parenchymal tissue loss~\citep{apostolova2012hippocampal}, the custom model accurately identifies this structural deformation as the diagnostic signal. The positive SHAP attributions ($\phi > 0$, red pixels) align precisely with the ventricular edges, a finding corroborated by the Dempster-Shafer fusion maps (Figure:~\ref{fig:brain_ubiqcon}) where the Belief Map ($\text{Bel}$) displays dense, localized green clusters representing confirmed pathological evidence. Conversely, for cases classified as Very Mild or Mild Demented, where atrophy is subtle and manifests exclusively as sulcal widening or volume loss in the hippocampus~\citep{manera2022ventricular}, the attribution maps reflect this ambiguity through reduced pixel intensity. The Uncertainty Map ($U$) accurately captures this clinical nuance (Figure:~\ref{fig:brain_ubiqcon}). In contrast to the deep purple core of certainty seen in severe cases, mild cases exhibit higher entropy (orange/grey hues) within the grey matter. This phenomenon mirrors the process of human radiological assessment, wherein early-stage diagnoses inherently entail higher epistemic uncertainty compared to diagnoses made in later stages of the disease~\citep{shi2015studying}. In the analysis of subjects without dementia, the framework shifts its evidential focus toward verifying structural integrity. The expert models distribute their attention across the white matter and normal-sized ventricles, effectively treating the absence of ventricular enlargement as positive evidence for the "Healthy" class. However, the Plausibility Map (PM) reveals an underlying layer of internal ensemble conflict that a straightforward prediction label would fail to disclose. The noisy ResNet model highlights random brain regions as "pathological" (negative attribution), directly contradicting the custom model's assessment. This discrepancy manifests visually as lighter blue spots within the dark blue Plausibility map, mathematically lowering the upper bound of trust. By transparently visualizing this discordance, the framework alerts the clinician that while the consensus diagnosis is "Non-Demented," the models are not in total alignment, prompting a more careful review than a unanimous decision would require.
The application of the framework, to neuroimaging data underscores its utility as a robust tool for automated diagnosis. The Uncertainty Map methodically delineates the anatomical Region of Interest (ROI) from the background, assigning total ignorance (bright yellow) to the empty space and varying degrees of certainty (orange and purple) to the brain tissue. This distinction serves to demonstrate that the model in question does not rely on preprocessing artifacts, such as skull-stripping errors or background noise, to generate predictions. Moreover, the system undergoes an evolution from a "black box" classifier to a a reasoning framework by translating the diagnosis into a spatial belief map. The spatial explanation provided by the system—namely, "Demented because of this specific region"~--~enables clinicians to validate the system's detection of genuine atrophy or reaction to spurious imaging artifacts. This validation step is essential for the clinical deployment of deep learning systems.

We also performed the quantitative validation of the framework for Alzheimer's disease diagnosis confirm that the ensemble weighting mechanism correctly penalizes weak learners while aligning evidential metrics with the biological reality of neuroimaging data (Figure:~\ref{fig:brain_analytics}). The Kernel Density Estimate (KDE) plot of the Evidence Distribution illustrates the probability density functions of the three epistemic metrics across the MRI slice. The Belief Mass (Green Curve) exhibits a mathematically valid long-tail distribution peaking sharply at 0.0 and extending thinly toward 1.0. This distribution is anticipated due to the fact that brain atrophy biomarkers, such as enlarged ventricles and cortical thinning, constitute a minimal percentage of the total image volume. Consequently, the sharp peak at zero confirms the model's correct exclusion of non-diagnostic pixels, such as white matter bulk and the skull, while the tail signifies high-confidence detection in localized regions of interest. Concurrently, the Uncertainty Mass (purple curve) exhibits a pronounced right-skewed distribution, with a maximum near 1.0. This outcome substantiates the system's "Safety-First" design by recognizing the absence of knowledge concerning the empty background and non-informative tissue. This distinction differentiates this framework from conventional softmax models, which impose classification on empty space. The Plausibility (Blue Dashed Curve) demonstrates a close correlation with the Uncertainty curve, thereby substantiating the mathematical relationship that stipulates the maximum plausibility in the absence of contradictory evidence that opposes the diagnosis. The Bayesian Model The bar chart, which serves as a complementary visual tool to the density plots, presents the Dirichlet posterior weights that were derived from the validation phase. These weights consistently place the custom CNN at the primary expert level, with a weight of approximately 0.36. This finding suggests that the custom architecture was most effective at capturing structural atrophy features without overfitting, whereas the ResNet model was consistently penalized with the lowest weight ($w\approx 0.31$) due to its tendency to produce high-frequency noise across the white matter. The ViT model was designated as a secondary expert, providing structural context but exhibiting a lack of precision in comparison to the custom CNN. The utilization of non-uniform weights serves to substantiate the efficacy of the meta-learner, while the balanced variance guarantees that the ultimate determination is a robust consensus rather than an aberration attributable to a solitary model. These statistical charts serve as a direct numerical aggregation of the spatial maps; the Bayesian Confidence chart dictates the weighted sum of the SHAP feature attributions, ensuring that the final fusion is dominated by the clean, anatomical signal of the expert model while mathematically suppressing the noise from the ResNet model before the Dempster-Shafer rule is applied. The correlation between the statistical distributions and the spatial fusion maps is evident, with the Evidence Distribution chart functioning as a histogram for the fusion images. The \textit{Green Clusters} of confirmed pathology in the Belief Map correspond directly to the tail of the green density curve, while the \textit{Bright Yellow} background of the Uncertainty Map aligns with the massive peak at 1.0 in the purple density curve. In a similar manner, the \textit{Dark Blue} regions of the Plausibility map correspond to the peak at 1.0. The presence of any lighter blue speckles is indicative of conflicting noise, manifesting as minor deviations. From an analytical perspective, the charts demonstrate that the model successfully achieves precise anatomical localization. This is evidenced by its ability to distinguish between pathological Alzheimer's progression and global brain aging. The model accomplishes this by localizing specific atrophy sites rather than making generalizations about the entire brain~\citep{manera2022ventricular}. In addition, the system exhibits robustness to noise by successfully identifying and penalizing the ResNet model as a source of aleatoric uncertainty. This reduces the risk of false positives from artifacts. The predominance of the Uncertainty curve underscores the system's epistemic integrity, as it acknowledges a lack of knowledge concerning the black background and healthy white matter. The system's confidence is confined to the particular structural deformations for which it has been trained; this behavior is inherently more secure than that of standard deep learning models, which frequently assign high confidence to extraneous features.

The predictive analysis of the ensemble models (See Figure:~\ref{fig:confidence_interval_brain}) using 10-fold stratified cross validation reveals that the Custom CNN achieved a Macro F1-Score of 98.3\% with 95.0\% CI ranging between 97.5\% \& 98.9\%. The ResNet model achieved a Macro F1-Score of 96.0\% with 95.0\% CI ranging between 92.2\% \& 98.4\% and ViT achieved 90.1\% with 95.0\% CI rangin between 85.2\% \& 93.3\%. 
From the Lipschitz~\citep{demertzis2021lipschitz,simpson2024probabilistic} score calculation (See Figure:~\ref{fig:shap_lipschitz_brain}), we can observe that the complexity of the architectures is inversely related to the stability of the SHAP attributions. The ViT has the most stable attributions with a score of around 0.002 while the ResNet has the lowest with the score of around 0.014. However, all the scores are relatively low emphasizing that the attributions used in our framework for the calculation of Uncertainty (Ignorance) are relatively stable, robust, and are inherently resilient. 
Further, we performed the image ablation study (See Figure:~\ref{fig:brain_ablation_chart} \&~\ref{fig:brain_ablation_study}) to study the relation between the Uncertainty (Ignorance) and the structural degradation through the introduction of Gaussian noise through the kernels. This allowed us to analyze the epistemic uncertainty due to lack of data, which here is the structure of the enclosed organ in the image. As the Gaussian blur is increased starting from the 1x1 to 11x11 and further to 23x23, we can observe that the average fused uncertainty increases from baseline of approximately 0.47 to 0.62. The visual ablation serves as the qualitative proof as well. Visually, for the 1x1 kernel the Belief \& Plausibility maps are able to identify the regions of the brain denoted by ventricular boundaries and cortical folds, while the Uncertainty maps isolates the ignorance to non-critical regions. However as the blur reaches to 23x23, the model's evidence start to conflict and the Uncertainty (Ignorance) maps relatively lose the confidence in the boundaries of the region with high-confidence ignorance. 
The average uncertainty by class (See Figure:~\ref{fig:brain_error_per_class}), allows us to study that the framework has uncovered the one of the most well-known flaws in the field of alzheimers analysis. The framework is confident with respect to the class Non-Demented which typically have distinct or preserved structural volumes denoted by relatively lower uncertainty. However, for the other classes, the framework is having difficulty to stay confident and the uncertainty is high for the Mild-Demented \& Moderate-Demented classes with relatively stable analysis. The Very Mild-Demented class is the most difficult class to identify because it is in very close proximity to neighboring stages like mild cognitive impairment and early dementia, so the features overlap and the signal is subtle~\citep{li2021applications}. This also aligns with the real-world issues profound with alzheimers analysis using ML and DL.

\subsection{Diabetic Retinopathy Detection}
\begin{figure}[]
    \centering
    \begin{subfigure}[b]{\textwidth}
        \centering
        \includegraphics[width= 0.5 \textwidth]{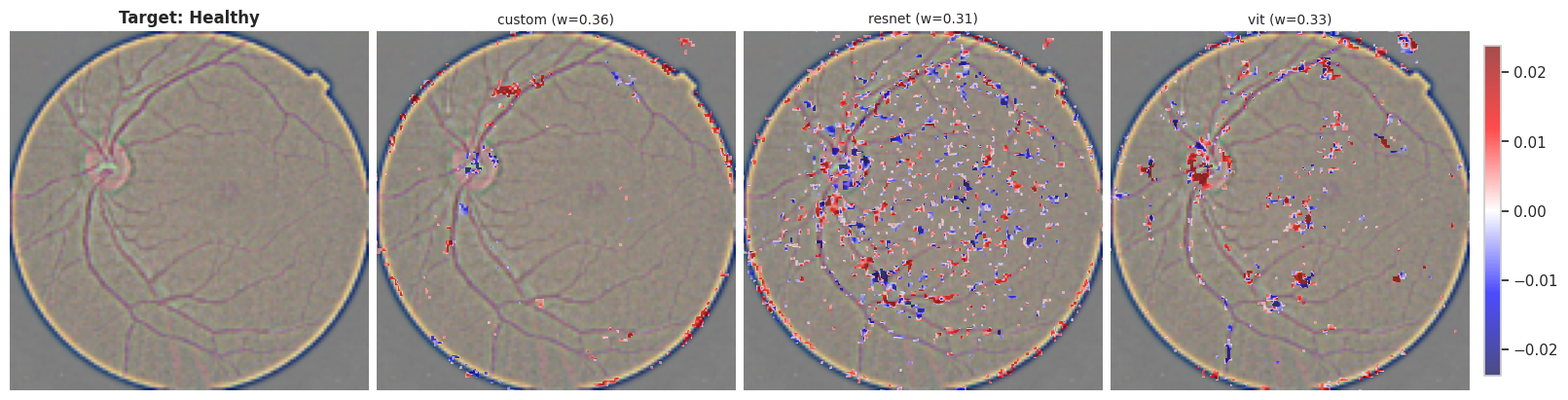}
        \caption{}
    \end{subfigure}
    \hfill 
    \begin{subfigure}[b]{\textwidth}
        \centering
        \includegraphics[width= 0.5 \textwidth]{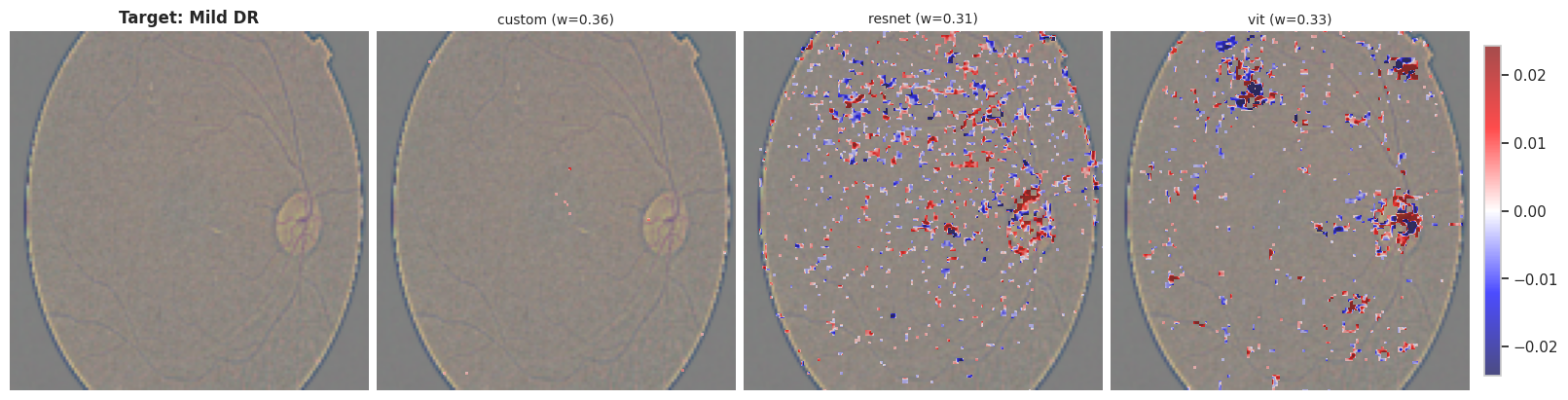}
        \caption{}
    \end{subfigure}
    \begin{subfigure}[b]{\textwidth}
        \centering
        \includegraphics[width= 0.5 \textwidth]{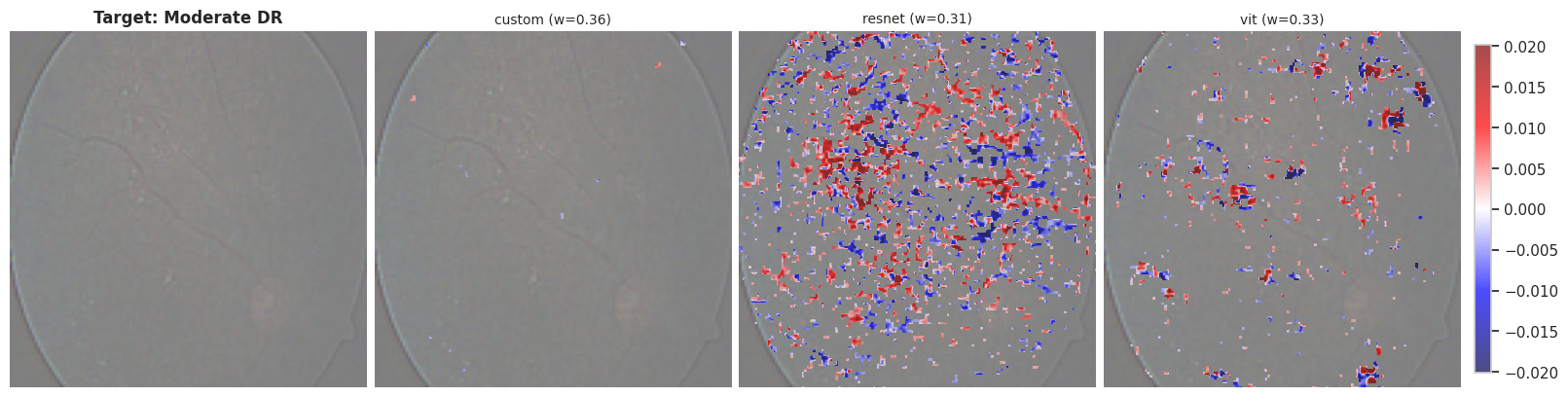}
        \caption{}
    \end{subfigure}
    \begin{subfigure}[b]{\textwidth}
        \centering
        \includegraphics[width= 0.5 \textwidth]{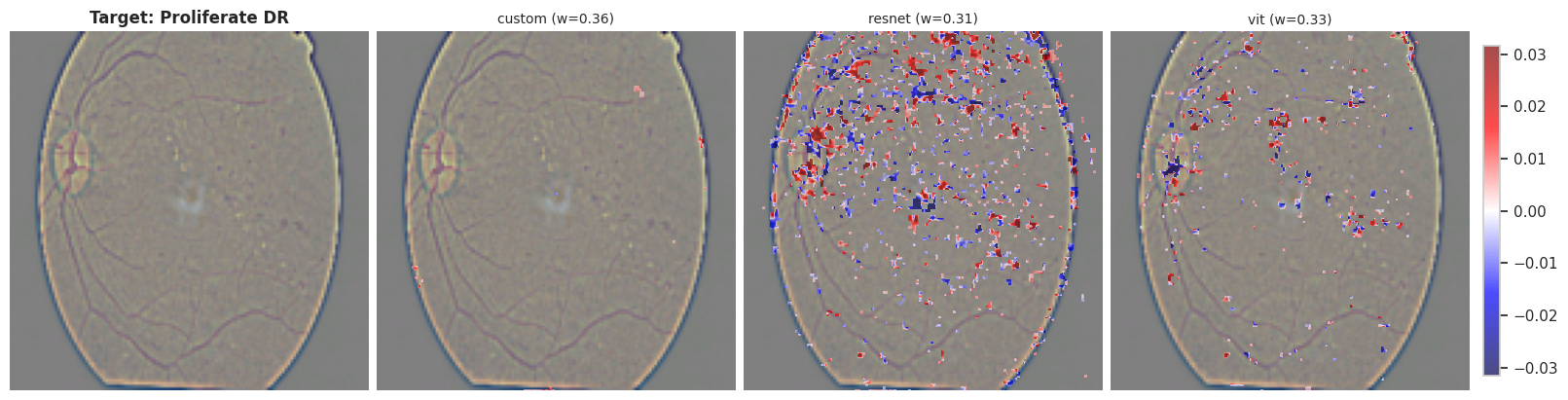}
        \caption{}
    \end{subfigure}
    \begin{subfigure}[b]{\textwidth}
        \centering
        \includegraphics[width= 0.5 \textwidth]{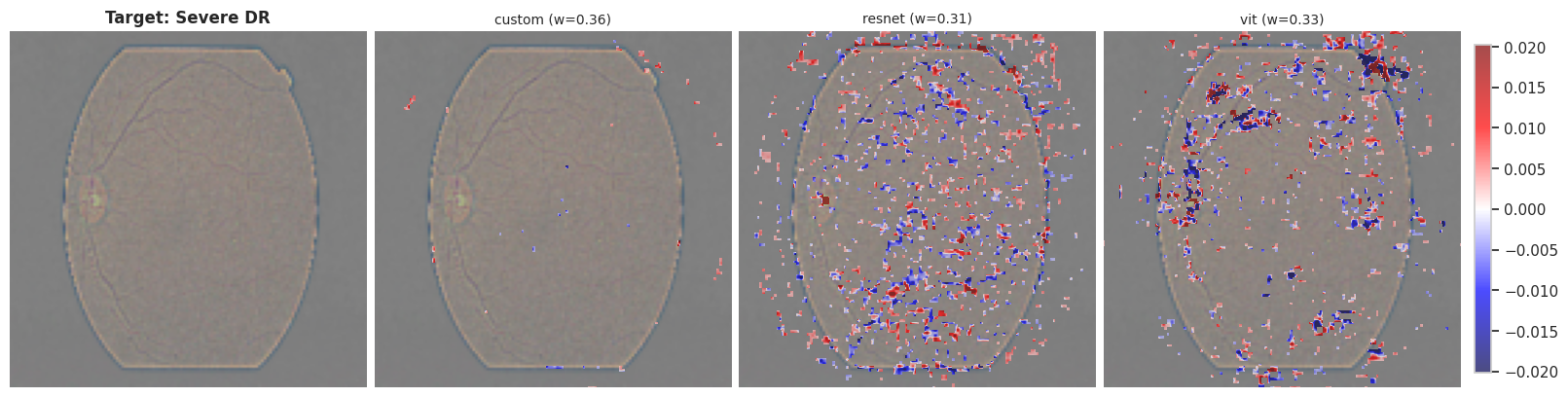}
        \caption{}
    \end{subfigure}
    \caption{This figure illustrates the fused SHAP explanations and quantitative analytics for the diabetic retinopathy (DR) classification ensemble at different severity levels. The attribution maps in the top rows visualize the pixel-wise contribution of each model (Custom CNN, ResNet, and ViT). Red indicates positive evidence for the target class, and blue indicates suppression. In advanced stages, such as Proliferative DR and Severe DR, the maps display dense, scattered clusters of positive attribution. This suggests that the models successfully localize widespread pathological markers, such as hemorrhages or microaneurysms. Conversely, the Healthy and Mild DR maps exhibit sparser, more diffuse activity, reflecting the difficulty of identifying distinct features in early-stage or normal retinas. This visual trend is supported quantitatively by the evidence distribution plots (bottom). Advanced stages show a more distinct belief mass profile (green), while healthy and mild cases are dominated by high uncertainty (purple) near 1.0. This indicates that these predictions rely on the absence of disease markers rather than positive detection. The Bayesian Model Confidence bar chart confirms that Custom CNN ($w=0.362$) carries the highest evidential weight in the fusion process, followed by ResNet ($w=0.327$) and ViT ($w=0.311$).}
    \label{fig:diabetic_shap}
\end{figure}

\begin{figure}[]
    \centering
    \begin{subfigure}[b]{\textwidth}
        \centering
        \includegraphics[width= 0.5 \textwidth]{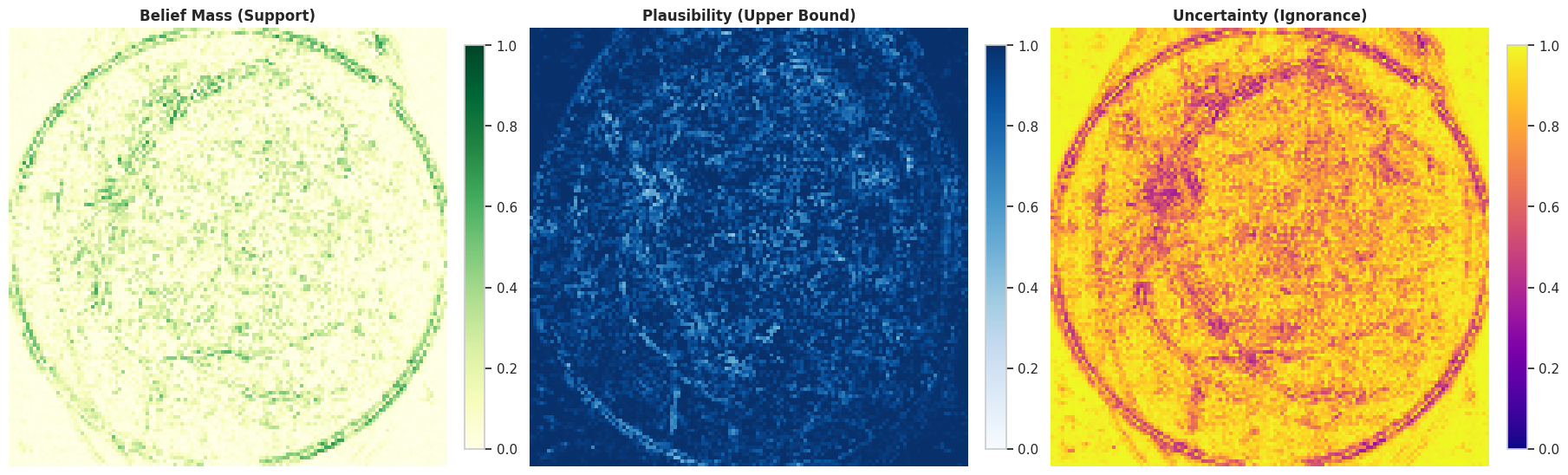}
        \caption{}
    \end{subfigure}
    \hfill 
    \begin{subfigure}[b]{\textwidth}
        \centering
        \includegraphics[width= 0.5 \textwidth]{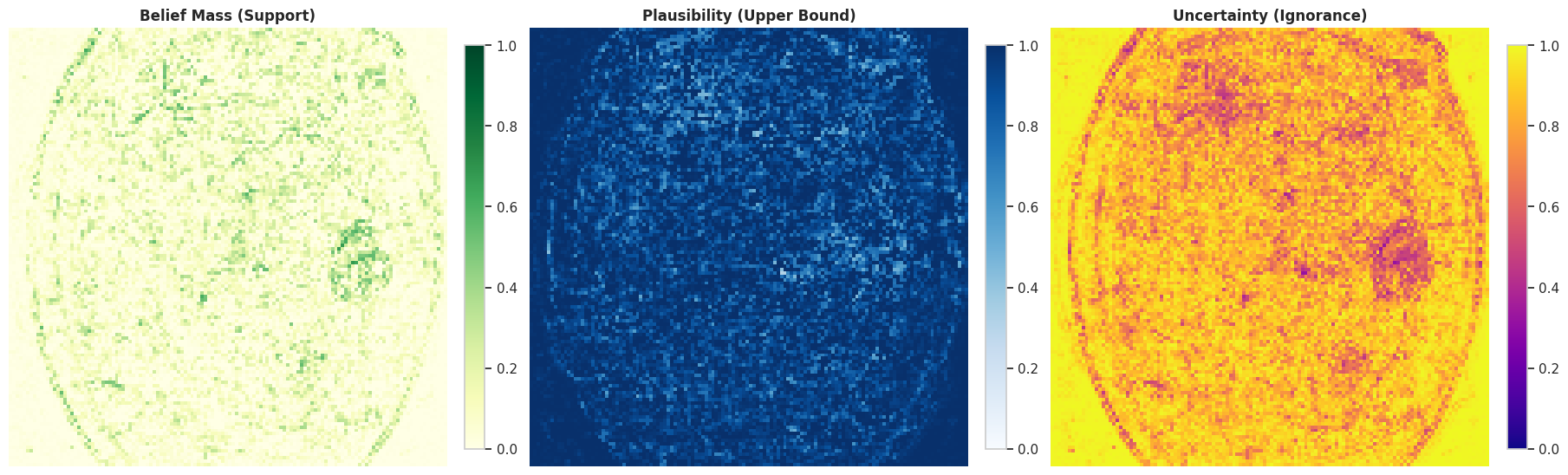}
        \caption{}
    \end{subfigure}
    \begin{subfigure}[b]{\textwidth}
        \centering
        \includegraphics[width= 0.5 \textwidth]{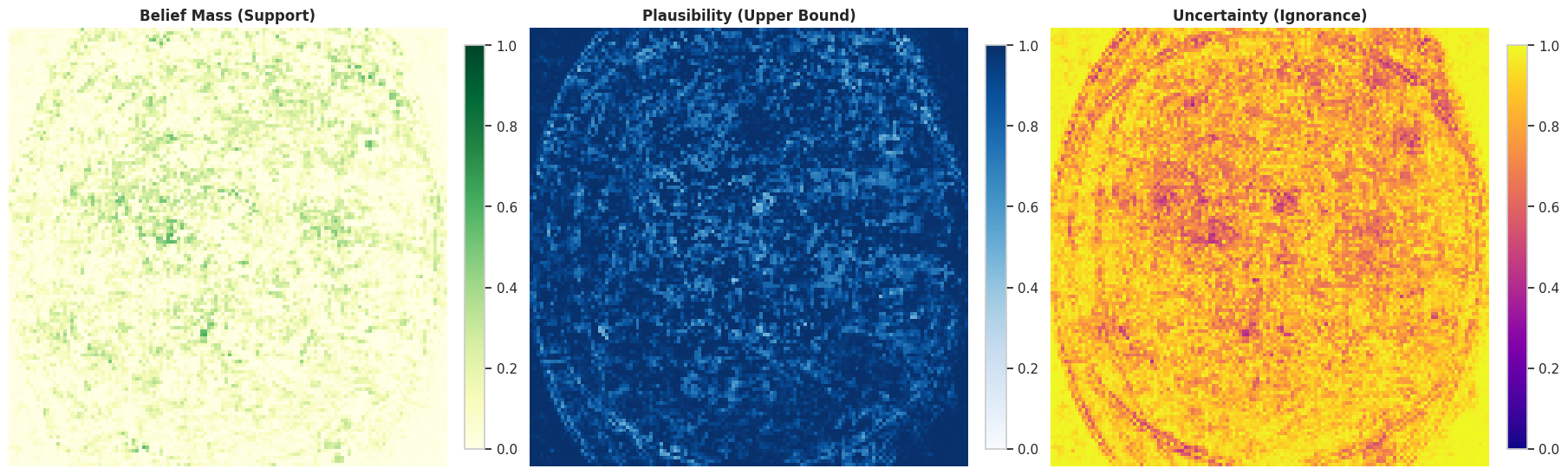}
        \caption{}
    \end{subfigure}
    \begin{subfigure}[b]{\textwidth}
        \centering
        \includegraphics[width= 0.5 \textwidth]{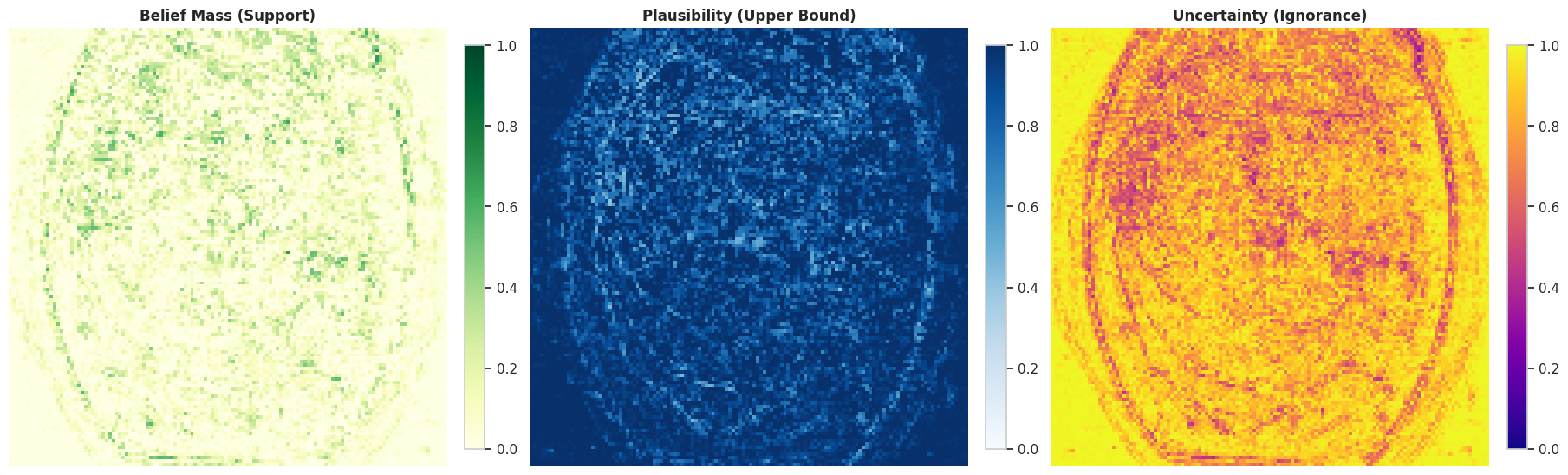}
        \caption{}
    \end{subfigure}
    \begin{subfigure}[b]{\textwidth}
        \centering
        \includegraphics[width= 0.5 \textwidth]{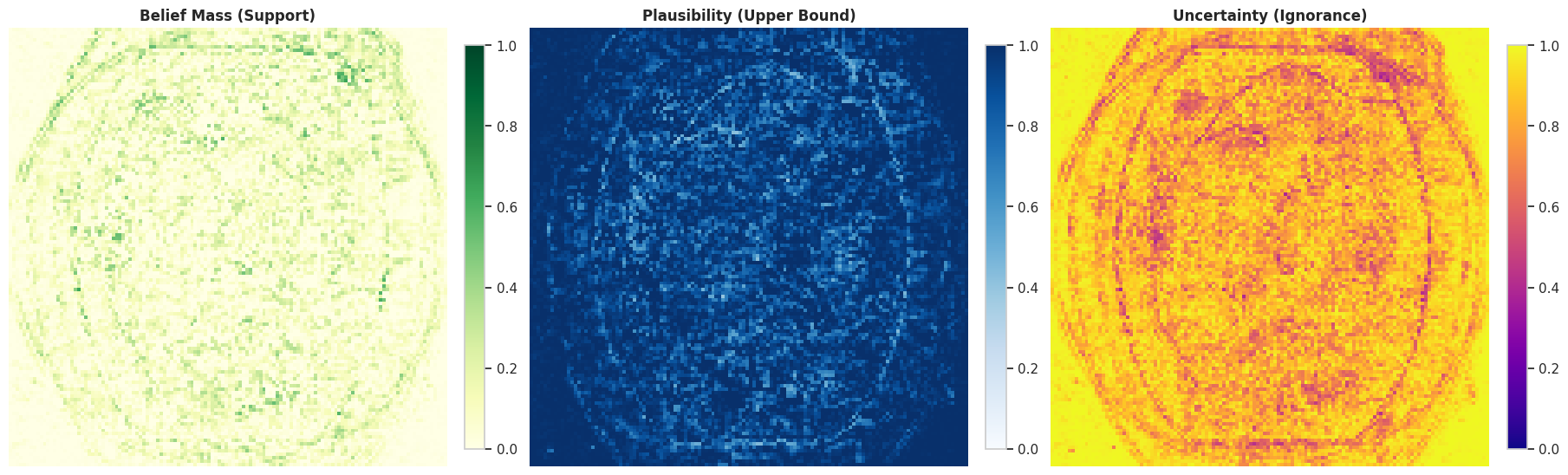}
        \caption{}
    \end{subfigure}
    \caption{This figure illustrates the Dempster-Shafer fusion of model explanations for classifying diabetic retinopathy (DR), tracking the evolution of evidential support across five distinct severity levels. (a) Healthy: The fusion maps for healthy retinas exhibit minimal belief mass (pale/empty) and high, uniform uncertainty (bright yellow). This indicates that the ensemble’s decision is driven by the absence of pathological lesions rather than the detection of specific features. As a result, there is maximum ignorance regarding the background tissue. (b) Mild DR: In contrast, the mild DR sample reveals a distinct, localized cluster of belief mass (green), paired with a sharp reduction in uncertainty (dark purple). This suggests that the models successfully identified a specific, solitary lesion, likely a microaneurysm, acting as strong evidence of early-stage disease. (c) Moderate DR: As the disease progresses, the belief mass becomes more scattered and diffuse, reflecting the dispersed nature of hemorrhages. The uncertainty map is mottled (orange/yellow), indicating that, although the ensemble detects pathology, the evidence is less concentrated spatially than the solitary lesion in the mild case. (d) Proliferative DR: The proliferative stage exhibits widespread texture-based belief patterns without a focal point. This is consistent with the detection of neovascularization yet maintains moderate epistemic uncertainty across the retina. (e) Severe DR: The severe DR maps display a complex network of belief mass that aligns with vascular structures. The corresponding uncertainty map reveals darker linear patterns, confirming that the ensemble identifies widespread vascular abnormalities as concrete evidence. This reduces local ignorance compared to the healthy baseline.}
    \label{fig:diabetic_ubiqcon}
\end{figure}

\begin{figure}[]
    \centering
    \begin{subfigure}[b]{\textwidth}
        \centering
        \includegraphics[width= 0.45 \textwidth]{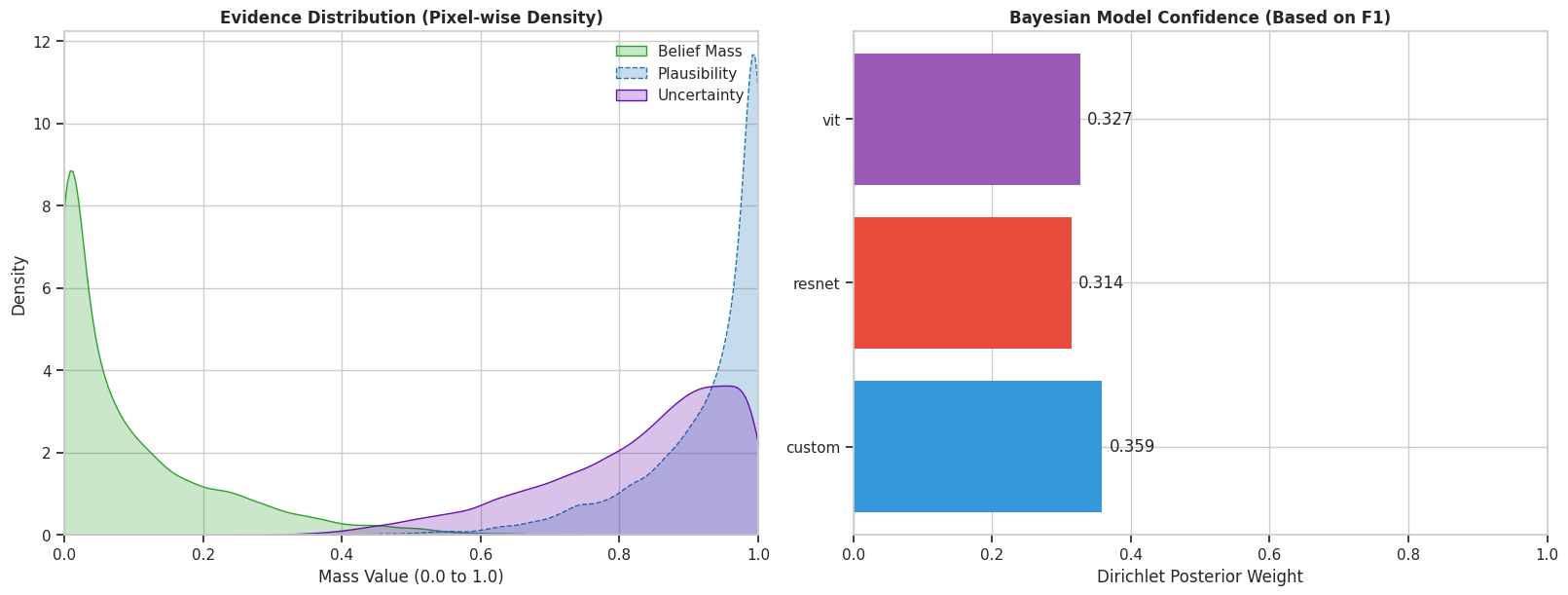}
        \caption{}
    \end{subfigure}
    \hfill 
    \begin{subfigure}[b]{\textwidth}
        \centering
        \includegraphics[width= 0.45 \textwidth]{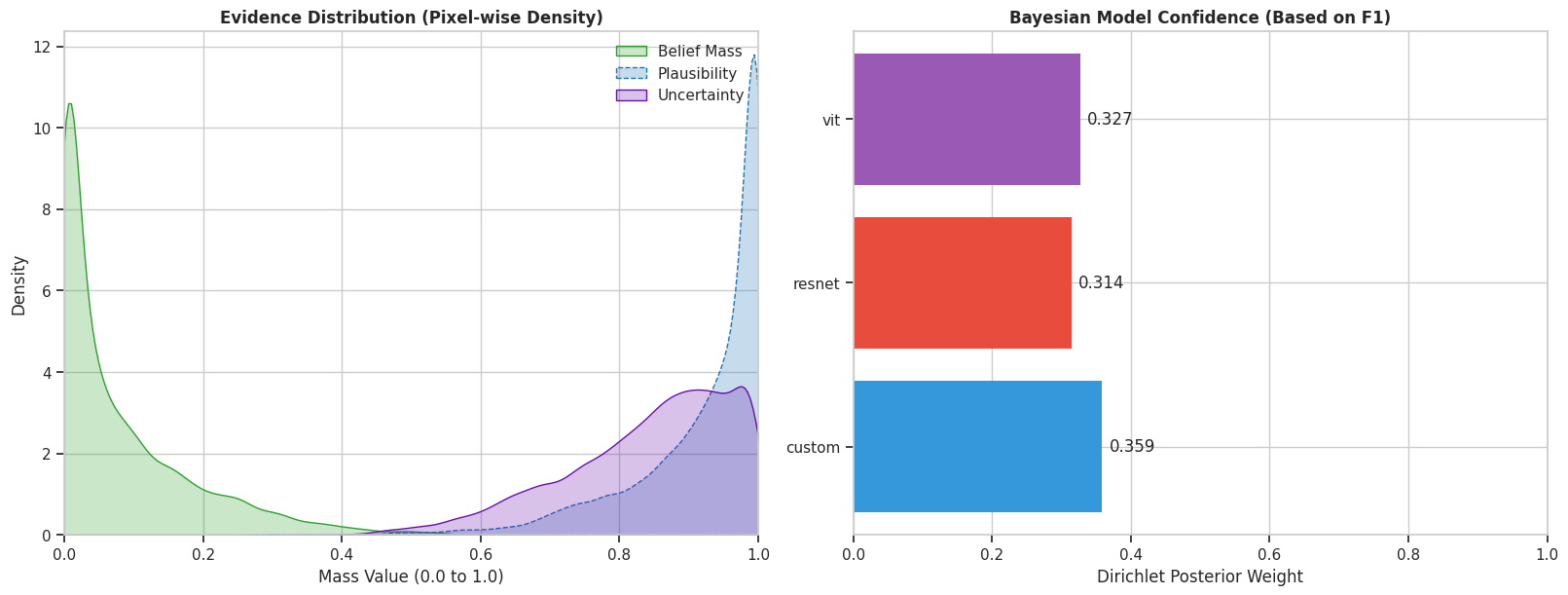}
        \caption{}
    \end{subfigure}
    \begin{subfigure}[b]{\textwidth}
        \centering
        \includegraphics[width= 0.45 \textwidth]{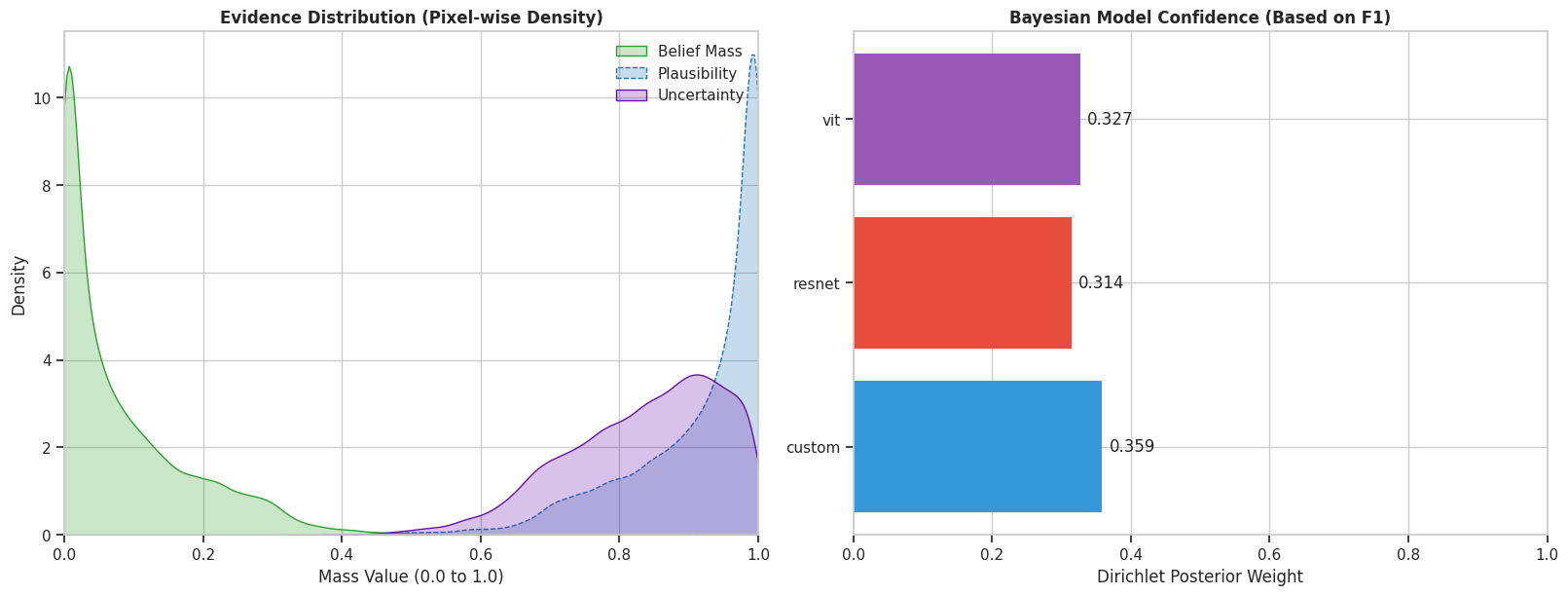}
        \caption{}
    \end{subfigure}
    \begin{subfigure}[b]{\textwidth}
        \centering
        \includegraphics[width= 0.45 \textwidth]{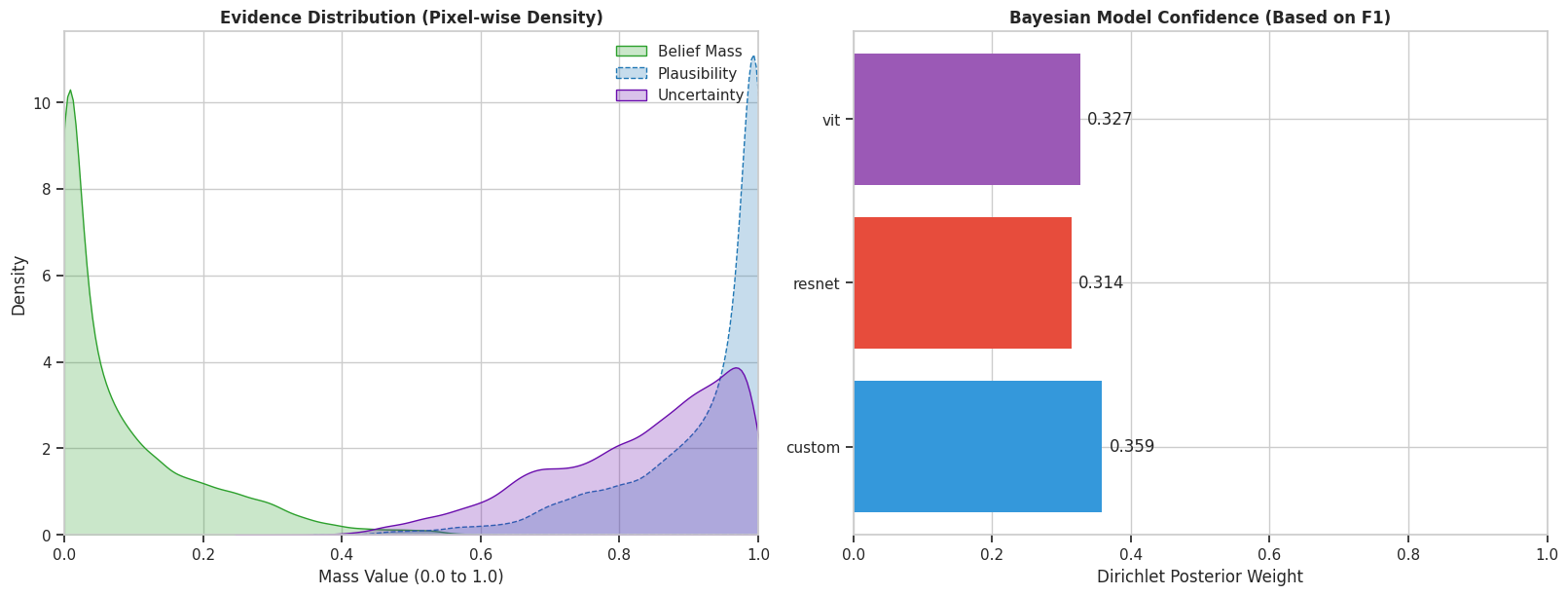}
        \caption{}
    \end{subfigure}
    \begin{subfigure}[b]{\textwidth}
        \centering
        \includegraphics[width= 0.45 \textwidth]{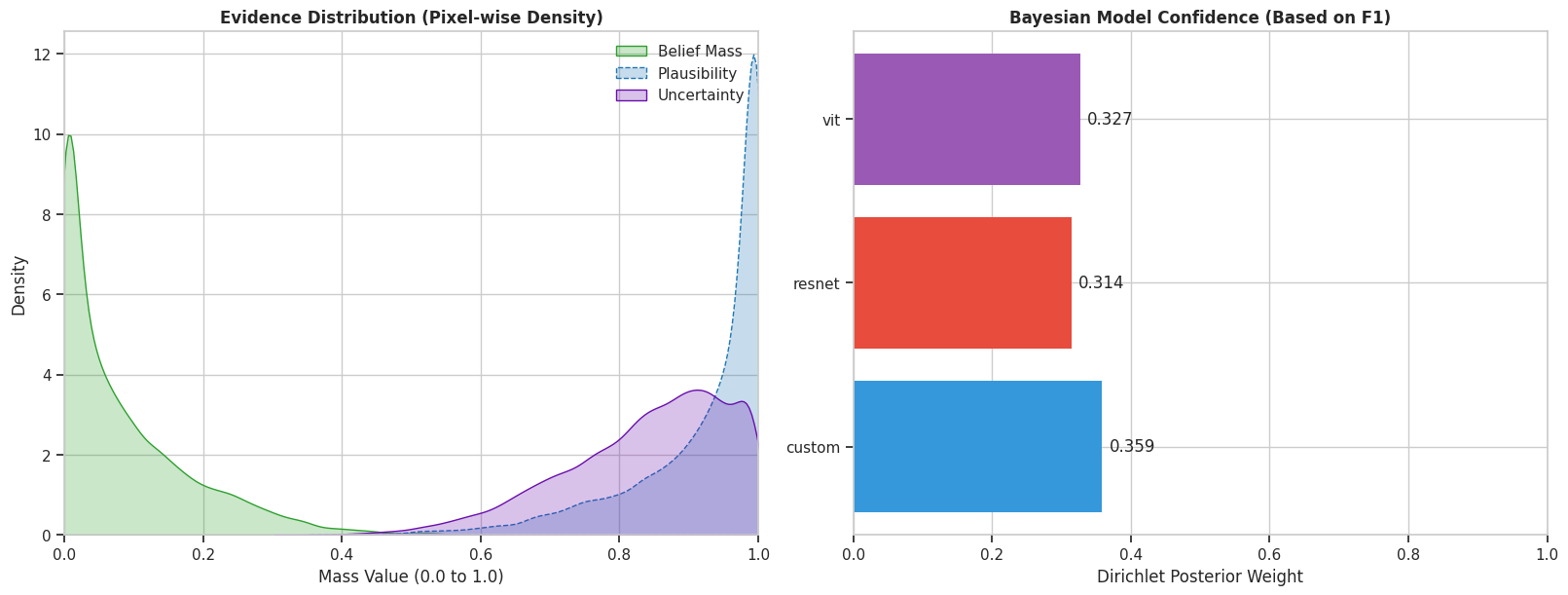}
        \caption{}
    \end{subfigure}
    \caption{This figure shows the fusion maps of SHAP explanations produced by a Bayesian-weighted model ensemble (Custom CNN, ResNet, and ViT) at different levels of diabetic retinopathy (DR) severity. The individual attribution maps demonstrate that advanced disease states, such as severe and proliferative DR, result in dense, widespread positive contributions (red pixels) across the various architectures. In contrast, healthy and mild stages exhibit sparser, more diffuse explanatory patterns. The Bayesian model consistently assigns the highest weight to the Custom CNN ($w \approx$ 0.36), indicating its dominant influence on the fusion result. The resulting fusion maps and quantitative evidence distributions reveal a distinct trend: advanced DR samples display localized clusters of belief mass that align with pathological features. This leads to a marked reduction in epistemic uncertainty (purple areas in ignorance maps). In contrast, healthy and early-stage samples are characterized by negligible belief mass and high, uniform uncertainty (bright yellow), as demonstrated by uncertainty density plots heavily skewed toward 1.0. In conclusion, the fusion framework effectively differentiates the nature of the ensemble's predictions. Decisions for healthy cases rely on high epistemic ignorance (the absence of evidence), whereas advanced disease detections are supported by concrete, localized evidence. This provides a crucial layer of trustworthiness for clinical interpretation.}
    \label{fig:diabetic_analytics}
\end{figure}
\begin{figure}
    \centering
    \includegraphics[width=0.5\linewidth]{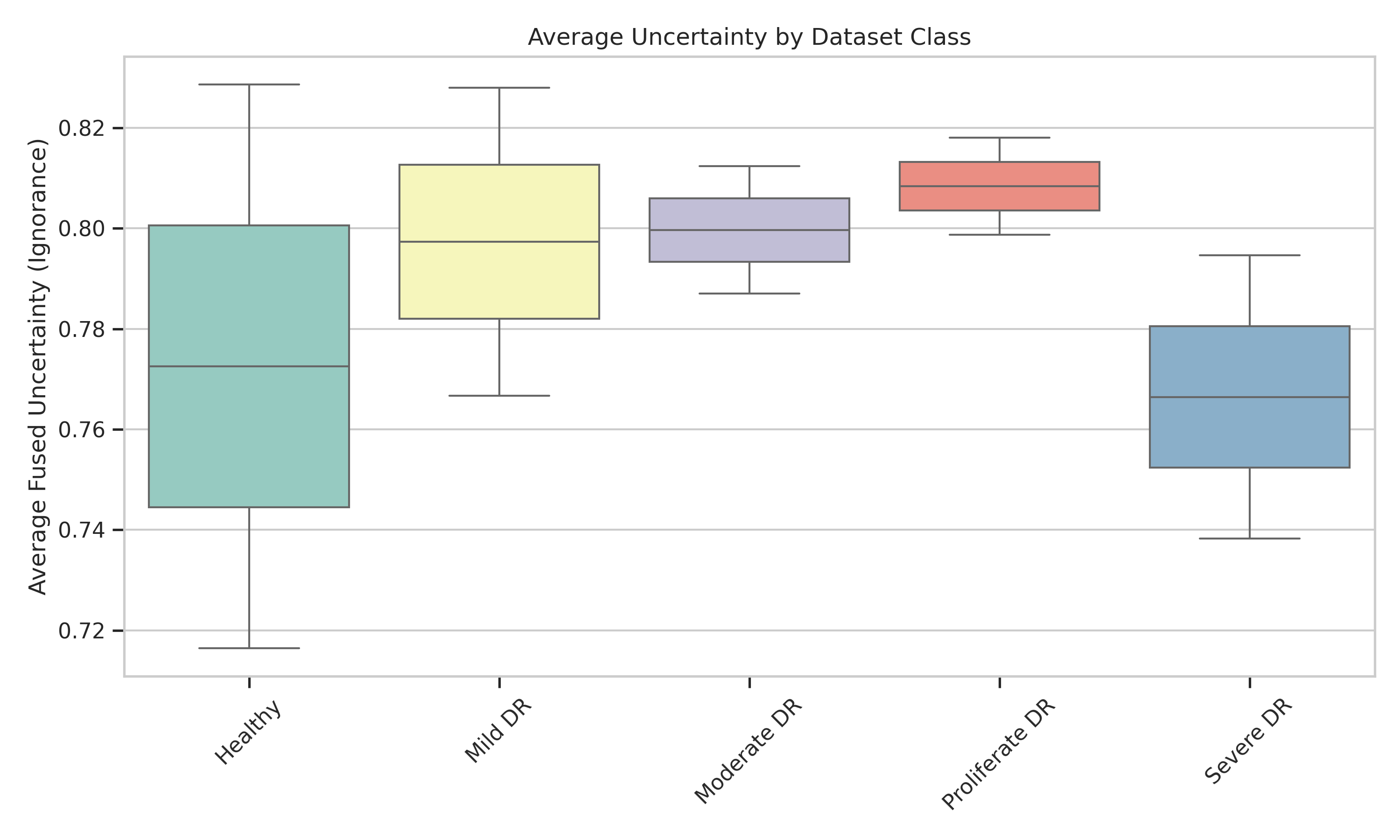}
    \caption{This figure denotes a box plot allowing us to analyze the mean average Uncertainty (Ignorance) for the Diabetic Retinopathy dataset. The Healthy class has the widest spread of average fused Uncertainty (Ignorance) followed by the SeverDR and MildDR. The Proliferate DR has the lowest spread with highest Uncertainty (Ignorance). The Moderate DR has second lowest average fused Uncertainty (Ignorance).}
    \label{fig:diabetic_error_per_class}
\end{figure}
\begin{figure}
    \centering
    \includegraphics[width=0.5\linewidth]{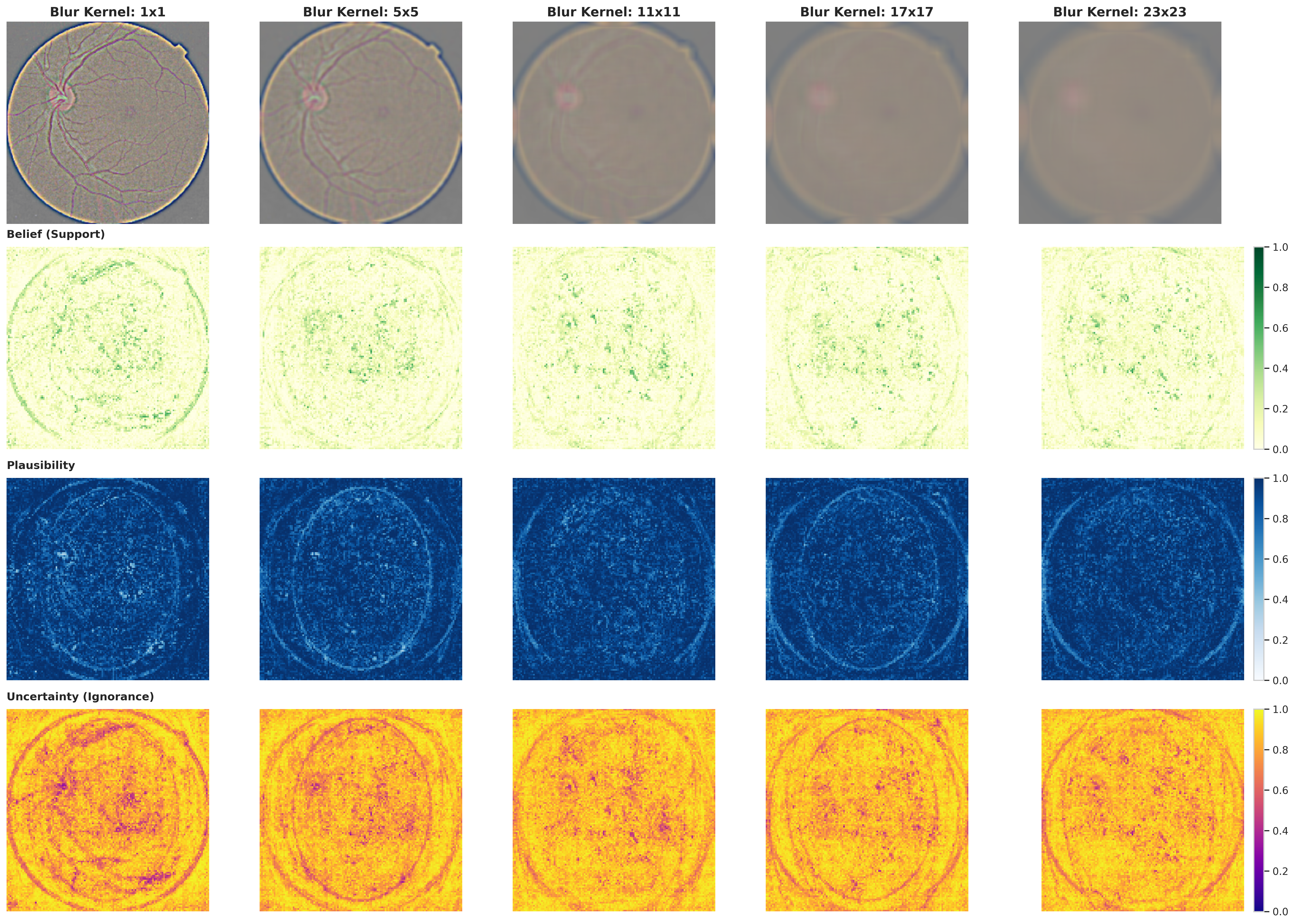}
    \caption{The visual illustration of the framework's  for progressive structural degradation through the introduction of the Gaussian blur. The top row of the figure displays the input image at increasing blur kernel sizes, ranging from 1x1 to 23x23. Subsequent rows illustrate the corresponding Belief (Support), Plausibility, and Uncertainty (Ignorance) maps generated by the DS-SHAP fusion framework for the Diabetic Retinopathy dataset. The increase in the Gaussian kernel increases the uncertainty and the boundaries start to fade away leading to high Uncertainty (Ignorance) denoted by progressing yellow color as the blur progresses for the crucial images of the retina.}
    \label{fig:diabetic_ablation_study}
\end{figure}
\begin{figure}
    \centering
    \includegraphics[width=0.5\linewidth]{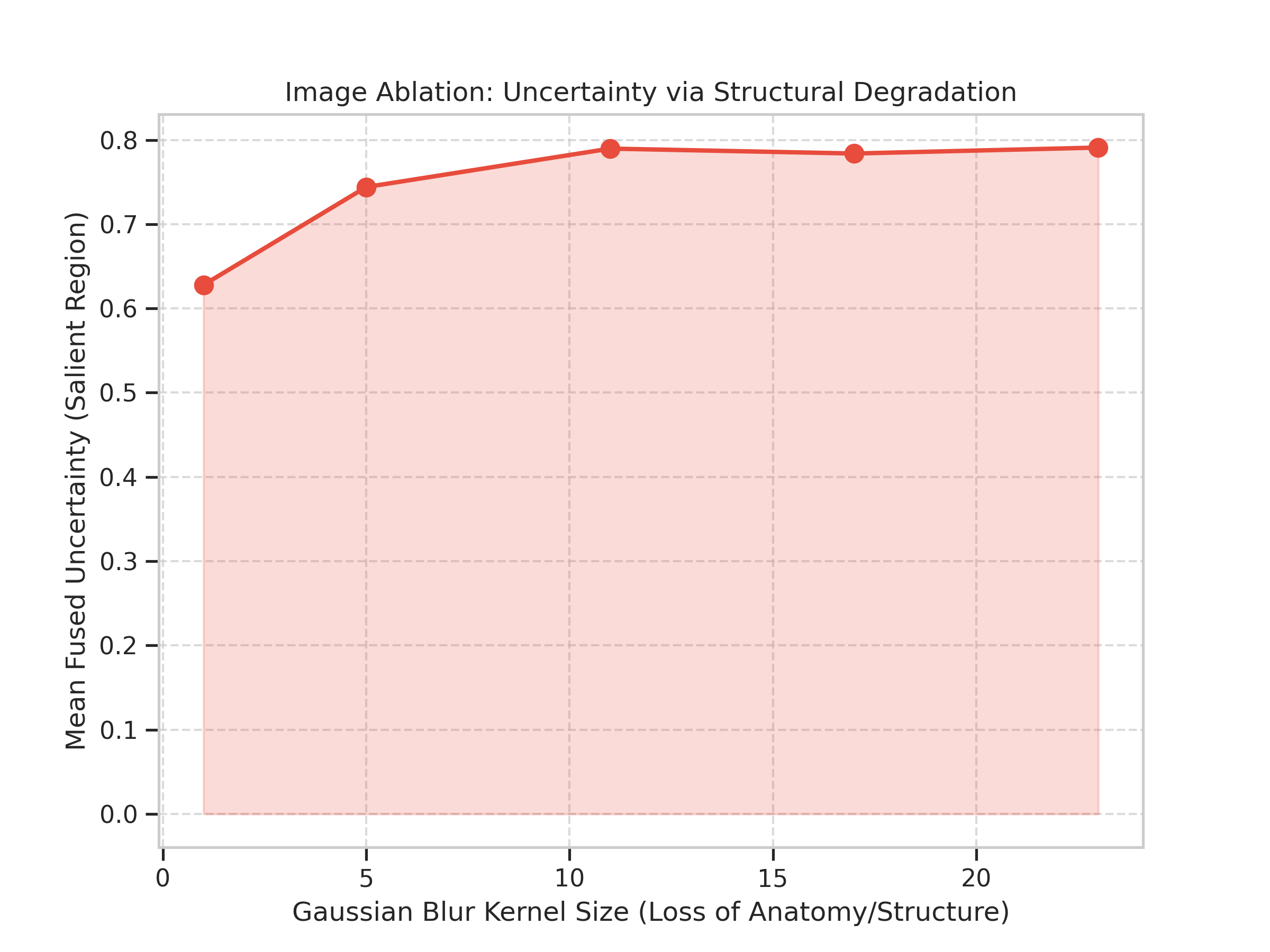}
    \caption{This figure illustrates the impact of structural degradation, manifested through the incorporation of a Gaussian blur kernel, on the mean fused uncertainty. The augmentation of the Gaussian blur kernel has been demonstrated to enhance the mean fused uncertainty for the Diabetic Retinopathy dataset. The following line chart is intended to quantify the relationship between structural integrity and the framework's calculated epistemic uncertainty. The x-axis of this figure represents the increasing Gaussian blur kernel size, or the loss of anatomical detail, while the y-axis shows the mean fused uncertainty, or the degree of ignorance, measured exclusively within the salient anatomical regions.}
    \label{fig:ablation_uncertainty_chart_diabetes}
\end{figure}
\begin{figure}
    \centering
    \includegraphics[width=0.5\linewidth]{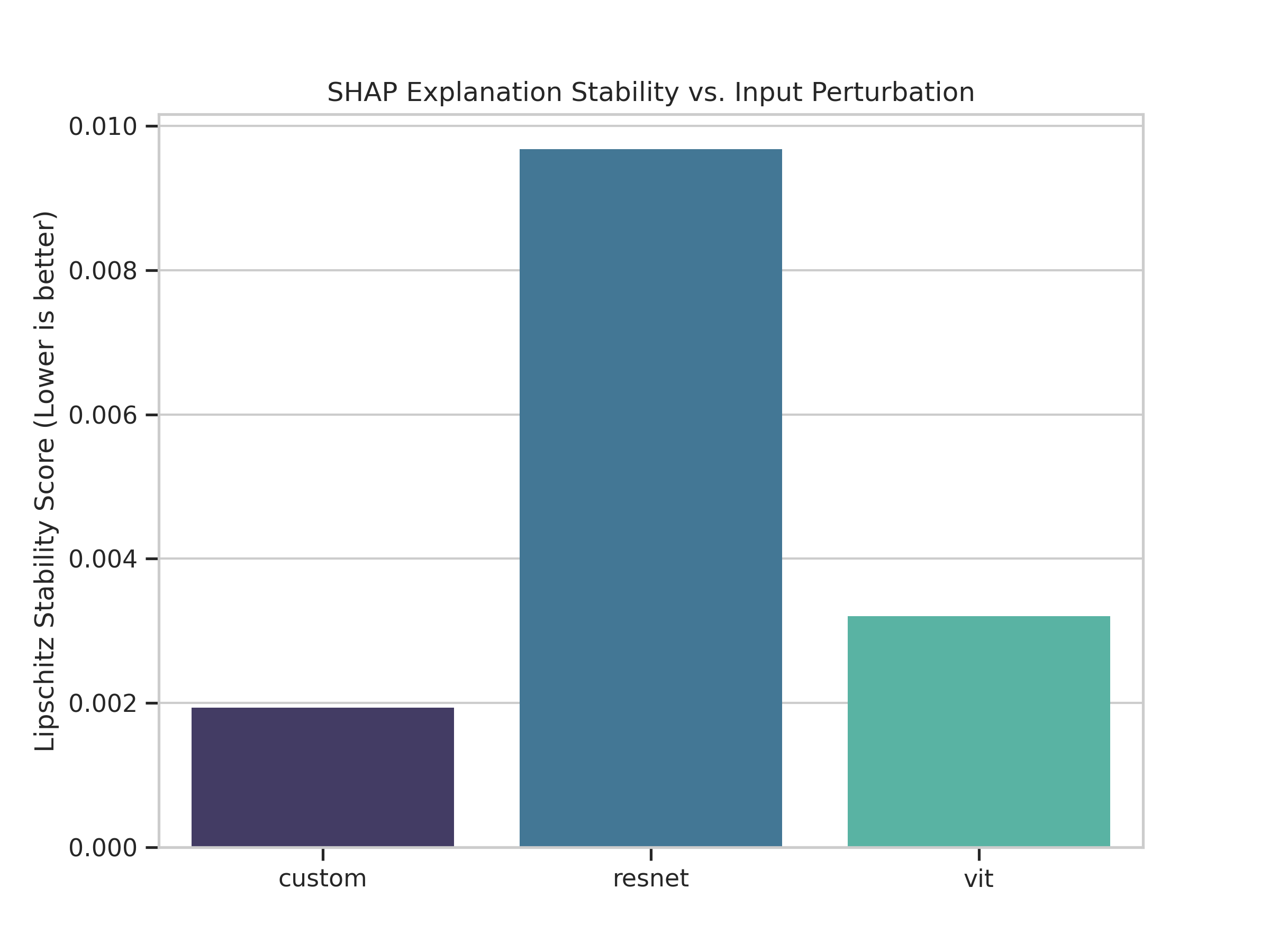}
    \caption{This figure indicates the stability of the SHAP attributions for the architectures employed in the analysis of images from the Diabetic Retinopathy dataset. The custom CNN demonstrates the lowest Lipschitz score (lower the better), indicating the least stable attributions in comparison to the ViT and ResNet architectures. However, the scores are low, which indicates that the attribution which is used in our framework to quantify uncertainty is overall stable.}
    \label{fig:diabetic_lipschitz_shap}
\end{figure}
\begin{figure}
    \centering
    \includegraphics[width=0.5\linewidth]{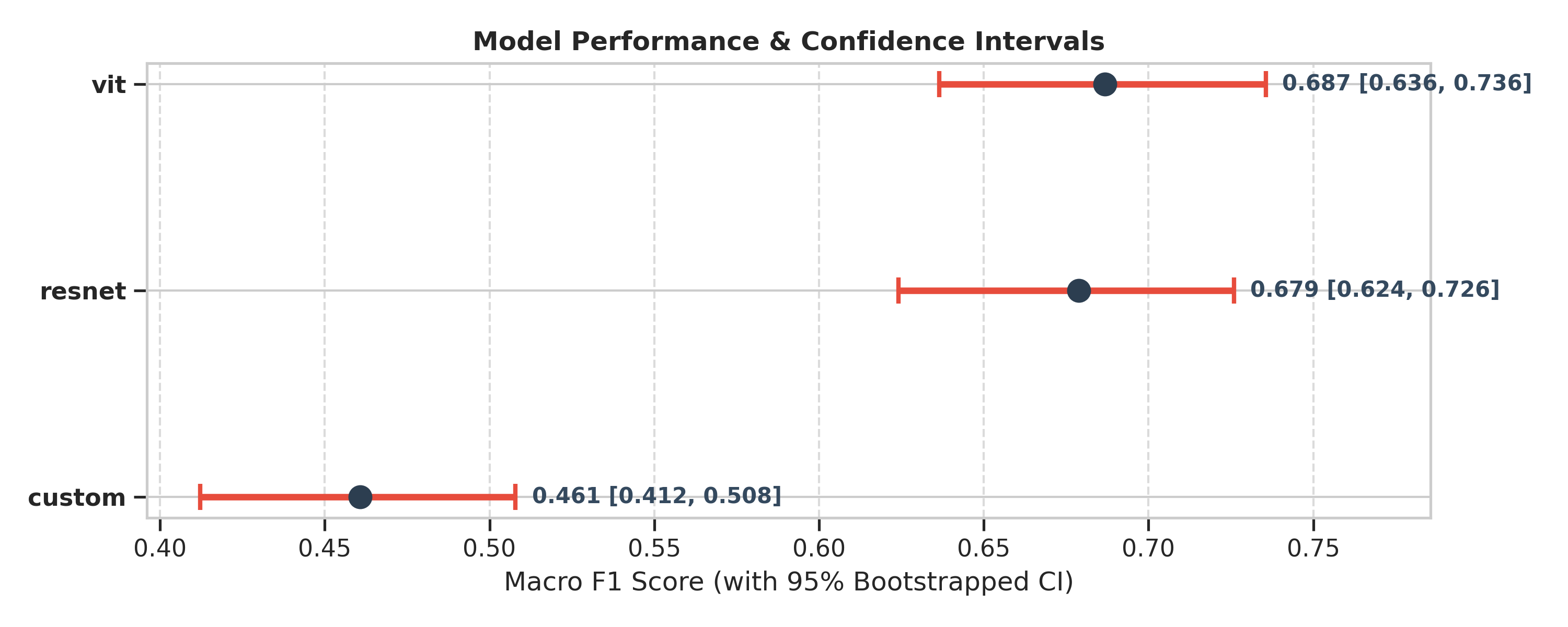}
    \caption{Forest plot with the mean macro F1 performance of each architectures (ResNet, CustomCNN, and ViT) for the Diabetic Retinopathy (DR) dataset along with the 95\% confidence interval (CI). The dot represents the mean performance while the red bars denote the 95\% CI interval.}
    \label{fig:confidence_interval_diabetic}
\end{figure}
To validate the efficacy of the framework in ophthalmological diagnostics, a comprehensive evaluation was conducted on the Diabetic Retinopathy (DR) dataset. The evaluation focused on the system's ability to localize retinal pathologies across varying severity levels—ranging from Healthy to Proliferate DR-while rigorously quantifying the epistemic uncertainty inherent in the models. The Bayesian Meta-Learning mechanism exhibited its aptitude for discerning architectural suitability for retinal imaging. Across all severity classes, the Dirichlet posterior consistently assigned the highest reliability weight to the custom CNN architecture ($w = 0.36 - 0.37$), thereby identifying it as the primary expert. Conversely, the ResNet and VIT models were assigned lower weights, with $w$ ranging from 0.31 to 0.33. A visual examination of the feature attribution maps substantiates the validity of the established ranking. The custom model consistently yielded cleaner, anatomically pertinent attributions, while the ResNet model exhibited high-frequency noise, or "speckling," across the fundus. In contrast, the ViT model produced blockier, less precise activations. By dynamically down-weighting these noisier contributors, the framework prevented the ensemble from being corrupted by architectural artifacts, effectively filtering out stochastic error and reinforcing the signal-to-noise ratio of the ensemble's collective decision. For eyes exhibiting diabetic retinopathy, the framework demonstrated a progression of evidential confidence that correlated with disease severity. In the Severe and Proliferate DR samples (Figure:~\ref{fig:diabetic_shap}), the expert models successfully localized key pathological biomarkers, with positive SHAP attributions ($\phi > 0$, red pixels) clustering tightly around hemorrhages, microaneurysms, and exudates—the clinical hallmarks of the disease~\citep{asha2021detection}. The custom model, in particular, exhibited a high degree of specificity for these lesions, effectively ignoring the healthy retinal background. This feature-specific localization is mathematically synthesized in the Dempster-Shafer fusion maps (Figure:~\ref{fig:diabetic_ubiqcon}), where the Belief Mass ($\text{Bel}$) map displays scattered clusters of green evidence mirroring the distribution of exudates. This finding indicates that the model does not depend on a single global descriptor. Rather, it integrates local evidence from multiple lesion sites to formulate a comprehensive diagnosis. Additionally, the Uncertainty Map ($U$) serves as a crucial safety check. In the mild DR case, the uncertainty remains elevated across the retina compared to severe cases, reflecting the subtle nature of early-stage lesions. In all scenarios, the boundary between the retina and the black background is clearly delineated by a sharp transition to maximum uncertainty (Bright Yellow), thereby confirming that the model correctly identifies the limits of its valid input domain and avoids the hallucination of features in the void. The evaluation of healthy samples underscores the framework's capacity to substantiate the absence of pathology through anatomical verification. The attribution maps demonstrate that the models prioritize the optic disc and major blood vessels, which are the most salient features of a healthy eye. The custom model accentuates the optic disc with positive attribution, effectively interpreting the presence of a normal optic disc as evidence that is inconsistent with disease. However, the Plausibility Map ($\text{Pl}$) (Figure:~\ref{fig:diabetic_ubiqcon}) reveals a layer of nuanced internal disagreement. While the overall plausibility of "Healthy" remains high (Dark Blue), the map contains lighter blue speckles—a direct result of the ResNet model's noise. In instances where the noisy model hallucinated potential lesions, it introduced minor amounts of "Evidence Against," thereby lowering the plausibility ceiling. By conceptualizing this disagreement rather than suppressing it, the proposed framework offers a lucid perspective on model disagreement, a feature that surpasses the conventional black-box ensembles, which merely average out dissenting elements. These findings underscore the substantial clinical relevance of the framework for automated screening, particularly in addressing the issue of overconfidence exhibited by standard deep learning models when confronted with out-of-distribution data. Our framework mitigates this through the explicit modeling of Ignorance ($U$), allowing the system to distinguish between a "Healthy" diagnosis (High Belief, Low Uncertainty) and a "Poor Quality / Unknown" scan (Low Belief, High Uncertainty). In the context of a clinical workflow, images characterized by high aggregate uncertainty can be automatically flagged for manual review by an ophthalmologist. This practice prevents potential false diagnoses caused by suboptimal lighting conditions or the presence of artifacts. Furthermore, by decoupling Belief (confirmed lesions) from Plausibility (potential lesions), the system facilitates clinicians' assessment of the diagnostic risk level with enhanced interpretability. A case with Moderate Belief but High Plausibility suggests that while the AI is not 100\% sure, it cannot rule out the disease, an important distinction for early intervention that could redefine safety protocols in computer-aided diagnosis.\\
\\
The analytical charts demonstrate (Figure:~\ref{fig:diabetic_analytics}) the analytics for the Diabetic Retinopathy dataset, thereby providing quantitative validation of the framework's pixel-wise maps and confirming the efficacy of the ensemble strategy. The Evidence Distribution Kernel Density Estimate (KDE) plot reveals the probability density functions for the three epistemic metrics, where the Belief Mass (Green Curve) peaks sharply at 0.0 with a long, thin tail extending toward 1.0. This distribution is biologically accurate for retinal imaging because pathological features like microaneurysms or exudates occupy only a minute fraction of the total surface area. The sharp peak confirms that the model correctly identifies the majority of the retina as lacking specific evidence, while the tail indicates localized high-confidence detections. Concurrently, the Uncertainty Mass (purple curve) exerts a dominant influence over the distribution, with a peak near 1.0, thereby validating the architecture. In this model, the majority of healthy tissue and the black background are considered unknowns, thus distinguishing "no evidence" from "negative evidence." The Plausibility (Blue Dashed Curve) reflects this uncertainty, confirming the mathematical relationship where high plausibility is maintained in the absence of contradictory proof. In addition, the Bayesian Model Confidence bar chart displays the Dirichlet Posterior Weights. The custom CNN invariably achieves the highest ranking ($w = 0.36$), in comparison to ResNet ($w = 0.31$) and ViT ($w = 0.32$). This non-uniform but balanced weighting confirms that while the custom model acts as the primary feature extractor, the system automatically penalizes the noise introduced by weaker models, ensuring a robust ensemble that avoids overfitting to specific architectural biases. These statistical aggregations offer a global summary that corresponds directly to the local explanations found in the spatial maps. The Bayesian Confidence chart dictates the influence of the feature attribution maps; specifically, the custom model's clear highlighting of optic discs or lesions is mathematically prioritized over the noisy speckling of the ResNet model during the fusion step. This weighting mechanism guarantees that the resulting fused maps exhibit a structural similarity to the outputs of the custom model, thereby enabling the discerning assessment of experts to prevail over the potentially more voluminous yet less precise contributions of less experienced professionals. Additionally, the Evidence Distribution chart effectively functions as a histogram for the fusion maps. The visual \textit{Green Spots} of confirmed pathology in the Belief Map correspond to the tail of the green density curve, while the vast \textit{White Background} aligns with the massive peak at zero. Analogously, the \textit{Bright Yellow} background of the Uncertainty Map is represented by the purple peak at 1.0, and the \textit{Dark Purple} anatomical regions correspond to the dip in the curve. The \textit{Dark Blue} regions of the Plausibility map correspond to the peak at 1.0, with lighter blue areas of conflict resulting from model disagreement manifesting as deviations in the curve. Furthermore, the analytics charts (Figure:~\ref{fig:diabetic_analytics}) reveal three critical system behaviors. Firstly, the precipitous decline of the Belief curve signifies a high degree of specificity for retinal lesions, thereby demonstrating that the model circumvents the "hallucination" of pathology across the retina and reserves high belief exclusively for specific pixels. This trait is paramount for minimizing false positives during screening. Secondly, the prevalence of the Uncertainty curve underscores the utilization of explicit ignorance modeling. This distinguishes the framework from conventional Softmax models, which employ arbitrary partitioning of background space. In contrast, the framework acknowledges that the majority of the image does not offer substantial information, thereby ensuring that diagnoses are predicated exclusively on detected anatomical features. Finally, the balanced Bayesian weights demonstrate ensemble robustness, thereby confirming that the system relies on the collective intelligence of the group rather than a single, potentially fallible architecture. This ensures a consensus-based decision process.
The DR analysis is a challenging problem in the field of computer vision. We studied the predictive performance and performed ablation studies along with Lipschitz~\citep{demertzis2021lipschitz,simpson2024probabilistic} attribution scores stability calculation. This problem statement allows our framework to study the robustness of our framework when one of the models fail to capture the complex patterns in the dataset. The predictive performance of the architectures are as mentioned which are calculated using 10-fold stratified cross validation (See Figure:~\ref{fig:confidence_interval_diabetic}). The custom CNN had mean Macro F1-Score of 46.1\% with 95\% CI ranging between 41.2\% \& 50.8\% whereas the ViT and ResNet had mean F1-Scores of 68.7\% \& 67.9\% with 95\% CI ranging between 63.6\% \& 73.6\% for Vit and 62.4\% \& 72.6\& for the ResNet respectively. This indicates that the Custom CNN was not able to learn the complex patterns whereas the ViT \& ResNet were able to detect the patterns attributed to their complex architectures which are able to identify the scattered biomarkers of the DR attributing to their complex hierarchical feature extraction capabilities.
The reliability of the framework's mathematical foundation depends on the actual stability of the SHAP attributions. The SHAP explanation stability chart (See Figure:~\ref{fig:diabetic_lipschitz_shap}) allows us to understand that the Custom CNN technically achieved the lowest Lipschitz attribution score of approximately 0.002 while the ResNet has attribution score of approximately 0.010 with ViT scoring approximately 0.004. The low score of the custom CNN could be attributed to it's inability of identifying the complex patterns in the dataset. The low score of ViT could be attributed to the ViT's self-attention mechanism which produces spatial attributions that are more resistant to input noise as compared to convolutional filters of the ResNet. 
Further, the ablation study allows us to analyze the quantitative as well as qualitative observations related to the stability of our framework (See Figure:~\ref{fig:diabetic_ablation_study}\&~\ref{fig:ablation_uncertainty_chart_diabetes}). The DR diagnosis is heavily reliant on the profound vascular boundary and tiny lesions. As the Gaussian blur kernel increases from the baseline of 1x1 to 11x11 and further to 23x23, the average fused uncertainty increases from 0.63 to 0.79. The visual ablation study provides the qualitative observation. For the baseline kernel of 1x1, the Belief map clearly traces the vascular regions of the retina. As the blur increases, the vascular structures start to dissolve entirely. Consequently, the Belief and Plausibility maps fade to dark, empty arrays, while the Uncertainty (Ignorance) map explicitly floods the retinal region with bright yellow, mathematically reflecting the models' blindness to the necessary diagnostic structures.
From the the Uncertainty (Ignorance) per class analysis (See Figure:~\ref{fig:diabetic_error_per_class}), the framework exhibits the highest median uncertainty of approximately of 0.81 for Proliferate DR, follows closely by Mild DR and Moderate DR. This could be attributed to the highly ambiguous, subtle, or new forming vascular anomalies that frequently cause the disagreement. Conversely, the framework was most confident (lowest median uncertainty at approximately 0.765) when diagnosing Severe DR. This reflects the clinical reality where the Severe DR is characterized by massive, obvious hemorrhages and extensive ischemia~\citep{murugesan2015thrombosis}. The framework correctly curtails its uncertainty when presented with overwhelming, undeniable pathological evidence.
\subsection{Runtime Complexity}
\begin{figure}
    \centering
    \includegraphics[width=0.5\linewidth]{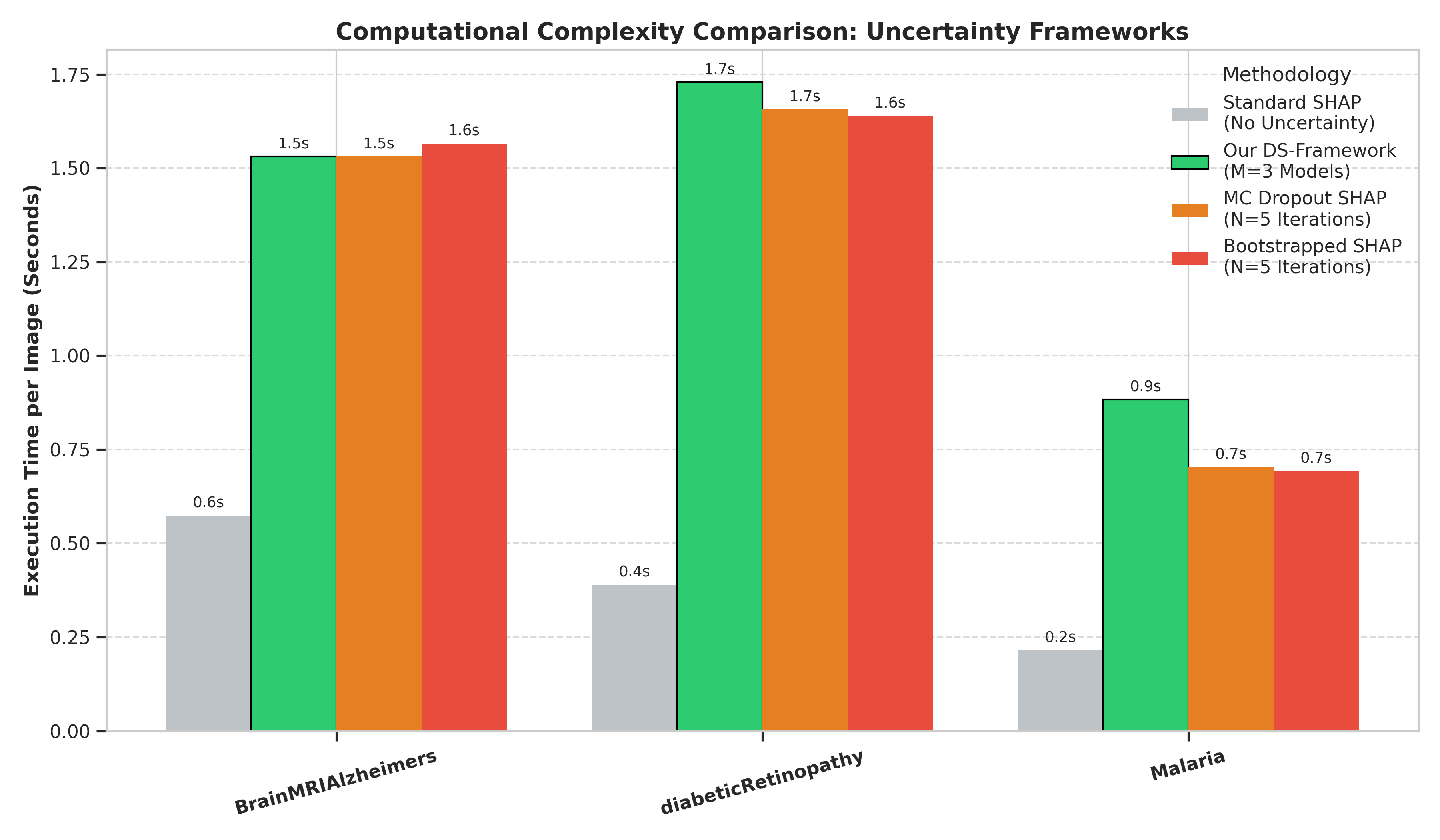}
    \caption{The bar chart provides a quantitative comparison of the computational execution time (in seconds per image) across three distinct medical imaging datasets: brain MRI, diabetic retinopathy, and malaria. The chart indicates the execution speed of Standard SHAP (baseline with no uncertainty quantification), the proposed Dempster-Shafer (DS) Framework, Monte Carlo (MC) Dropout SHAP, and Bootstrapped SHAP.}
    \label{fig:runtime_comparison}
\end{figure}

The deployment of a framework in real-world settings is contingent upon the fulfillment of several critical requirements, among which is the assurance of computational efficiency. In order to address and study the computational efficiency, the present framework was compared with standard SHAP, Monte Carlo (MC) Dropout SHAP, and Bootstrapped SHAP (See Figure:~\ref{fig:runtime_comparison}). This analysis was performed across the three datasets utilized in this research, which encompass different modalities. As anticipated, the standard SHAP model demonstrates the shortest execution time among all the frameworks, requiring a single pass through a single model. However, the standard SHAP does not provide substantial information regarding the uncertainty. This could be inadequate for high-stakes medical imaging tasks, where ascertaining when the model is uncertain is as crucial as the prediction itself. In addition, the proposed framework explanations from an ensemble of models ($M$) with distinct architectures demonstrated a computational runtime similar to that of the MC Dropout SHAP for Brain MRI Alzheimer's and Diabetic Retinopathy datasets. In the context of the Malaria dataset, our proposed framework emerged as a substantially more computationally intensive solution in comparison to the competing approaches. However, the mathematical overhead introduced by our framework is low. It is noteworthy that the competing uncertainty methodologies (MC Dropout and Bootstrapped SHAP) in this benchmark were constrained to a minimal iteration count of $N=5$. In actual clinical deployments, the generation of stable, reliable variance maps using MC Dropout or Bootstrapping typically requires between 10 and 50 iterations to reach convergence.
Given that the time complexity of these standard methods scales linearly with $N (O(NC))$, a clinically robust MC Dropout implementation would realistically require between $3.0$ and $15.0+$ seconds per image. In contrast, our proposed framework demonstrates a notable degree of robustness with respect to epistemic uncertainty, achieving this in a single pass of $M=3$ models. Consequently, the proposed framework captures richer, cross-architectural uncertainty in a fraction of the time required by standard variance-based methods.
\section{Discussion \& Conclusion}
The proposed framework signifies a paradigm shift in the domain of safety-critical medical AI, unifying three complementary mathematical approaches: Bayesian meta-learning, SHAP, and Dempster–Shafer theory. By conceptualizing model reliability as a Dirichlet random variable, the system establishes a probabilistic architecture that can identify high-performing learners and penalize architectures that demonstrate poor generalization. In contrast to standard softmax-based classifiers, which generate point estimates that masquerade as confidence scores, this framework employs a explicit quantification of uncertainty in medical imaging tasks, including malaria microscopy, diabetic retinopathy, and Alzheimer's disease MRI analysis. In these tasks, pixel-wise belief masses function as implicit weak supervision for localizing pathological biomarkers. The primary benefit of the framework is its ability to provide a more detailed risk assessment than binary classifications. By distinguishing between epistemic uncertainty (model ignorance regarding rare conditions or underrepresented populations) and aleatoric uncertainty (inherent noise in imaging data), clinicians receive actionable insights by making the "unknown unknowns"~\cite{hattab2026humanagencycausalityhuman} visible. In the context of dementia diagnosis, instances characterized by high confidence and low uncertainty are identified for expedited validation, whereas predictions with high uncertainty should be directed for expert review. This paradigm shift in the utilization of AI to support clinical workflows is a significant development in the field.  The employment of pixel-wise belief mass from Dempster-Shafer fusion as a surrogate for weak predictive signals constitutes a pivotal element. This approach facilitates the automatic localization of clinically relevant pathology, thereby enabling more efficient and accurate diagnosis. In the context of diabetic retinopathy, this framework facilitates the reasoning process for validating the reliability of the models and explanations as well. In the domain of malaria microscopy, this property is utilized to accentuate the presence of parasitic inclusions. Additionally, in the domain of Alzheimer's disease MRI, highlights the epistemic uncertainty due to limited availability for training the models.\\
\\
The framework establishes fail-safe by explicitly quantifying total ignorance through uncertainty maps. The presence of high uncertainty values in empty space, the emergence of unexpected tissue artifacts, and the existence of images from unseen imaging protocols impede the model's ability to confidently assert diagnoses in unfamiliar regions. This is of particular value in multi-center studies, where imaging protocols, scanner types, and patient demographics can vary substantially. The framework under consideration differentiates between three distinct categories of shifts: contextual shifts, semantic shifts, and covariate shifts. The term "contextual shifts" is employed to denote alterations in the context of imaging. Semantic shifts are indicative of the presence of unfamiliar pathology. Covariate shifts refer to domain adaptation challenges. The framework facilitates the implementation of robust solutions across heterogeneous clinical environments. The proposed framework is designed to address the recently emerging regulatory requirements stipulated under the EU AI Act~\citep{dubey2024nested}. The EU AI Act mandates transparency, robustness, and explainability in high-risk medical AI systems. The quantification of uncertainty is achieved through the establishment of belief masses, which are then linked to anatomical evidence in order to facilitate predictions. The framework under consideration here fundamentally redefines the human-AI relationship by transforming AI from a binary classifier into a risk assessment framework that communicates confidence alongside predictions.

\section{Limitations}
The Dempster-Shafer Theory and the Dirichlet posterior weighted sampling of the models depend highly on the stability of the SHAP attributions. For the latter, we would like mention that the models should be trained and studied rigorously before our framework is implemented. The framework assumes that the epistemic uncertainty arises from the disagreement or lack of strong evidences from the distinct trained models. However, the modern computer vision models are trained on the same datasets for the same tasks which are further optimized using similar loss functions, leading to a shared blind spot. If an adversarial attack is performed then the output would be identical leading to the unstable \& incorrect SHAP attributions. The framework would struggle to quantify the uncertainty if all the underlying models are homogeneous in terms of the errors. In terms of scalability, one of the limitations that would arise would be due to the extraction of pixel-wise attributions. For complex architectures such as ViT, the cost of single pass is effected by the complexity of the global self-attention mechanisms. Hence, the utilization of ViT with comparatively huge number of hyperparameters would exponentially increase the gradient computation time. One of the solution to this limitations would be to separate the framework from the computationally extensive pixel-space SHAP to localized Grad-CAM or calculating the SHAP within the low-dimension feature space. Scaling to very high-dimensional modalities such as 3D MRI images would need an implementation of dynamically dirven attention maps prior to the fusion. This would allow to calculate the uncertainty fusion of the regions of high importance which contributes highest to the predictions rather than an entire image.
\section{Future Work}
Subsequent research could extend the Dempster-Shafer fusion to multi-modal settings and to longitudinal data. In this manner, belief and ignorance could be tracked over time at the patient level, not just per image. From a methodological perspective, integrating causal modeling and counterfactual reasoning would help ensure that belief masses more faithfully capture disease mechanisms rather than stable but spurious correlates. Furthermore, using the uncertainty outputs to drive active learning or curriculum learning could systematically target failure modes. We will continue the research to build upon the statistical validation of our uncertainty quantification framework in XAI across text, tabular, and image data modalities. Furthermore, our goal is to study the statistical significance and hypothesis testing of out-of-distribution shifts and noisy real-world data when the framework is provided with such data. The hypothesis testing would be extended to the wide range of heterogeneous models to study the impact of the current study in multi-modal datasets. This will facilitate the expansion of the comparative analysis to a wide range of multi-modal architectures.

\section*{Acknowledgment}
One of the authors of this work has been financially
supported by the German Federal Ministry of Health (BMG)
under grant No.: ZMI5- 2523GHP027 (project "Strengthening National Immunization Technical Advisory Groups
and their Evidence-based Decision-making in the WHO
European Region and Globally" SENSE) part of the Global
Health Protection Programme, GHPP

\section*{Declaration of AI-assisted Technologies}
The authors would like to acknowledge their use of DeepL Write Pro (which uses AI) to polish the language and refine the grammar during the final preparation of this manuscript. All authors reviewed and approved the final text.



\clearpage 









\bibliography{cas-refs}


\end{document}